\documentclass{article}

\PassOptionsToPackage{numbers, compress}{natbib}
 \usepackage[preprint]{neurips_2026}

\usepackage[utf8]{inputenc} % allow utf-8 input
\usepackage[T1]{fontenc}    % use 8-bit T1 fonts
\makeatletter
\let\kernel@command@changed@warning\@gobble
\makeatother
\usepackage{hyperref}       % hyperlinks
\usepackage{booktabs}       % professional-quality tables
\usepackage{amsfonts}       % blackboard math symbols
\usepackage{nicefrac}       % compact symbols for 1/2, etc.
\usepackage{microtype}      % microtypography
\usepackage[table]{xcolor}         % colors

\usepackage{subcaption}
\usepackage{graphicx}
\usepackage{algorithm}
\usepackage{algpseudocode}
\usepackage{amsmath}
\usepackage{amssymb}
\usepackage{rotating}
\usepackage{multirow}
\usepackage{adjustbox}
\usepackage{array}
\usepackage{placeins}
\usepackage{afterpage}
\usepackage[ruled,vlined,algo2e]{algorithm2e}
\SetKwComment{Comment}{$\triangleright$\ }{}

\makeatletter
\newcommand{\labelalias}[2]{%
  \@bsphack
  \protected@write\@auxout{}%
    {\string\newlabel{#1}{{\theequation}{\thepage}{}{}{}}}%
  \@esphack
}
\makeatother

\title{VAGS: Velocity Adaptive Guidance Scale for Image Editing and Generation}

\author{%
  Yan Luo\textsuperscript{1,*},
  Ahmadou Aidara\textsuperscript{1,*},
  Jingyi Lu\textsuperscript{2,*},
  Jeremy Moebel\textsuperscript{1},
  Kai Han\textsuperscript{2,$\dagger$},
  Mengyu Wang\textsuperscript{1,3,$\dagger$}
  \\
  \textsuperscript{1}Harvard AI and Robotics Lab, Harvard University\\
  \textsuperscript{2}School of Computing and Data Science, The University of Hong Kong \\
  \textsuperscript{3}Kempner Institute for the Study of Natural and Artificial Intelligence, Harvard University
}

\begin{document}

\maketitle

\begingroup
\renewcommand{\thefootnote}{\fnsymbol{footnote}}
\footnotetext[1]{Equal contribution as co-first authors.}
\footnotetext[2]{Equal contribution as co-senior authors.}
\endgroup

\begin{abstract}
Classifier-free guidance (CFG) is the primary control over how strongly text
semantics move a flow-based sampler, yet standard practice holds its scale
fixed across the entire ODE trajectory. This is a fundamental mismatch: early
steps are noise-dominated and carry weak semantic signal, while late steps
commit image structure and demand stronger directional commitment; more
critically, the value of any guidance strength depends on whether the guided
velocity is consistent with the model's current dynamics or working against
them. We propose \textit{Velocity-Adaptive Guidance Scale} (VAGS), a
training-free replacement that multiplies the nominal scale by a bounded factor
combining a temporal signal-level term with the cosine similarity between
task-relevant velocity fields. For inversion-free editing, VAGS measures the
alignment between source- and target-guided velocities, so edit strength at
each step reflects local compatibility between preservation and transformation.
For generation, VAGS-Gen uses the alignment between unconditional and
conditional velocities as the analogous signal. Neither variant requires
fine-tuning, auxiliary networks, or extra forward passes, and fixed CFG is
recovered as a special case. On PIE-Bench and DIV2K for editing, and COCO17,
CUB-200, and Flickr30K for generation, VAGS consistently improves structural
fidelity and generation quality over fixed CFG and recent training-free
guidance variants. The code is publicly available at \url{https://github.com/Harvard-AI-and-Robotics-Lab/Velocity_Adaptive_Guidance_Scale}.
\end{abstract}
% \begin{abstract}
%   The abstract paragraph should be indented \nicefrac{1}{2}~inch (3~picas) on both the left- and right-hand margins. Use 10~point type, with a vertical spacing (leading) of 11~points. The word \textbf{Abstract} must be centered, bold, and in point size 12. Two line spaces precede the abstract. The abstract must be limited to one paragraph.
% \end{abstract}

\afterpage{%
  \begin{figure}[t]
  \centering
  \includegraphics[width=\linewidth]{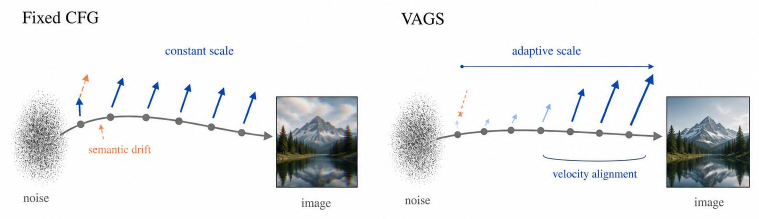}
  \caption{Overview of Velocity-Adaptive Guidance Scale (VAGS). 
Fixed CFG uses a constant scale along the sampling trajectory, which can cause
semantic drift and structural distortion. VAGS instead adaptively adjusts the
guidance scale according to denoising progress and velocity alignment, improving
semantic control while preserving image structure without extra training.}
  \label{fig:teaser}
  \vspace{-3ex}
\end{figure}

}

\section{Introduction}
\label{sec:intro}

Diffusion and flow-based generative models now define the dominant paradigm for
text-conditioned image synthesis, from iterative denoising
~\citep{ho2020ddpm,song2021scorebased} and latent diffusion
~\citep{rombach2022ldm} to DDIM-style sampling~\citep{song2021ddim},
flow matching~\citep{lipman2023flow}, rectified flow
~\citep{liu2023rectifiedflow}, and recent large-scale text-to-image systems
~\citep{esser2024sd3,flux2024}. Across these models, classifier-free guidance
(CFG)~\citep{ho2022cfg} is the standard mechanism for steering generation. It
extrapolates from unconditional to conditional predictions, compressing the
entire trade-off between prompt adherence and sample fidelity into a single
scalar. In flow-matching and rectified-flow models, that scalar directly
modulates the velocity field driving the sampling ODE, and therefore governs
the full trajectory from noise to image. The same control has become central to
image editing, including inversion-based editors
~\citep{hertz2022prompt2prompt,tumanyan2023pnp,mokady2023nulltext,
wallace2023edict,brooks2023instructpix2pix}, rectified-flow inversion methods
~\citep{rout2024rfinversion,wang2024rfsolver,deng2024fireflow,xu2025ftedit},
and recent inversion-free editors
~\citep{xu2023infedit,kulikov2024flowedit,yang2025irfds,yoon2025splitflow},
where edits are produced by integrating differences between source- and
target-conditioned velocities. CFG is not a peripheral sampling detail: it is
the primary control over how strongly text semantics move a flow-based sampler.
Fig.~\ref{fig:teaser} illustrates this fixed-scale failure mode: a constant
guidance scale can accumulate semantic drift along the sampling trajectory,
whereas VAGS adapts the scale using velocity alignment to reinforce reliable
directions and suppress harmful ones.

Holding this scalar constant across the entire ODE trajectory is the standard
practice, and it is the wrong one. Flow-based sampling does not evolve
uniformly. In the noisy regime, velocity differences carry weak or unstable
semantic signal, and strong guidance amplifies uninformative discrepancies that
pull the trajectory off the data manifold. In the clean regime, structure has
accumulated, semantic directions are sharp, and strong guidance is needed to
commit the visual details a prompt demands. Dynamic guidance schedules partially
recognize this nonuniformity~\citep{chang2023muse,wang2024analysis,
intervalguidance2024,papalampidi2025dynamic,galashov2025learn}, and recent
rectified-flow analyses connect CFG to off-manifold drift
~\citep{fan2025cfg,saini2025rectified}. However, monotone or searched schedules
still bypass the question that actually matters at each step: whether the
guided direction is consistent with the model's current dynamics, or whether it
is working against structure that has already formed.

The consequences are concrete. In text-driven editing, methods must transform
the requested content while preserving unrelated source structure. Existing
approaches expose this tension through attention control, inversion refinement,
or source-target velocity transport~\citep{hertz2022prompt2prompt,
tumanyan2023pnp,mokady2023nulltext,kulikov2024flowedit,yoon2025splitflow}. A
large fixed target scale achieves the requested object change but also
displaces background texture, identity, pose, and layout from the source image;
reducing it preserves the source but weakens or loses the edit. In generation,
a single scalar must serve both early global composition and late local
refinement, so the value that improves prompt adherence simultaneously
introduces saturation, duplicate objects, or brittle fine structure. The
problem is not finding the right average guidance strength; it is recognizing
that no fixed scalar can be right at every step.

Flow-based samplers already expose the signal needed to resolve this. Each
model call returns a velocity: the direction in latent space the sampler is
about to follow. When two relevant velocities are aligned, stronger guidance
reinforces motion the model already supports. When they conflict, stronger
guidance is an extrapolation against the reference dynamics, exactly the
condition under which unrelated structure drifts and artifacts emerge. Cosine
similarity directly quantifies this angular relationship. It is insensitive to
the timestep-dependent scale of the ODE and responds instead to the
compatibility that varies with prompt difficulty, image content, and sampling
stage. This is a clean, principled signal, and it has gone unused.

We introduce \textit{Velocity-Adaptive Guidance Scale} (VAGS), a training-free,
plug-and-play replacement for fixed CFG. Instead of applying one constant scale
throughout sampling, VAGS modulates the nominal scale at each step with a
bounded factor determined by denoising progress and velocity alignment. The
temporal term attenuates guidance in the noisy regime and strengthens it in the
clean regime; the geometric term amplifies guidance when task-relevant
velocities agree and suppresses it when they conflict. Fixed CFG is recovered as
a special case, and the added cost per step is a single inner product.

The formulation applies directly to both tasks. For inversion-free editing,
VAGS adapts the target scale from the alignment between the source-guided
velocity and a pilot target-guided velocity, so edit strength at each step
reflects the local compatibility between preservation and transformation. For
text-to-image generation, VAGS-Gen uses the alignment between unconditional and
conditional velocities as the analogous measure of compatibility with the data
prior. Neither variant requires extra model evaluations, external evaluators,
fine-tuning, or auxiliary networks.

Experiments confirm that velocity-adaptive guidance improves both tasks. On
PIE-Bench~\citep{ju2023direct} and DIV2K~\citep{agustsson2017ntire}, VAGS
improves source preservation without sacrificing edit strength when plugged into
FlowEdit~\citep{kulikov2024flowedit} and SplitFlow~\citep{yoon2025splitflow}.
On COCO17~\citep{lin2014microsoft}, CUB-200~\citep{welinder2010caltech}, and
Flickr30K~\citep{plummer2015flickr30k}, VAGS-Gen improves FID and Inception
Score over fixed CFG and recent training-free guidance variants
~\citep{jin2025flops,li2025self}. Ablations against interval
~\citep{kynkaanniemi2024applying}, monotone~\citep{wang2024analysis},
zero-initialized~\citep{fan2025cfg}, and matched-mean CFG schedules isolate
step-level velocity regulation as the source of the gains, ruling out any
benefit from a shifted average scale.

Our contributions are:
\begin{itemize}
    \item We identify fixed CFG as a shared bottleneck in flow-based image
    editing and generation, where a uniform scale ignores both the temporal
    structure of the ODE and the step-wise compatibility of the guided velocity.
    \item We propose VAGS, a unified adaptive guidance rule that combines signal
    level with velocity alignment at negligible computational overhead.
    \item We instantiate VAGS for editing and generation without additional
    training, networks, or model forward passes.
    \item We demonstrate consistent gains across five benchmarks and confirm
    through scheduler and matched-mean ablations that step-level adaptivity,
    not average scale, drives the improvement.
\end{itemize}

\section{Related Work}
\label{sec:related_work}

\noindent\textbf{Diffusion and Flow-based Generative Models.}
Diffusion models~\citep{ho2020ddpm,song2021scorebased} generate images by
iterative denoising, with later improvements including latent diffusion
~\citep{rombach2022ldm}, classifier-free guidance~\citep{ho2022cfg}, DDIM
sampling~\citep{song2021ddim}, and consistency distillation
~\citep{song2023consistency}. Flow Matching~\citep{lipman2023flow} and
Rectified Flow~\citep{liu2023rectifiedflow} instead learn an ODE velocity
field that transports noise to data, yielding a direct geometric object for
sampling control. This formulation underlies recent high-quality text-to-image
systems such as Stable Diffusion 3~\citep{esser2024sd3} and
FLUX~\citep{flux2024}.

\noindent\textbf{Text-driven Image Editing.}
Text-driven image editing modifies a real image according to a target prompt
while preserving unrelated content. Inversion-based diffusion editors first
map the image to a noisy latent and then regenerate it, often with attention
or inversion refinements~\citep{hertz2022prompt2prompt,tumanyan2023pnp,
mokady2023nulltext,wallace2023edict,brooks2023instructpix2pix}. Similar
inversion-based strategies have been adapted to rectified flows
~\citep{rout2024rfinversion,wang2024rfsolver,deng2024fireflow,xu2025ftedit},
but inversion error continues to trade editability against reconstruction.
Inversion-free editors avoid this by integrating source/target velocity
differences directly, as in InfEdit~\citep{xu2023infedit},
FlowEdit~\citep{kulikov2024flowedit}, iRFDS~\citep{yang2025irfds}, and
SplitFlow~\citep{yoon2025splitflow}. VAGS targets this latter family by
adapting the target guidance scale rather than changing the editing ODE.

\noindent\textbf{Guidance Scheduling.}
CFG is commonly applied with a fixed scale, although dynamic schedules often
improve the fidelity/adherence trade-off. Muse~\citep{chang2023muse} and
subsequent analyses~\citep{wang2024analysis,intervalguidance2024} show that
the best guidance strength varies across noise levels, while other methods
learn or search for dynamic scales~\citep{papalampidi2025dynamic,
galashov2025learn}. For rectified flows, recent work studies CFG-induced
off-manifold drift and proposes corrected guidance updates
~\citep{fan2025cfg,saini2025rectified}; for editing, FGS~\citep{cho2025fgs}
uses scheduled faithfulness guidance. VAGS is complementary: it is
training-free, uses no external controller, and adapts guidance through the
velocity alignment already available inside the sampler.

\section{Velocity-Adaptive Guidance Scale}
\label{sec:method}

We propose \emph{Velocity-Adaptive Guidance Scale} (VAGS) to replace the
constant scale with a per-step value controlled by two signals already exposed
by the ODE: the current signal level and the angular alignment of relevant
velocity fields. The same rule applies to inversion-free image editing and
text-to-image generation. Full pseudocodes are provided in
Algorithms~\ref{alg:vags} and~\ref{alg:vag_gen} in the appendix.
Fig.~\ref{fig:schematic} summarises the adaptive scale by a two-axis multiplier and the task-specific
workflows.

\begin{figure}[t]
  \centering
  \includegraphics[width=0.8\linewidth]{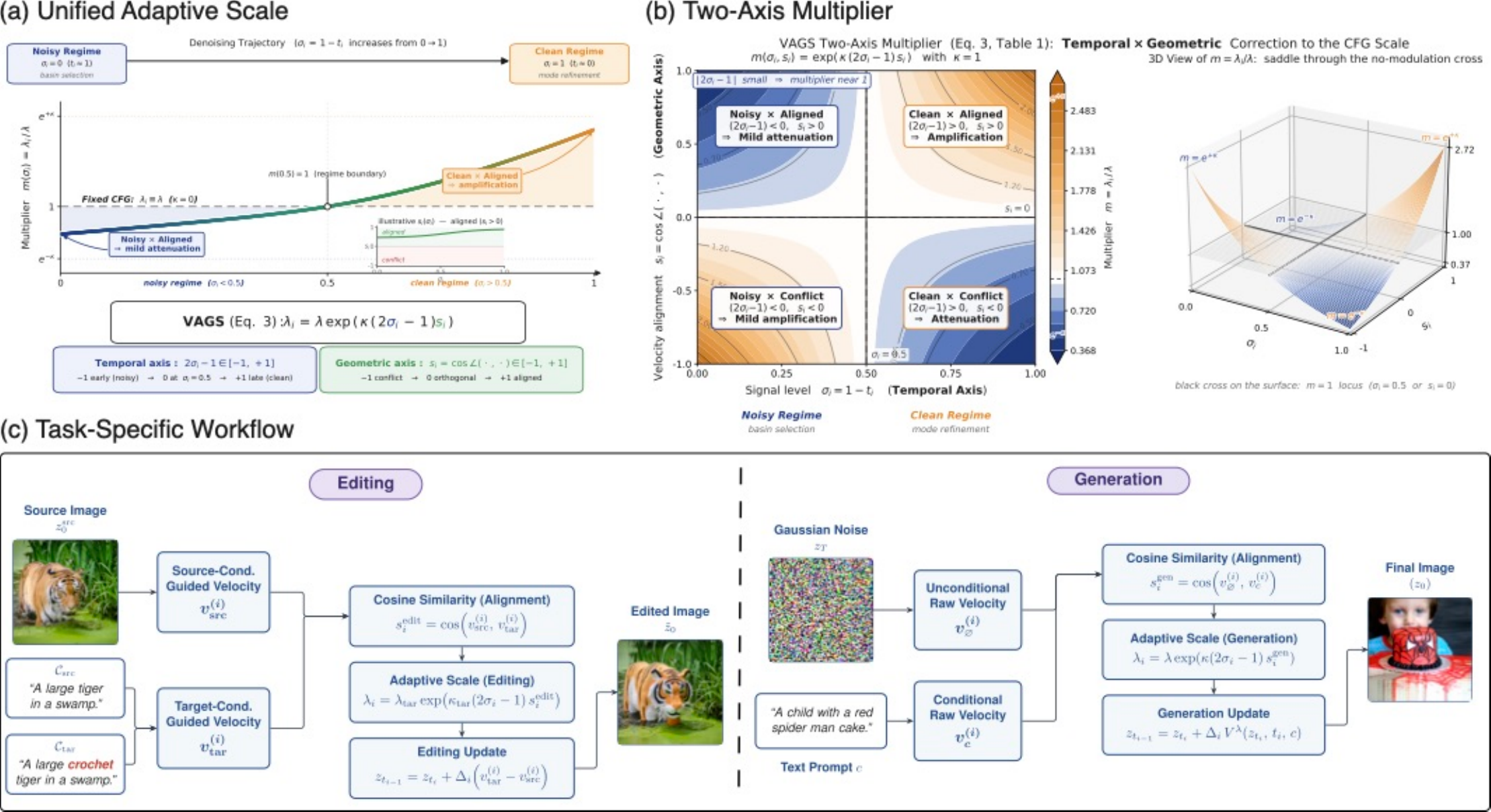}
  \caption{Workflow of Velocity-Adaptive Guidance Scale (VAGS). VAGS replaces fixed CFG
  with $\lambda_i=\lambda\exp(\kappa(2\sigma_i-1)s_i)$, where \(\sigma_i\)
  tracks denoising progress and \(s_i\) measures velocity alignment. The same
  rule applies to editing with source/target alignment and to generation with
  unconditional/conditional alignment.}
  \label{fig:schematic}
  \vspace{-3ex}
\end{figure}

% \subsection{A Unified Framework for Adaptive Guidance}
% \label{sec:framework}

\noindent\textbf{Flow matching and CFG.}
Flow-matching models~\citep{lipman2023flow,liu2023rectifiedflow} learn a
velocity field \(v_\theta\) that transports noise to data along the interpolant
\(z_t=(1-t)x+t\epsilon\), with \(z_0=x\) and \(z_1=\epsilon\). Sampling
integrates \(\mathrm{d}z_t/\mathrm{d}t=v_\theta(z_t,t,c)\) from \(t=1\) to
\(t=0\) over \(\{t_i\}_{i=1}^{N}\). At step \(i\), vanilla CFG combines the
unconditional and conditional velocities as
\begin{equation}
    V^{\lambda}\!\bigl(z_{t_i},t_i,c\bigr)
    =
    v_\varnothing^{(i)}
    +
    \lambda\,\bigl(v_c^{(i)}-v_\varnothing^{(i)}\bigr),
    \label{eq:cfg}
\end{equation}
where \(\lambda\) is held fixed in standard rectified-flow samplers such as
SD3~\citep{esser2024sd3} and FLUX~\citep{flux2024}.

\noindent\textbf{Shared sampler form.}
Both generation and inversion-free editing can be written as an Euler update
using a reference and target guided velocity:
\begin{equation}
    z_{t_{i-1}}
    =
    z_{t_i}
    +
    \Delta_i\,
    \mathcal{F}\!\bigl(V^{\mathrm{ref}}_{t_i},V^{\mathrm{tgt}}_{t_i}\bigr),
    \qquad
    \Delta_i=t_{i-1}-t_i<0 .
    \label{eq:unified_step}
\end{equation}
For generation, \(V^{\mathrm{ref}}=0\) and
\(\mathcal{F}(0,V^{\mathrm{tgt}})=V^{\mathrm{tgt}}\). For
inversion-free editing~\citep{kulikov2024flowedit,yoon2025splitflow},
\(V^{\mathrm{ref}}\) and \(V^{\mathrm{tgt}}\) are source- and
target-prompt guided velocities, and
\(\mathcal{F}(V^{\mathrm{ref}},V^{\mathrm{tgt}})=
V^{\mathrm{tgt}}-V^{\mathrm{ref}}\). Thus CFG controls the strength of the
same target field in both tasks.

\noindent\textbf{Fixed-scale bottleneck.}
A constant \(\lambda\) treats all steps as equally trustworthy. This conflicts
with the trajectory geometry: early latents are noisy and weakly semantic,
whereas late latents contain committed structure and sharper prompt
directions. A constant scale is also content-blind. It cannot distinguish a
step where the target direction agrees with the model's current dynamics from
one where guidance fights the trajectory, the latter being the setting where
high CFG produces off-manifold artefacts in rectified flows
~\citep{fan2025cfg,saini2025rectified}.

\noindent\textbf{Per-step adaptive scale.}
The adaptive rule should satisfy three constraints. First, it should preserve
the familiar CFG scale as the nominal strength, so existing sampler settings
remain meaningful. Second, it should vary smoothly across the trajectory,
because abrupt scale changes can introduce artifacts of their own. Third, it
should be symmetric around fixed CFG: agreement should raise the scale and
conflict should lower it, with the range controlled by one modulation
parameter.

Let \(\sigma_i=1-t_i\) denote signal level and let \(s_i\in[-1,1]\) be the
cosine similarity between two task-specific velocities at step \(i\). VAGS
sets
\begin{equation}
    \lambda_i
    =
    \lambda
    \exp\!\Bigl(\kappa(2\sigma_i-1)s_i\Bigr),
    \label{eq:adaptive}
\end{equation}
where \(\kappa\) controls modulation strength
(Fig.~\ref{fig:schematic}a). The temporal factor
\((2\sigma_i-1)\) changes sign from noisy to clean regimes, while the cosine
measures whether the local velocity geometry supports stronger guidance. In
the clean regime, aligned directions are amplified and conflicting directions
are damped to preserve committed structure. In the noisy regime, the
correction remains mild because the temporal factor is small near the middle
of the trajectory and the multiplier is bounded by
\([e^{-\kappa},e^\kappa]\); Fig.~\ref{fig:schematic}b visualises the
resulting four quadrants of the temporal--geometric multiplier. Setting
\(\kappa=0\) recovers fixed CFG exactly.
The exponential form makes the adaptation multiplicative rather than an
absolute schedule, keeping the base scale interpretable while letting local
velocity geometry reshape the trajectory.

\noindent\textbf{VAGS for image editing.}
Given a source image \(x_\mathrm{src}\), source prompt \(c_\mathrm{src}\), and
target prompt \(c_\mathrm{tar}\), inversion-free editing couples noisy source
and target latents as
\begin{equation}
    \hat{z}^{\,\mathrm{src}}_{t_i}
        =(1-t_i)x_\mathrm{src}+t_i\epsilon_i,
    \qquad
    \hat{z}^{\,\mathrm{tar}}_{t_i}
        =z^\mathrm{edit}_{t_i}+\hat{z}^{\,\mathrm{src}}_{t_i}-x_\mathrm{src},
    \label{eq:edit_couple}
\end{equation}
with \(\epsilon_i\sim\mathcal{N}(0,I)\), then advances the edit by
\begin{equation}
    z^\mathrm{edit}_{t_{i-1}}
    =
    z^\mathrm{edit}_{t_i}
    +
    (t_{i-1}-t_i)\bigl(V^\mathrm{tar}_{t_i}-V^\mathrm{src}_{t_i}\bigr).
    \label{eq:edit_step}
\end{equation}
VAGS keeps the source scale fixed and adapts only the target scale, since the
target field controls the requested semantic change.

This choice separates the two roles in the editing update. The source-guided
velocity acts as the preservation anchor: it describes how the original image
would move under its own prompt. The target-guided velocity supplies the
requested semantic change. Adapting the source scale would move the anchor
itself, making preservation changes harder to attribute. We therefore keep the
source branch fixed and ask whether the target branch is locally compatible
with that anchor.

The editing alignment signal is the cosine between the source guided velocity
and a one-shot pilot target velocity. Let \(u^\mathrm{src}_i,p^\mathrm{src}_i\)
and \(u^\mathrm{tar}_i,p^\mathrm{tar}_i\) be unconditional and conditional raw
predictions for the source and target latents. We form
\begin{equation}
    \lambda^{\mathrm{tar},\mathrm{base}}_i
    =
    \lambda_\mathrm{tar}\exp\!\bigl(\kappa_\mathrm{tar}(2\sigma_i-1)\bigr),
    \qquad
    V^{\mathrm{tar},\mathrm{pilot}}_{t_i}
    =
    u^\mathrm{tar}_i+
    \lambda^{\mathrm{tar},\mathrm{base}}_i(p^\mathrm{tar}_i-u^\mathrm{tar}_i),
    \label{eq:pilot}
\end{equation}
and then set
\begin{equation}
    s^\mathrm{edit}_i
    =
    \frac{V^\mathrm{src}_{t_i}\cdot V^{\mathrm{tar},\mathrm{pilot}}_{t_i}}
         {\|V^\mathrm{src}_{t_i}\|\,
          \|V^{\mathrm{tar},\mathrm{pilot}}_{t_i}\|},
    \qquad
    \lambda^\mathrm{tar}_i
    =
    \lambda_\mathrm{tar}
    \exp\!\Bigl(\kappa_\mathrm{tar}(2\sigma_i-1)s^\mathrm{edit}_i\Bigr).
    \label{eq:edit_adaptive}
\end{equation}
% \labelalias{eq:s_edit}{eq:edit_adaptive}
The pilot avoids iterative recomputation, and all quantities come from the
same four-way batch already used by the editing baseline.

\noindent\textbf{VAGS for Text-to-Image Generation.}
For generation, the latent starts from \(z_{t_N}\sim\mathcal{N}(0,I)\) and
follows
\begin{equation}
    z_{t_{i-1}}
    =
    z_{t_i}+(t_{i-1}-t_i)V^{\lambda}\!\bigl(z_{t_i},t_i,c\bigr).
    \label{eq:gen_step}
\end{equation}
There is no source field, so VAGS-Gen uses the unconditional and conditional
raw velocities already required by CFG:
\begin{equation}
    s^\mathrm{gen}_i
    =
    \frac{v^{(i)}_\varnothing\cdot v^{(i)}_c}
         {\|v^{(i)}_\varnothing\|\,
          \|v^{(i)}_c\|},
    \qquad
    \lambda_i
    =
    \lambda
    \exp\!\Bigl(\kappa(2\sigma_i-1)s^\mathrm{gen}_i\Bigr).
    \label{eq:gen_adaptive}
\end{equation}
% \labelalias{eq:s_gen}{eq:gen_adaptive}
Each step therefore performs the same two-way unconditional/conditional batch
as vanilla CFG, followed by one cosine similarity and scalar re-weighting.
VAGS-Gen is a drop-in replacement for fixed CFG with no auxiliary network,
fine-tuning, or extra forward pass.

\section{Experiments}
\label{sec:experiments}

% \subsection{Setup}

\noindent\textbf{Evaluation scope.}
The evaluation tests VAGS on two uses of classifier-free guidance: inversion-free
editing and text-to-image generation. The main editing criterion is the
preservation/editability trade-off, while the generation study tests whether the
same velocity-alignment signal improves sample quality without weakening prompt
alignment. Scheduler and matched-mean ablations isolate adaptive guidance from
simple changes to the average CFG scale, and \(\kappa\) sweeps measure
modulation sensitivity.

\noindent\textbf{Datasets and metrics.}
For editing, we evaluate on PIE-Bench~\citep{ju2023direct} and
DIV2K~\citep{agustsson2017ntire}. PIE-Bench provides 700 image-prompt pairs
across ten edit categories with foreground masks, and DIV2K stresses
preservation on high-resolution natural images with fine texture and detail.
Editing metrics cover structure distance~\citep{tumanyan2023pnp},
background PSNR/LPIPS~\citep{zhang2018perceptual}/MSE/SSIM, and whole-image
and edited-region CLIP similarity~\citep{wu2021godiva}. For generation, we
use COCO17~\citep{lin2014microsoft}, CUB-200~\citep{welinder2010caltech},
and Flickr30K~\citep{plummer2015flickr30k}, spanning broad captions,
fine-grained bird species, and everyday scenes. We report
FID~\citep{heusel2017gans}, IS~\citep{salimans2016improved},
CLIPScore~\citep{hessel2021clipscore}, and caption metrics
BLEU-4~\citep{papineni2002bleu}, METEOR~\citep{banerjee2005meteor},
ROUGE-L~\citep{lin2004rouge}, and CLAIR.

\noindent\textbf{Implementation.}
All experiments use Stable Diffusion~3.5 Large. Images are resized and
center-cropped to \(512\times512\) before encoding. Runs use
PyTorch~\citep{paszke2019pytorch} and
\texttt{diffusers}~\citep{von-platen-etal-2022-diffusers} on one NVIDIA
H100 80GB GPU. For editing, VAGS is plugged into
FlowEdit~\citep{kulikov2024flowedit} and
SplitFlow~\citep{yoon2025splitflow} with \(N=50\), \(N_{\max}=33\),
\(\lambda_\mathrm{src}=3.5\), \(\lambda_\mathrm{tar}=13.5\), and
\(\kappa_\mathrm{tar}=0.9\). For generation, VAGS-Gen uses \(N=25\), base
CFG \(\lambda=7.0\), \(\kappa=1.0\) on COCO17 and CUB-200, and
\(\kappa=0.9\) on Flickr30K.

\noindent\textbf{Baselines.}
Editing baselines cover diffusion inversion and attention editors, including
DDIM+P2P~\citep{song2021ddim,hertz2022prompt2prompt},
DDIM+PnP~\citep{song2021ddim,tumanyan2023pnp}, and
Null-text+P2P~\citep{mokady2023nulltext,hertz2022prompt2prompt};
rectified-flow inversion methods
~\citep{rout2024rfinversion,wang2024rfsolver,deng2024fireflow,xu2025ftedit};
and inversion-free editors~\citep{kulikov2024flowedit,yoon2025splitflow,
yang2025irfds}. Scheduler ablations compare
Interval~\citep{kynkaanniemi2024applying}, Monotone~\citep{wang2024analysis},
and Zero-Init~\citep{fan2025cfg}. For generation, we compare fixed CFG
SDv3.5, A-Euler~\citep{jin2025flops}, and
Self-Guidance~\citep{li2025self}.

% \FloatBarrier
% \subsection{Quantitative Evaluation}
% \label{sec:quant}

\noindent\textbf{Quantitative Evaluation for Editing.}
Tab.~\ref{tab:sota_comparison} shows that VAGS improves preservation without
sacrificing edit strength. On PIE-Bench, FlowEdit+VAGS reduces Dist from
30.31 to 13.84 and MSE by 62\%, while CLIP scores remain stable or improve.
SplitFlow+VAGS follows the same trend (Dist \(28.03\rightarrow14.67\),
Tab.~\ref{tab:guidance_ablation}). On DIV2K, VAGS gives the best Dist, PSNR,
MSE, and LPIPS among all compared methods, including inversion-based editors.
These results support the main claim: adapting guidance by velocity alignment
improves preservation while retaining semantic edits.

This pattern is important because the metrics reward different behavior.
Structure distance and background metrics measure whether the source image
survives the edit, while CLIP-based scores measure whether the target prompt is
still expressed. A trivial CFG reduction would usually improve preservation at
the cost of edit strength. VAGS instead improves the preservation metrics
while keeping semantic metrics stable, the expected signature of step-level
regulation rather than global weakening.

\FloatBarrier
% Required packages: booktabs, multirow, graphicx (for \resizebox)
\begin{table*}[t!]
\centering
\small
\setlength{\tabcolsep}{4.5pt}
\renewcommand{\arraystretch}{1.08}
\resizebox{\textwidth}{!}{%
\begin{tabular}{l l c c c c c c c c c}
\toprule
\multirow{2}{*}{\textbf{Dataset}} &
\multirow{2}{*}{\textbf{Method}} &
\multirow{2}{*}{\textbf{Model}} &
\multicolumn{1}{c}{\textbf{Structure}} &
\multicolumn{5}{c}{\textbf{Background Preservation}} &
\multicolumn{2}{c}{\textbf{CLIP Similarity}} \\
\cmidrule(lr){4-4}\cmidrule(lr){5-9}\cmidrule(lr){10-11}
& & &
\textbf{Dist.\,$\times10^{3}\downarrow$} &
\textbf{PSNR $\uparrow$} &
\textbf{LPIPS $\times10^{3}\downarrow$} &
\textbf{MSE $\times10^{4}\downarrow$} &
\textbf{SSIM $\times10^{2}\uparrow$} &
\textbf{CLIP-I $\uparrow$} &
\textbf{Whole $\uparrow$} &
\textbf{Edited $\uparrow$}
\\
\midrule

%% ── PIE-Bench ──────────────────────────────────────────────────────────
\multirow{18}{*}{\textbf{PIE-Bench}}
& \multicolumn{9}{l}{\textit{Diffusion-based}} \\
& DDIM~\cite{song2021ddim} + P2P~\cite{hertz2022prompt2prompt}
  % & SD~1.4  & 69.40 & 17.87 & 208.80 & 219.88 & 22.44 & 71.14 & 25.01\\
  & SD~1.4 & 71.12 & 17.84 & 207.62 & 220.79 & 71.62 & 82.44 & 25.91 & 22.75\\
& DDIM~\cite{song2021ddim} + PnP~\cite{tumanyan2023pnp}
  % & SD~1.4  & 28.22 & 22.28 & 113.46 & 83.64 & 22.55 & 79.05 & 25.41\\
  & SD~1.4 & 27.27 & 22.36 & 110.22 & 81.94 & 79.65 & 86.93 & 26.13 & 22.79\\
& Null-text~\cite{mokady2023nulltext} + P2P~\cite{hertz2022prompt2prompt}
  % & SD~2.1  & 13.44 & 27.03 & 60.67  & 35.86 & 21.86 & 84.11 & 24.75\\
  & SD~1.4 & 15.69 & \textbf{26.92} & \textbf{66.42} & \underline{35.09} & 82.67 & 85.42 & 25.11 & 22.01\\
\cmidrule(l){2-11}
& \multicolumn{9}{l}{\textit{Flow-based / Rectified Flow}} \\
& RF-Inversion~\cite{rout2024rfinversion}
  % & FLUX.1  & 40.60 & 20.82 & 184.80 & 129.10 & 22.11 & 71.92 & 25.20\\
  & FLUX.1 & 47.19 & 20.64 & 193.68 & 123.33 & 76.34 & 80.23 & 26.16 & 22.38\\
& RF-Solver~\cite{wang2024rfsolver}
  % & FLUX.1  & 31.10 & 22.90 & 135.81 & 80.11 & 22.88 & 81.90 & 26.00\\
  & FLUX.1 & 48.04 & 20.52 & 181.15 & 128.54 & 77.81 & 82.20 & 26.63 & 23.46\\
& FireFlow~\cite{deng2024fireflow} + RF-Solver~\cite{wang2024rfsolver}
  % & FLUX.1  & 28.30 & 23.28 & 120.82 & 70.39 & 22.94 & 82.82 & 25.98\\
  & FLUX.1 & 25.88 & 22.90 & 129.27 & 74.73 & 82.03 & 84.76 & 25.96 & 22.02\\
& FlowEdit~\cite{kulikov2024flowedit}
  % & FLUX.1  & 27.70 & 21.91 & 111.70 & 94.00 & 22.70 & 83.39 & 25.61\\
  & FLUX.1 & 27.93 & 21.74 & 113.83 & 98.75 & 83.46 & 86.54 & 26.19 & 22.19\\
& FTEdit~\cite{xu2025ftedit} + AdaLN
  % & SD~3.5  & 18.17 & 26.62 & 80.55  & 40.24 & 22.27 & 91.50 & 25.74\\
  & SD~3.5 & \textbf{13.81} & 23.69 & 77.69 & 36.10 & \textbf{92.62} & \textbf{92.40} & 25.01 & 21.81\\
& iRFDS~\cite{yang2025irfds}
  % & SD~3.5  & 45.04 & 16.25 & 312.24 & 305.83 & 25.38 & 57.03 & 25.46\\
  & SD~3.5 & 67.64 & 19.42 & 150.68 & 147.86 & 79.47 & 77.69 & 25.45 & 21.73\\
& FlowEdit~\cite{kulikov2024flowedit}
  % & SD~3.5  & 30.32 & 18.31 & 207.09 & 169.11 & 27.56 & 68.96 & 27.63\\
  & SD~3.5 & 30.31 & 21.87 & 117.06 & 92.70 & 82.54 & 83.69 & \textbf{27.62} & \textbf{23.80}\\
& SplitFlow ~\cite{yoon2025splitflow}
  & SD~3.5  & 28.03 & 22.32 & 111.49 & 85.31 & 83.17 & 84.10 & \underline{27.52} & \underline{23.77}\\
%& SplitFlow (Official Repo)~\cite{yoon2025splitflow}
 % & SD~3  & 27.20 & 21.94 & 105.48 & 92.35 & 83.41 & - & 27.51 & 23.89\\
%& SplitFlow (SplitFlow Paper)~\cite{yoon2025splitflow}
 % & SD~3  & 25.96 & 22.45 & 102.14 & 81.99 & 83.91 & - & 26.96 & 23.83\\
\cmidrule(l){2-11}
& \multicolumn{9}{l}{\textit{Ours}} \\

  & \cellcolor{gray!15}FlowEdit + VAGS
  % & SD~3.5  & 23.52 & 19.74 & 186.19 & 125.96 & 27.44 & 71.43 & 27.53\\
  & \cellcolor{gray!15}SD~3.5 & \cellcolor{gray!15}\underline{13.84} & \cellcolor{gray!15}\underline{26.38} & \cellcolor{gray!15}\underline{70.38} & \cellcolor{gray!15}\textbf{34.86} & \cellcolor{gray!15}\underline{87.68} & \cellcolor{gray!15}\underline{87.44} & \cellcolor{gray!15}26.92 & \cellcolor{gray!15}23.08\\
  % & FlowEdit + VAGS-Cosine (Updated w/ $\kappa$=0.9) & SD~3.5 &  23.52 & 23.49 & 103.56 & 66.92 & 83.97 & 84.45 & 27.54 & 23.65\\
  % & FlowEdit + VAGS-Cosine (alt run,  $\kappa$=1.5) & SD~3.5 & 15.61 & 25.46 & 99.51 & 41.40 & 84.31 & 86.42 & 27.01 & 23.00\\
% & FlowEdit + VAGS-Ratio
  % & SD~3.5  & 24.21 & 19.44 & 183.17 & 131.64 & 27.40 & 71.86 & 27.50\\
  % & SD~3.5 & 13.54 & 26.29 & 64.35 & 35.50 & 88.24 & 87.85 & 26.93 & 23.04\\
  % & FlowEdit + VAGS-Cosine (Updated w/ $\kappa$=0.5) & SD~3.5 &  25.55 & 22.84 & 104.88 & 75.61 & 83.81 & 84.44 & 27.50 & 23.76\\
  % & FlowEdit + VAGS-Ratio (alt run,  $\kappa$=0.5) & SD~3.5 & 14.08 & 25.35 & 77.91 & 43.82 & 86.31 & 87.80 & 26.96 & 23.18\\

\midrule

%% ── DIV2K ───────────────────────────────────────────────────────────────
\multirow{18}{*}{\textbf{DIV2K}}
& \multicolumn{9}{l}{\textit{Diffusion-based}} \\
& DDIM~\cite{song2021ddim} + P2P~\cite{hertz2022prompt2prompt}
  % & SD~1.4  & 78.50 & 13.74 & 383.42 & 455.00 & 29.50 & 51.35 & 29.82\\
  & SD~1.4 & 64.46 & 15.00 & 323.57 & 359.10 & 60.93 & 81.89 & 26.45 & -\\
& DDIM~\cite{song2021ddim} + PnP~\cite{tumanyan2023pnp}
  % & SD~1.4  & 31.60 & 18.02 & 228.35 & 179.00 & 28.30 & 65.31 & 28.56\\
  & SD~1.4 & 30.66 & 18.75 & 205.72 & 154.73 & 68.61 & 83.83 & 28.37 & -\\
& Null-text~\cite{mokady2023nulltext} + P2P~\cite{hertz2022prompt2prompt}
  % & SD~2.1  & 24.50 & 19.15 & 170.20 & 142.40 & 27.25 & 76.20 & 27.59\\
  & SD~1.4 & 15.82 & 21.35 & 137.28 & 93.17 & 74.37 & \underline{88.32} & 27.61 & -\\
\cmidrule(l){2-11}
& \multicolumn{9}{l}{\textit{Flow-based / Rectified Flow}} \\
& RF-Inversion~\cite{rout2024rfinversion}
  % & FLUX.1  & 55.40 & 16.20 & 320.60 & 285.40 & 30.85 & 62.40 & 30.99\\
  & FLUX.1 & 31.87 & 20.05 & 252.75 & 115.45 & 73.37 & 79.57 & 29.60 & -\\
& RF-Solver~\cite{wang2024rfsolver}
  % & FLUX.1  & 48.20 & 16.55 & 245.20 & 272.10 & 30.80 & 64.15 & 30.96\\
  & FLUX.1 & 34.63 & 18.56 & 265.56 & 161.60 & 72.98 & 79.54 & 29.90 & -\\
& FireFlow~\cite{deng2024fireflow} + RF-Solver~\cite{wang2024rfsolver}
  % & FLUX.1  & 25.40 & 19.11 & 210.40 & 136.00 & 29.60 & 71.06 & 29.75\\
  & FLUX.1 & 20.52 & 21.02 & 197.26 & 92.75 & 77.88 & 82.86 & 28.77 & -\\
& FlowEdit~\cite{kulikov2024flowedit}
  % & FLUX.1  & 27.00 & 17.83 & 191.09 & 183.00 & 29.85 & 73.03 & 30.06\\
  & FLUX.1 & 24.84 & 18.73 & 173.21 & 152.33 & 78.81 & 83.93 & 29.07 & -\\
& FTEdit~\cite{xu2025ftedit} + AdaLN
  % & SD~3.5  & 26.00 & 19.80 & 184.76 & 118.00 & 29.05 & 79.50 & 29.13\\
  & SD~3.5 & \underline{14.43} & \underline{23.02} & \underline{125.97} & \underline{56.65} & \underline{83.73} & \textbf{88.97} & 27.03 & -\\
& iRFDS~\cite{yang2025irfds}
  % & SD~3.5  & 65.40 & 18.50 & 208.30 & 195.50 & 28.35 & 71.20 & 28.55\\
  & SD~3.5 & 61.14 & 15.86 & 317.41 & 320.07 & 63.02 & 76.79 & 27.47 & -\\
& FlowEdit~\cite{kulikov2024flowedit}
  % & SD~3.5  & 25.15 & 20.27 & 145.63 & 103.77 & 30.25 & 82.15 & 30.33\\
  & SD~3.5 & 18.27 & 20.27 & 145.55 & 103.37 & 82.15 & 82.31 & \textbf{30.06} & -\\
& SplitFlow~\cite{yoon2025splitflow}
  % & SD~3.5  & 24.77 & 20.53 & 141.11 & 97.52 & 30.45 & 82.73 & 30.57\\
  & SD~3.5 & 17.61 & 20.53 & 141.99 & 97.19 & 82.63 & 82.44 & \underline{29.95} & -\\
\cmidrule(l){2-11}
& \multicolumn{9}{l}{\textit{Ours}} \\
& \cellcolor{gray!15}FlowEdit + VAGS
  % & SD~3.5  & NA & 21.70 & NA & 75.90 & 29.87 & 82.35 & 30.01\\
  & \cellcolor{gray!15}SD~3.5 & \cellcolor{gray!15}\textbf{8.62} & \cellcolor{gray!15}\textbf{23.61} & \cellcolor{gray!15}\textbf{98.43} & \cellcolor{gray!15}\textbf{50.22} & \cellcolor{gray!15}\textbf{84.41} & \cellcolor{gray!15}84.07 & \cellcolor{gray!15}29.61 & \cellcolor{gray!15}-\\
% & FlowEdit + VAGS-Ratio
  % & SD~3.5 & NA & 21.01 & NA & 86.87 & 29.81 & 81.56 & 29.91\\
  % & SD~3.5 & 9.38 & 23.14 & 95.70 & 54.67 & 84.39 & 84.39 & 29.62 & -\\

\bottomrule
\end{tabular}%
}
\caption{Quantitative comparison on PIE-Bench and DIV2K. FlowEdit + VAGS ranks first on all background preservation metrics while maintaining competitive edit strength.}
\label{tab:sota_comparison}
\end{table*}

\noindent\textbf{Quantitative Evaluation Generation.}
Tab.~\ref{tab:perf_comparison} evaluates VAGS-Gen on three datasets. On
COCO17, VAGS-Gen improves FID from 28.46 to 26.07 and IS from 33.64 to 35.15,
outperforming fixed CFG, A-Euler, and Self-Guidance. The gain is larger on
CUB-200, where FID drops from 24.92 to 20.60, and remains positive on
Flickr30K (79.58 to 75.65). Caption metrics stay competitive, indicating that
the FID/IS gains do not come from ignoring the prompt. Appendix trajectory
diagnostics show that VAGS-Gen reshapes the CFG path using
unconditional/conditional velocity alignment, while prompt-wise mean scales
remain concentrated.

\FloatBarrier
% Required packages: booktabs, multirow, graphicx, xcolor
\begin{table*}[t!]
\centering
\small
\setlength{\tabcolsep}{4.5pt}
\renewcommand{\arraystretch}{1.08}
\resizebox{\textwidth}{!}{%
\begin{tabular}{l l c c c c c c c}
\toprule
\multirow{2}{*}{\textbf{Dataset}} &
\multirow{2}{*}{\textbf{Method}} &
\multirow{2}{*}{\textbf{FID} {$\downarrow$}} &
\multirow{2}{*}{\textbf{IS} {$\uparrow$}} &
\multirow{2}{*}{\textbf{CLIPScore} {$\uparrow$}} &
\multicolumn{4}{c}{\textbf{Caption-based Metrics}} \\
\cmidrule(lr){6-9}
& & & & &
\textbf{BLEU-4} {$\uparrow$} &
\textbf{METEOR} {$\uparrow$} &
\textbf{ROUGE-L} {$\uparrow$} &
\textbf{CLAIR} {$\uparrow$} \\
\midrule

\multirow{4}{*}{\textbf{COCO17} \cite{lin2014microsoft}}
& SDv3.5 (CFG=7) & 28.46 & 33.64 & \textbf{32.84} & \underline{7.99} & \underline{29.17} & 35.11 & \underline{71.45} \\
& SDv3.5 w/ A-Euler \cite{jin2025flops} & \underline{27.64} & \underline{34.63} & \textbf{32.84} & 7.96 & 29.09 & \underline{35.17} & 71.23 \\
& SDv3.5 w/ Self-Guidance \cite{li2025self} & 40.98 & 28.54 & 31.21 & 6.52 & 26.18 & 31.59 & 62.26 \\
& \cellcolor{gray!15}SDv3.5 w/ VAGS (CFG=7, $\kappa$=1.0)
  & \cellcolor{gray!15}\textbf{26.07} & \cellcolor{gray!15}\textbf{35.15} & \cellcolor{gray!15}\textbf{32.84}
  & \cellcolor{gray!15}\textbf{8.28} & \cellcolor{gray!15}\textbf{29.53} & \cellcolor{gray!15}\textbf{35.21} & \cellcolor{gray!15}\textbf{71.62} \\
\midrule

\multirow{4}{*}{\textbf{CUB-200} \cite{welinder2010caltech}}
& SDv3.5 (CFG=7) & 24.92 & 4.81 & 32.40 & 0.13 & 16.57 & \underline{17.64} & \textbf{66.69} \\
& SDv3.5 w/ A-Euler \cite{jin2025flops} & \underline{22.98} & \underline{5.05} & \textbf{32.50} & 0.12 & 16.39 & 17.51 & 66.22 \\
& SDv3.5 w/ Self-Guidance \cite{li2025self} & 61.16 & 4.98 & 31.25 & \textbf{0.18} & \textbf{17.85} & \textbf{17.78} & 62.28 \\
& \cellcolor{gray!15}SDv3.5 w/ VAGS (CFG=7, $\kappa$=1.0)
  & \cellcolor{gray!15}\textbf{20.60} & \cellcolor{gray!15}\textbf{5.17} & \cellcolor{gray!15}\underline{32.45}
  & \cellcolor{gray!15}\underline{0.15} & \cellcolor{gray!15}\underline{17.00} & \cellcolor{gray!15}17.52 & \cellcolor{gray!15}\underline{66.52} \\
\midrule

\multirow{4}{*}{\textbf{Flickr30K} \cite{plummer2015flickr30k}}
& SDv3.5 (CFG=7) & 79.58 & \underline{17.95} & \textbf{33.79} & \underline{4.12} & \textbf{23.88} & \textbf{29.30} & \textbf{68.03} \\
& SDv3.5 w/ A-Euler \cite{jin2025flops} & \underline{78.57} & 17.83 & 33.71 & 3.91 & 22.99 & 29.00 & \underline{67.56} \\
& SDv3.5 w/ Self-Guidance \cite{li2025self} & 93.07 & 14.44 & 31.60 & 3.35 & 20.80 & 26.27 & 57.35 \\
& \cellcolor{gray!15}SDv3.5 w/ VAGS (CFG=7, $\kappa$=0.9)
  & \cellcolor{gray!15}\textbf{75.65} & \cellcolor{gray!15}\textbf{18.21} & \cellcolor{gray!15}\underline{33.77}
  & \cellcolor{gray!15}\textbf{4.17} & \cellcolor{gray!15}\underline{23.43} & \cellcolor{gray!15}\underline{29.25} & \cellcolor{gray!15}\underline{67.65} \\
\bottomrule
\end{tabular}%
}
\caption{Text-to-image generation on COCO17, CUB-200, and Flickr30K. BLIP-2 \cite{li2023blip} is used to generate captions from the generated images, enabling evaluation with the caption-based metrics. VAGS-Gen yields consistent FID and IS gains.
% , with the largest improvement on fine-grained generation.
}
\label{tab:perf_comparison}
% \vspace{-2ex}
\end{table*}

\FloatBarrier
% Required packages: booktabs, multirow, graphicx (for \resizebox)
\begin{table*}[t!]
\centering
\small
\setlength{\tabcolsep}{4.5pt}
\renewcommand{\arraystretch}{1.08}
\resizebox{\textwidth}{!}{%
\begin{tabular}{l l l c c c c c c c c}
\toprule
\multirow{2}{*}{\textbf{Dataset}} &
\multirow{2}{*}{\textbf{Editor}} &
\multirow{2}{*}{\textbf{Scheduler}} &
\multicolumn{1}{c}{\textbf{Structure}} &
\multicolumn{5}{c}{\textbf{Background Preservation}} &
\multicolumn{2}{c}{\textbf{CLIP Similarity}} \\
\cmidrule(lr){4-4}\cmidrule(lr){5-9}\cmidrule(lr){10-11}
& & &
\textbf{Dist.\,$\times10^{3}\downarrow$} &
\textbf{PSNR $\uparrow$} &
\textbf{LPIPS $\times10^{3}\downarrow$} &
\textbf{MSE $\times10^{4}\downarrow$} &
\textbf{SSIM $\times10^{2}\uparrow$} &
\textbf{CLIP-I $\uparrow$} &
\textbf{Whole $\uparrow$} &
\textbf{Edited $\uparrow$}
\\
\midrule

%% ── PIE-Bench ──────────────────────────────────────────────────────────
\multirow{10}{*}{\textbf{PIE-Bench}}
& \multirow{5}{*}{FlowEdit}
% & No Scheduler
%   & 11.65 & 27.22 & 54.55 & 32.86 & 22.10 & 84.76 & 25.02\\
 & No Scheduler
  % & 30.32 & 18.31 & 207.09 & 169.11 & 27.56 & 68.96 & 27.63\\
   & 30.31 & 21.87 & 117.06 & 92.70 & 82.54 & 83.69 & \underline{27.62} & \underline{23.80}\\
&&  Interval~\cite{kynkaanniemi2024applying} 
  % & 29.82 & 18.35 & 202.47 & 167.47 & 27.51 & 69.61 & 27.57\\
  & 29.82 & 21.93 & 114.14 & 91.72 & 82.88 & 84.03 & 27.58 & 23.75\\
&& Monotone~\cite{wang2024analysis}
  % & 15.91 & 21.49 & 132.86 & 85.86 & 26.96 & 78.48 & 27.05\\
  & \underline{15.91} & \underline{25.43} & \textbf{69.87} & \underline{43.47} & \underline{87.67} & \underline{84.73} & 27.05 & 23.26\\
&& Zero-Init~\cite{fan2025cfg} 
  % & 194.67 & 8.98 & 560.19 & 1385.85 & 28.46 & 32.46 & 28.44\\
  & 194.67 & 11.69 & 354.57 & 867.59 & 57.64 & 75.28 & \textbf{28.44} & \textbf{24.24}\\
&& \cellcolor{gray!15}VAGS (Ours)
  % & 23.52 & 19.74 & 186.19 & 125.96 & 27.44 & 71.43 & 27.53\\
  & \cellcolor{gray!15}\textbf{13.84} & \cellcolor{gray!15}\textbf{26.38} & \cellcolor{gray!15}\underline{70.38} & \cellcolor{gray!15}\textbf{34.86} & \cellcolor{gray!15}\textbf{87.68} & \cellcolor{gray!15}\textbf{87.44} & \cellcolor{gray!15}26.92 & \cellcolor{gray!15}23.08\\
  % && VAGS-Ratio (Ours)
  %   & 24.21 & 19.44 & 183.17 & 131.64 & 27.40 & 71.86 & 27.50\\
  %   & 13.54 & 26.29 & 64.35 & 35.50 & 88.24 & 87.85 & 26.93 & 23.04\\
\cmidrule(l){2-11}

  & \multirow{5}{*}{ SplitFlow~\cite{yoon2025splitflow}}
& No Scheduler
  % & 28.04 & 18.62 & 199.44 & 158.13 & 27.45 & 69.99 & 27.52\\
  & 28.03 & 22.32 & 111.49 & 85.31 & 83.17 & 84.10 & \textbf{27.52} & \textbf{23.77}\\
&& Interval~\cite{kynkaanniemi2024applying}
  % & N/A & 18.96 & N/A & 147.42 & 27.39 & 71.49 & 27.43\\
  & 27.28 & 23.05 & 106.26 & 77.64 & 84.17 & 83.25 & 27.25 & 23.22\\
&& Monotone~\cite{wang2024analysis}
  % & N/A & 22.02 & N/A & 76.42 & 26.87 & 79.66 & 26.98\\
  & \underline{16.06} & \textbf{26.25} & \textbf{66.73} & \underline{38.77} & \textbf{88.12} & \underline{85.94} & 26.83 & 22.76\\
&& Zero-Init~\cite{fan2025cfg} 
  % & 21.66 & 19.85 & 170.21 & 121.57 & 27.23 & 73.37 & 27.30\\
  & 21.65 & 23.57 & 94.01 & 65.07 & 84.90 & 85.36 & \underline{27.30} & \underline{23.57}\\
&& \cellcolor{gray!15}VAGS (Ours)
  & \cellcolor{gray!15}\textbf{14.67} & \cellcolor{gray!15}\underline{25.99} & \cellcolor{gray!15}\underline{73.43} & \cellcolor{gray!15}\textbf{38.41} & \cellcolor{gray!15}\underline{87.31} & \cellcolor{gray!15}\textbf{87.22} & \cellcolor{gray!15}27.06 & \cellcolor{gray!15}23.28\\
  % && VAGS-Ratio (Ours)
  %   & 13.53 & 26.24 & 64.24 & 36.54 & 88.18 & 88.00 & 26.91 & 23.15\\
\midrule

%% ── DIV2K ──────────────────────────────────────────────────────────────
\multirow{10}{*}{\textbf{DIV2K}}
& \multirow{5}{*}{FlowEdit}
& No Scheduler
  % & 22.80 & 19.80 & 165.80 & 138.20 & 29.60 & 77.15 & 29.85\\
  & 18.27 & 20.27 & 145.55 & 103.37 & 82.15 & 82.31 & \textbf{30.06} & -\\
&& Interval~\cite{kynkaanniemi2024applying}
  % & N/A & 20.19 & N/A & 105.24 & 29.84 & 79.76 & 29.94\\
  & 18.11 & 19.94 & 144.88 & 111.55 & 79.51 & 82.32 & \underline{30.03} & -\\
&& Monotone~\cite{wang2024analysis} 
  % & N/A & 22.86 & N/A & 58.78 & 29.35 & 85.91 & 29.66\\
  & \underline{10.89} & \underline{22.47} & \underline{102.91} & \underline{63.91} & \underline{83.64} & \textbf{84.10} & 29.60 & -\\
&& Zero-Init~\cite{fan2025cfg}
  % & N/A & 10.50 & N/A & 944.15 & 30.78 & 44.12 & 30.73\\
  & 194.67 & 11.69 & 354.57 & 867.59 & 57.64 & 75.28 & 28.44 & -\\
&& \cellcolor{gray!15}VAGS (Ours)
  % & NA & 21.70 & NA & 75.90 & 29.87 & 82.35 & 30.01\\
  & \cellcolor{gray!15}\textbf{8.62} & \cellcolor{gray!15}\textbf{23.61} & \cellcolor{gray!15}\textbf{98.43} & \cellcolor{gray!15}\textbf{50.22} & \cellcolor{gray!15}\textbf{84.41} & \cellcolor{gray!15}\underline{84.07} & \cellcolor{gray!15}29.61 & \cellcolor{gray!15}-\\
  % && VAGS-Ratio (Ours)
  %   & NA & 21.01 & NA & 86.87 & 29.81 & 81.56 & 29.91\\
  %   & 9.38 & 23.14 & 95.70 & 54.67 & 84.39 & 84.39 & 29.62 & -\\
\cmidrule(l){2-11}

  & \multirow{5}{*}{ SplitFlow~\cite{yoon2025splitflow}}
& No Scheduler & 17.61 & 20.53 & 141.99 & 97.19 & 82.63 & 82.44 & \underline{29.95} & -\\
% & 24.77 & 20.53 & 141.11 & 97.52 & 30.45 & 82.73 & 30.57\\
& & Interval~\cite{kynkaanniemi2024applying} & 16.01 & 20.53 & 136.04 & 97.62 & 80.55 & 82.65 & 29.91 & -\\
% & N/A & 20.82 & N/A & 91.16 & 29.62 & 81.19 & 29.87\\
& & Monotone~\cite{wang2024analysis} & 9.63 & 23.01 & \textbf{97.16} & \underline{56.81} & \textbf{84.23} & \textbf{84.43} & 29.55 & -\\
% & N/A & 23.43 & N/A & 51.37 & 29.20 & 86.84 & 29.56\\
& & Zero-Init~\cite{fan2025cfg} & 46.72 & 14.83 & 378.94 & 345.27 & 48.95 & 75.93 & \textbf{31.58} & -\\
% & N/A & 21.86 & N/A & 73.26 & 29.30 & 82.98 & 29.39\\
& & \cellcolor{gray!15}VAGS (Ours) & \cellcolor{gray!15}\textbf{9.32} & \cellcolor{gray!15}\textbf{23.28} & \cellcolor{gray!15}\underline{102.58} & \cellcolor{gray!15}\textbf{53.57} & \cellcolor{gray!15}\underline{84.01} & \cellcolor{gray!15}\underline{84.08} & \cellcolor{gray!15}29.61 & \cellcolor{gray!15}-\\
  % & 22.19 & 22.23 & 117.55 & 69.94 & 26.64 & 83.20 & 26.70\\
  % & & VAGS-Ratio (Ours) & 9.43 & 23.17 & 96.62 & 54.62 & 84.34 & 84.37 & 29.46 & -\\
  % & 23.63 & 21.01 & 131.95 & 90.77 & 26.62 & 81.12 & 26.74\\
\bottomrule
\end{tabular}%
}
\caption{Ablation of guidance scheduling strategies on FlowEdit and SplitFlow. VAGS gives the strongest overall preservation and edit-quality trade-off, improving substantially over constant CFG and prior adaptive schedules.}
\label{tab:guidance_ablation}
\vspace{-3ex}
\end{table*}

\noindent\textbf{Scheduler ablation.}
Tab.~\ref{tab:guidance_ablation} isolates the role of velocity-aware
adaptivity. Interval scheduling gives small gains, Monotone improves
preservation but underperforms VAGS, and Zero-Init collapses structure. VAGS
is the only scheduler that consistently improves both preservation and edit
quality across FlowEdit and SplitFlow. Matched-mean controls in
Secs.~\ref{sec:appendix_editing_diagnostics} and
\ref{sec:appendix_generation_diagnostics} further show that VAGS is not merely
a lower or higher average CFG scale: editing with fixed CFG \(14.05\) remains
near the FlowEdit baseline (Dist \(30.55\) vs.\ \(30.31\)), while VAGS reaches
\(13.84\); generation with fixed CFG \(5.955\) only improves FID from 28.46 to
28.28, far short of VAGS-Gen at 26.07.

\FloatBarrier
\begin{figure*}[t]
  \centering
  \includegraphics[width=1\linewidth]{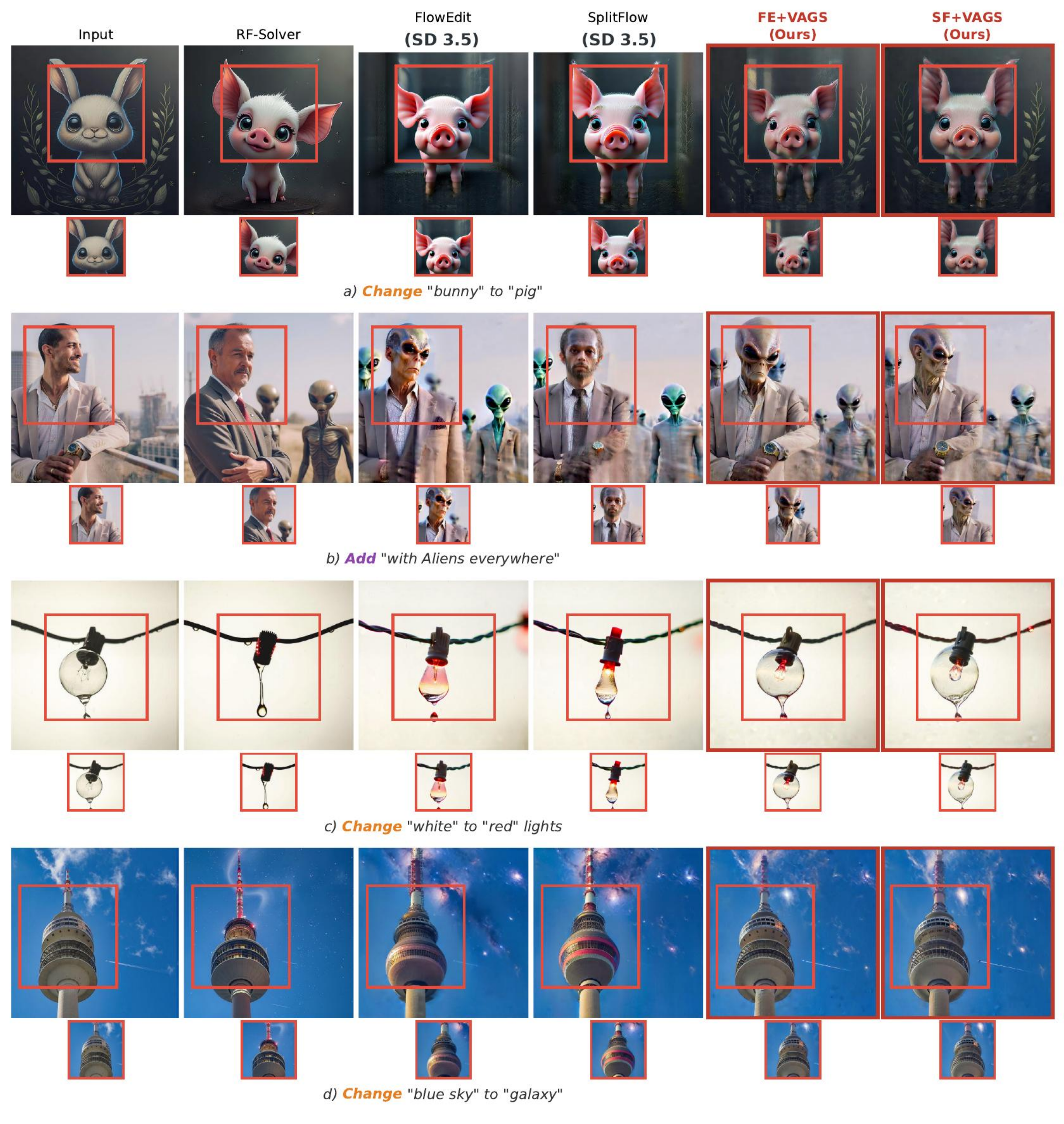}
  \caption{Qualitative editing comparison on PIE-Bench. Columns show,
  from left to right: Input, RF-Solver, FlowEdit (SD~3.5), SplitFlow
  (SD~3.5), FE+VAGS (Ours), and SF+VAGS (Ours). Fixed-guidance editors
  often corrupt backgrounds or under-edit the target region, while VAGS
  produces sharper semantic changes with stronger preservation of
  unedited content.}
  \label{fig:qualitative}
\end{figure*}

CLIP-scaled tables in Sec.~\ref{sec:appendix_clipscaled} preserve the same
conclusions after penalizing methods whose preservation or fidelity gains come
with weaker text-image alignment.

% \subsection{Qualitative Results}
% \label{sec:qual}

\noindent\textbf{Qualitative results on editing.}
Fig.~\ref{fig:qualitative} shows representative PIE-Bench and DIV2K edits.
Fixed guidance often under-edits or corrupts backgrounds, while time-only
schedules still drift on fine texture. VAGS produces sharper target changes
with stronger source preservation, matching the quantitative trends.
Per-pair editing diagnostics in Sec.~\ref{sec:appendix_editing_diagnostics} confirm that the guidance amplification correlates with the semantic regions undergoing the edit.

\noindent\textbf{Qualitative results on generation.}
Qualitative generation examples are provided in
Fig.~\ref{fig:qualitative_generation_main} and
Sec.~\ref{sec:appendix_generation_diagnostics}. VAGS-Gen better preserves
object counts, fine-grained species cues, and compositional layout than fixed
CFG and the training-free guidance baselines. Progressive denoising
comparisons further show that VAGS-Gen follows a different trajectory from
static CFG, even when the static scale is matched to VAGS' mean.

\FloatBarrier
\begin{figure}[t]
    \centering
    \includegraphics[width=\linewidth]{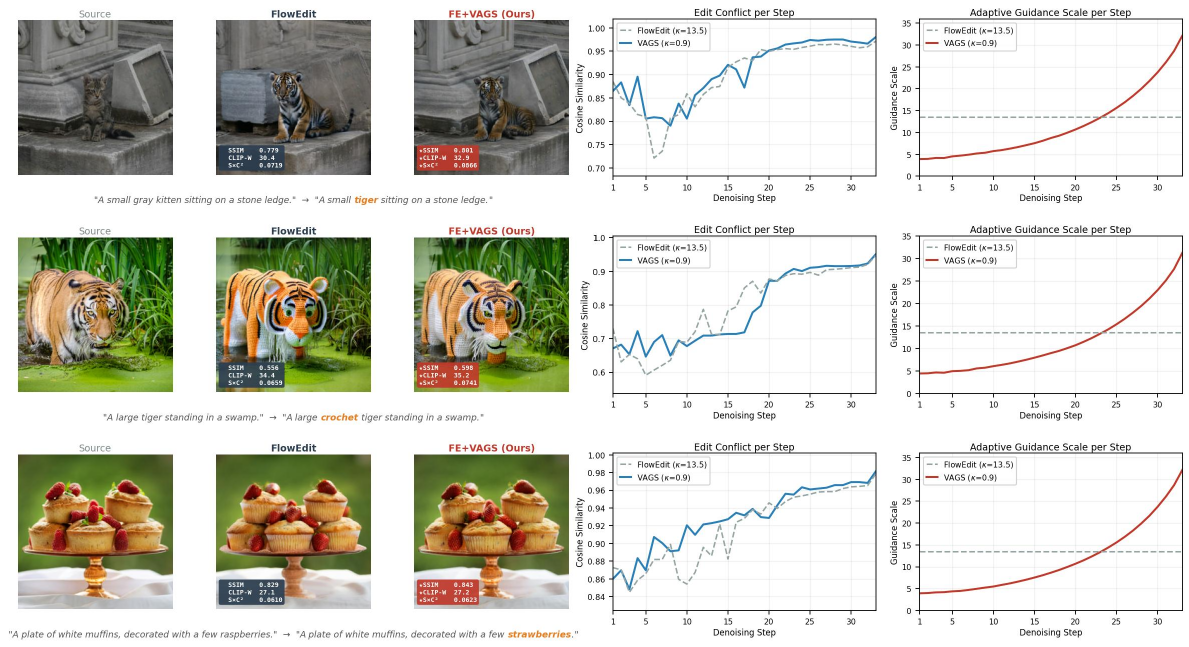}
    \caption{Editing trajectory analysis: per-step cosine similarity and adaptive guidance scale alongside source, baseline, and VAGS edited images.}
    \label{fig:trace_edit_main}
\end{figure}

% \subsection{Ablation Study}
% \label{sec:ablation}

\noindent\textbf{Analysis of modulation strength.}
Generation and editing \(\kappa\) ablations are reported in
Tabs.~\ref{tab:ablation_kappa_generation} and
\ref{tab:ablation_kappa_editing}. Moderate modulation improves FID and
editing preservation, while overly large values suppress or amplify guidance
too aggressively. We therefore use \(\kappa=1.0\) for COCO17/CUB-200,
\(\kappa=0.9\) for Flickr30K, and \(\kappa_\mathrm{tar}=0.9\) for editing.
Full plots are in Sec.~\ref{sec:appendix_kappa}. The chosen values prioritize
balanced quality: larger values can improve isolated fidelity metrics, but
caption or preservation metrics degrade under stronger modulation.

\noindent\textbf{Time complexity \& computational cost.}
The dominant cost of every flow-based sampler considered here is the per-step
forward pass through the velocity network \(v_\theta\), whose cost we denote
\(T_{\mathrm{net}}\); all other operations are \(\mathcal{O}(d)\) in the
latent dimension \(d\) and negligible by comparison. For editing, the
inversion-free baseline performs a single batched forward pass of size four
per step, covering unconditional and conditional predictions for both source
and target prompts. VAGS reuses these raw predictions to assemble the pilot
target velocity, compute one cosine similarity, and re-weight the target
velocity in closed form. The total complexity therefore remains
\(\mathcal{O}(N\,T_{\mathrm{net}})\) with no extra network evaluations.

For generation, constant CFG performs one batched forward pass of size two per
step. VAGS-Gen computes \(s^{\mathrm{gen}}_i\) directly from the same two raw
predictions, again adding only one inner product per step. In measured runtime
on a single NVIDIA H100, editing takes \(7.5\,\mathrm{s}\) per image for the
baseline and \(7.6\,\mathrm{s}\) for VAGS, a \(1.3\%\) overhead. Generation
takes \(11.3\,\mathrm{s}\) for the baseline and \(11.5\,\mathrm{s}\) for
VAGS-Gen, a \(1.8\%\) overhead.

\section{Conclusion}
\label{sec:conclusion}

We presented VAGS, a training-free replacement for fixed CFG in flow-based
editing and generation. Rather than holding guidance constant, VAGS combines a
temporal signal-level term with local velocity alignment, amplifying guidance
where the sampler's dynamics support it and restraining it where stronger
extrapolation would work against committed structure. The method is plug-and-play,
adds no forward passes, and delivers consistent gains in editing fidelity and
generation quality across five benchmarks. The broader implication is clear:
guidance in flow models is not a global scalar to be chosen before sampling
begins. It is a local control problem over velocities, one that can and should
be solved at every step.
\section*{Limitations and Broader Impact}
\label{sec:limitations_broader_impact}

\paragraph{Limitations.}
VAGS is designed for flow-based samplers that expose unconditional and
conditional velocity predictions. It is training-free and inexpensive, but it
does not correct failures caused by weak text encoders, missing training data,
or prompts outside the model distribution. The method also introduces one
modulation strength \(\kappa\), which is stable in our experiments but may
need light tuning for substantially different backbones, resolutions, or
sampling schedules.

\paragraph{Broader impact.}
VAGS improves controllability for image editing and generation, which can
benefit creative tools, dataset construction, and visual prototyping. The same
capability can also make synthetic or edited images more convincing. We do not
add new data sources or identity-specific mechanisms, and the method inherits
the safety profile and biases of the underlying pretrained model. Responsible
deployment should therefore follow the safeguards used for the base generator,
including provenance tracking, prompt filtering where appropriate, and care
around sensitive edits.

\clearpage

% \begin{ack}
% Use unnumbered first level headings for the acknowledgments. All acknowledgments

% Do {\bf not} include this section in the anonymized submission, only in the final paper. You can use the \texttt{ack} environment provided in the style file to automatically hide this section in the anonymized submission.
% \end{ack}

{
\small
\raggedright

\bibliographystyle{unsrtnat}   % or unsrtnat, abbrvnat
\bibliography{main}
}

\clearpage

%%%%%%%%%%%%%%%%%%%%%%%%%%%%%%%%%%%%%%%%%%%%%%%%%%%%%%%%%%%%

\appendix
\section{Technical appendices and supplementary material}
\label{sec:appendix}

This appendix collects implementation pseudocode, diagnostic evidence, and
secondary quantitative tables. The goal is to support the main claim without
interrupting the paper flow: VAGS works because it changes the shape of the
guidance trajectory in response to velocity alignment, not because it simply
changes the average CFG value.

\paragraph{Appendix contents.}
\begin{itemize}
    \item Sec.~\ref{sec:appendix_two_axis}: adaptive scale with two axes.
    \item Sec.~\ref{sec:appendix_algorithms}: editing and generation pseudocode.
    \item Sec.~\ref{sec:appendix_editing_diagnostics}: editing traces,
    alignment diagnostics, and matched-mean controls.
    \item Sec.~\ref{sec:appendix_generation_diagnostics}: generation examples,
    trajectories, and matched-mean controls.
    \item Sec.~\ref{sec:appendix_kappa}: sensitivity to modulation strength.
    \item Sec.~\ref{sec:appendix_clipscaled}: CLIP-scaled metric tables.
\end{itemize}

\clearpage
\subsection{Adaptive scale with two axes.}
\label{sec:appendix_two_axis}

VAGS resolves both issues by replacing the fixed target scale \(\lambda\)
with a step-dependent scale \(\lambda_i\) that is jointly modulated by a
\emph{temporal} factor and a \emph{geometric} factor, both computed from
quantities the sampler already produces. Let
\(\sigma_i=1-t_i\in[0,1]\) be the \emph{signal level} at step \(i\), which
grows monotonically from \(0\) (pure noise) to \(1\) (clean image), and
let \(s_i\in[-1,1]\) be a \emph{velocity-alignment} signal, defined as
the cosine similarity between two task-specific velocity fields available
at step \(i\) (made concrete in Sec.~\ref{sec:method}). VAGS sets
\begin{equation}
    \lambda_i
    \;=\;
    \lambda\cdot
    \exp\!\Bigl(\,\kappa\,(2\sigma_i-1)\,s_i\,\Bigr),
\end{equation}
where \(\kappa\!\geq\!0\) controls the modulation strength. The two
factors play complementary roles: the temporal factor encodes
\emph{which signal to trust} at the current stage of sampling, and the
geometric factor encodes \emph{whether the local velocity geometry
corroborates that trust enough to warrant a correction}
(Fig.~\ref{fig:schematic}).

\begin{itemize}
    \item \textbf{Temporal axis: \((2\sigma_i-1)\).} Negative in the noisy
    regime (\(\sigma_i\!<\!\tfrac12\)) and positive in the clean regime
    (\(\sigma_i\!>\!\tfrac12\)), this signed factor passes through zero at
    \(\sigma_i\!=\!\tfrac12\), where \(\lambda_i\) recovers the nominal
    scale exactly. Its sign reflects a qualitative shift in the sampler's
    task across the trajectory. Early steps perform \emph{basin
    selection}: from a near-isotropic latent the sampler must commit to a
    region of the data manifold, and the unconditional prior is the more
    reliable signal because raw conditional velocities are dominated by
    noise. Late steps perform \emph{mode refinement}: coarse content is
    fixed, and the prompt-conditioned signal must sharpen semantic detail
    without disturbing committed structure. The monotone ramp is
    consistent with prior empirical findings that increasing CFG
    schedules outperform constant
    guidance~\citep{chang2023muse,wang2024analysis}, but VAGS treats it
    not as a magnitude schedule alone, rather, as a \emph{regime
    indicator} that flips how the geometric axis should be read.

    \item \textbf{Geometric axis: \(s_i\).} The cosine similarity between
    the two velocity fields is invariant to the magnitude fluctuations of
    the raw predictions across timesteps and prompts, isolating their
    angular relationship. Critically, the \emph{informative event}
    differs between regimes. In the clean regime, where raw velocities
    are high-fidelity, agreement (\(s_i\!>\!0\)) means the
    prompt-conditioned direction is consistent with established
    structure: amplifying guidance sharpens the desired direction
    without disturbing unrelated content. Disagreement
    (\(s_i\!<\!0\)) means guidance is fighting committed structure and
    should be damped to avoid overrides or off-manifold drift. The
    polarity inverts in the noisy regime, where \(s_i\) is itself a
    low-fidelity quantity computed from two noisy vectors. The temporal
    factor \(|2\sigma_i-1|\) is correspondingly small there, so the
    multiplicative correction stays close to one regardless of how
    \(s_i\) is read; within that mild envelope, agreement indicates the
    prompt is broadly redundant with the unconditional prior (high
    \(\lambda\) buys little adherence at the cost of amplifying noise,
    motivating slight attenuation), while systematic disagreement is the
    most plausible early evidence that the prompt targets a region of
    the manifold the unconditional default would not naturally select,
    warranting a small early commitment toward that region.
\end{itemize}

The product \((2\sigma_i-1)\,s_i\) is the indicator that implements this
regime-dependent reading of \(s_i\). Its sign across the four
\((\sigma_i,s_i)\) quadrants is summarised in Table~\ref{tab:two_axes},
together with the sampler's local objective.

\begin{table}[h]
\centering
\renewcommand{\arraystretch}{1.2}
\scriptsize
\caption{Behaviour of the multiplicative correction across regimes.
Disagreement carries different information depending on which signal
the sampler should trust at the current stage; the temporal factor
flips the polarity, and its small magnitude in the noisy regime keeps
the correction mild precisely where \(s_i\) is least reliable.}
\label{tab:two_axes}
\begin{tabular}{@{}lll@{}}
\toprule
\textbf{Regime} & \textbf{Role of disagreement} (\(s_i\!<\!0\)) & \textbf{Effect on \(\lambda_i\)} \\
\midrule
Noisy (\(\sigma_i\!<\!\tfrac12\)): basin selection & Plausible signal of required basin shift & Mild amplification; agreement mildly attenuates \\
Clean (\(\sigma_i\!>\!\tfrac12\)): mode refinement & Conflict with committed structure & Attenuation; agreement amplifies \\
\bottomrule
\end{tabular}
\end{table}

This design is well suited to the training-free, plug-and-play setting
in which VAGS operates. The properties it relies on---that the
unconditional prior dominates trajectory shape under high noise, that
prompt-conditioned signals dominate under low noise, and that high CFG
late in sampling produces off-manifold artefacts when guidance fights
structure---are documented behaviours of pretrained flow
models~\citep{chang2023muse,wang2024analysis,fan2025cfg,saini2025rectified},
not behaviours VAGS induces. The temporal axis selects which of these
documented regimes the sampler is in; the geometric axis reads the
local velocity geometry to gate the correction. Crucially, in the
noisy regime where \(s_i\) is least reliable, \((2\sigma_i-1)\) is
small in magnitude, so the multiplier stays close to one and the cost
of misreading \(s_i\) is bounded by construction. In the clean regime
where the geometric reading is most trustworthy, \((2\sigma_i-1)\)
approaches its maximum, allowing the correction to take full effect
exactly where it is most reliable. The training-free nature of VAGS is
therefore not a limitation that the design tolerates; it is what the
design is calibrated for.

Two structural properties of Eq.~\eqref{eq:adaptive} are worth noting.
First, because \(s_i\) is bounded in \([-1,1]\), the multiplier is
bounded in \([\,e^{-\kappa},e^{\kappa}\,]\), so a single \(\kappa\)
controls the entire range of admissible scales and the final
\(\lambda_i\) is not guaranteed to be monotone in \(i\)---adaptivity is
traded for strict monotonicity by design. Second, setting
\(\kappa\!=\!0\) yields \(\lambda_i\!\equiv\!\lambda\), recovering the
fixed-CFG baseline exactly; VAGS is therefore a strict generalisation.
The remaining task is to specify the velocity pair from which \(s_i\)
is computed in each setting.

\subsection{Pseudocode}
\suppressfloats[t]
\label{sec:appendix_algorithms}

Algorithms~\ref{alg:vags} and~\ref{alg:vag_gen} expand the compact
description in Sec.~\ref{sec:method}. Both procedures reuse the raw velocity
predictions required by the underlying CFG sampler; VAGS only adds cosine
similarity and scalar re-weighting.

\begin{algorithm}[!htbp]
\DontPrintSemicolon
\SetAlgoLined
\SetNlSty{textbf}{}{}
\SetKwInput{KwNotation}{Notation}
\caption{Velocity-Adaptive Guidance Scale for Editing (VAGS)}
\label{alg:vags}

\KwIn{source image $x_\mathrm{src}$;\ prompts $c_\mathrm{src},c_\mathrm{tar}$;\ 
      base scales $\lambda_\mathrm{src},\lambda_\mathrm{tar}$;\ 
      modulation strength $\kappa_\mathrm{tar}$;\ 
      schedule $\{t_i\}_{i=1}^{N}$ with $0\!\approx\! t_1\!<\!\cdots\!<\! t_N\!\approx\! 1$.}
\KwOut{edited image $z^{\mathrm{edit}}_{t_1}$.}
\KwNotation{$\hat{z}^{\mathrm{s}}_i\!\equiv\!\hat{z}^{\,\mathrm{src}}_{t_i}$,\ 
            $\hat{z}^{\mathrm{t}}_i\!\equiv\!\hat{z}^{\,\mathrm{tar}}_{t_i}$;\ 
            $V^{\mathrm{s}}_i,\ \widetilde{V}^{\mathrm{t}}_i,\ V^{\mathrm{t}}_i$
            denote source, pilot-target, and adaptive-target guided velocities.}

\BlankLine
$z^{\mathrm{edit}}_{t_N} \leftarrow x_\mathrm{src}$ \tcp*[r]{initialise editing trajectory}
\For{$i = N, N{-}1, \ldots, 2$}{

  \tcc*[l]{\textsc{1.\ Couple noisy source/target latents}\hfill\eqref{eq:edit_couple}}
  $\epsilon_i \sim \mathcal{N}(0, I)$\;
  $\hat{z}^{\mathrm{s}}_i \leftarrow (1 - t_i)\, x_\mathrm{src} + t_i\, \epsilon_i$\;
  $\hat{z}^{\mathrm{t}}_i \leftarrow z^{\mathrm{edit}}_{t_i} + \hat{z}^{\mathrm{s}}_i - x_\mathrm{src}$\;

  \BlankLine
  \tcc*[l]{\textsc{2.\ One forward pass on the four-way batch}}
  $\bigl(u^{\mathrm{src}}_i,\, p^{\mathrm{src}}_i,\, u^{\mathrm{tar}}_i,\, p^{\mathrm{tar}}_i\bigr) \leftarrow
   v_\theta\bigl([\hat{z}^{\mathrm{s}}_i,\hat{z}^{\mathrm{s}}_i,\hat{z}^{\mathrm{t}}_i,\hat{z}^{\mathrm{t}}_i],\,
                 t_i,\,
                 [\varnothing, c_\mathrm{src}, \varnothing, c_\mathrm{tar}]\bigr)$\;

  \BlankLine
  \tcc*[l]{\textsc{3.\ Source velocity and pilot target velocity}\hfill\eqref{eq:pilot}}
  $\sigma_i \leftarrow 1 - t_i$\;
  $\lambda^{\mathrm{tar},\mathrm{base}}_i \leftarrow \lambda_\mathrm{tar}\,\exp\!\bigl(\kappa_\mathrm{tar}(2\sigma_i - 1)\bigr)$\;
  $V^{\mathrm{s}}_i \leftarrow u^{\mathrm{src}}_i + \lambda_\mathrm{src}\,(p^{\mathrm{src}}_i - u^{\mathrm{src}}_i)$\;
  $\widetilde{V}^{\mathrm{t}}_i \leftarrow u^{\mathrm{tar}}_i + \lambda^{\mathrm{tar},\mathrm{base}}_i\,(p^{\mathrm{tar}}_i - u^{\mathrm{tar}}_i)$\;

  \BlankLine
  \tcc*[l]{\textsc{4.\ Alignment signal and adaptive target scale}\hfill\eqref{eq:edit_adaptive}}
  $s^{\mathrm{edit}}_i \leftarrow
     \langle V^{\mathrm{s}}_i,\, \widetilde{V}^{\mathrm{t}}_i\rangle\,/\,
     \bigl(\|V^{\mathrm{s}}_i\|\,\|\widetilde{V}^{\mathrm{t}}_i\|\bigr)$\;
  $\lambda^{\mathrm{tar}}_i \leftarrow \lambda_\mathrm{tar}\,\exp\!\bigl(\kappa_\mathrm{tar}(2\sigma_i - 1)\, s^{\mathrm{edit}}_i\bigr)$\;

  \BlankLine
  \tcc*[l]{\textsc{5.\ Re-weight target velocity and advance the ODE}\hfill\eqref{eq:edit_step}}
  $V^{\mathrm{t}}_i \leftarrow u^{\mathrm{tar}}_i + \lambda^{\mathrm{tar}}_i\,(p^{\mathrm{tar}}_i - u^{\mathrm{tar}}_i)$\;
  $z^{\mathrm{edit}}_{t_{i-1}} \leftarrow z^{\mathrm{edit}}_{t_i} + (t_{i-1} - t_i)\bigl(V^{\mathrm{t}}_i - V^{\mathrm{s}}_i\bigr)$\;
}
\Return $z^{\mathrm{edit}}_{t_1}$\;
\end{algorithm}

\begin{algorithm}[!htbp]
\DontPrintSemicolon
\SetAlgoLined
\SetNlSty{textbf}{}{}
\SetKwInput{KwNotation}{Notation}
\caption{Velocity-Adaptive Guidance Scale for Generation (VAGS-Gen)}
\label{alg:vag_gen}

\KwIn{prompt $c$;\ base guidance scale $\lambda$;\ 
      modulation strength $\kappa$;\ 
      schedule $\{t_i\}_{i=1}^{N}$ with $0\!\approx\! t_1\!<\!\cdots\!<\! t_N\!\approx\! 1$.}
\KwOut{generated image $z_{t_1}$.}
\KwNotation{$v^{(i)}_\varnothing$ and $v^{(i)}_c$ denote the unconditional and
            conditional raw velocity predictions at $z_{t_i}$;\ 
            $V_i$ is the CFG-guided velocity used to advance the ODE.}

\BlankLine
$z_{t_N} \sim \mathcal{N}(0, I)$ \tcp*[r]{initialise generation trajectory}
\For{$i = N, N{-}1, \ldots, 2$}{

  \tcc*[l]{\textsc{1.\ One forward pass on the two-way batch}}
  $\bigl(v^{(i)}_\varnothing,\, v^{(i)}_c\bigr) \leftarrow
   v_\theta\bigl([z_{t_i}, z_{t_i}],\, t_i,\, [\varnothing,\, c]\bigr)$\;

  \BlankLine
  \tcc*[l]{\textsc{2.\ Alignment signal}\hfill\eqref{eq:gen_adaptive}}
  $\sigma_i \leftarrow 1 - t_i$\;
  $s^{\mathrm{gen}}_i \leftarrow
     \langle v^{(i)}_\varnothing,\, v^{(i)}_c\rangle\,/\,
     \bigl(\|v^{(i)}_\varnothing\|\,\|v^{(i)}_c\|\bigr)$\;

  \BlankLine
  \tcc*[l]{\textsc{3.\ Adaptive guidance scale}\hfill\eqref{eq:gen_adaptive}}
  $\lambda_i \leftarrow \lambda\,\exp\!\bigl(\kappa(2\sigma_i - 1)\, s^{\mathrm{gen}}_i\bigr)$\;

  \BlankLine
  \tcc*[l]{\textsc{4.\ Assemble guided velocity and advance the ODE}\hfill\eqref{eq:cfg},\,\eqref{eq:gen_step}}
  $V_i \leftarrow v^{(i)}_\varnothing + \lambda_i\bigl(v^{(i)}_c - v^{(i)}_\varnothing\bigr)$\;
  $z_{t_{i-1}} \leftarrow z_{t_i} + (t_{i-1} - t_i)\, V_i$\;
}
\Return $z_{t_1}$\;
\end{algorithm}

\clearpage
\subsection{Editing diagnostics}
\suppressfloats[t]
\label{sec:appendix_editing_diagnostics}

\begin{figure*}[t]
    \centering
    \begin{subfigure}[t]{0.32\textwidth}
        \centering
        \includegraphics[width=\linewidth]{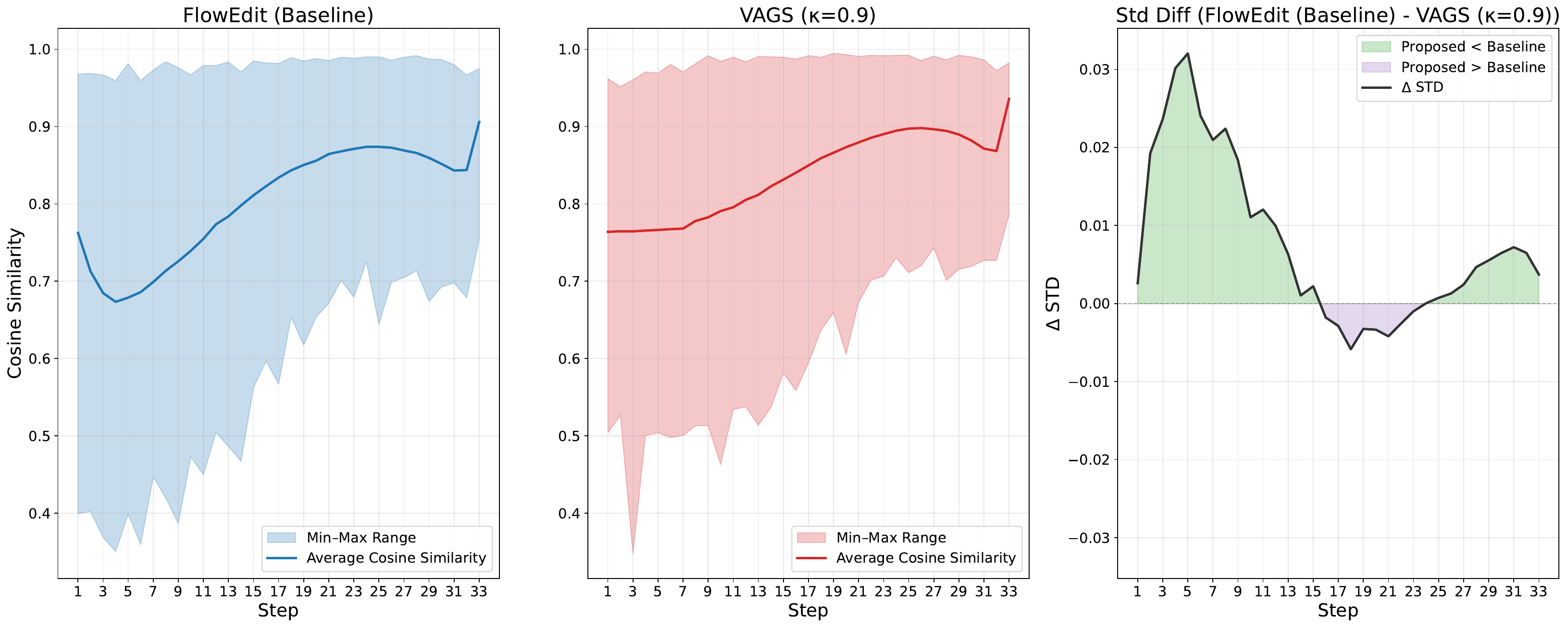}
        \caption{Step-wise alignment}
        \label{fig:cos_sim_step_piebench}
    \end{subfigure}
    \hfill
    \begin{subfigure}[t]{0.32\textwidth}
        \centering
        \includegraphics[width=\linewidth]{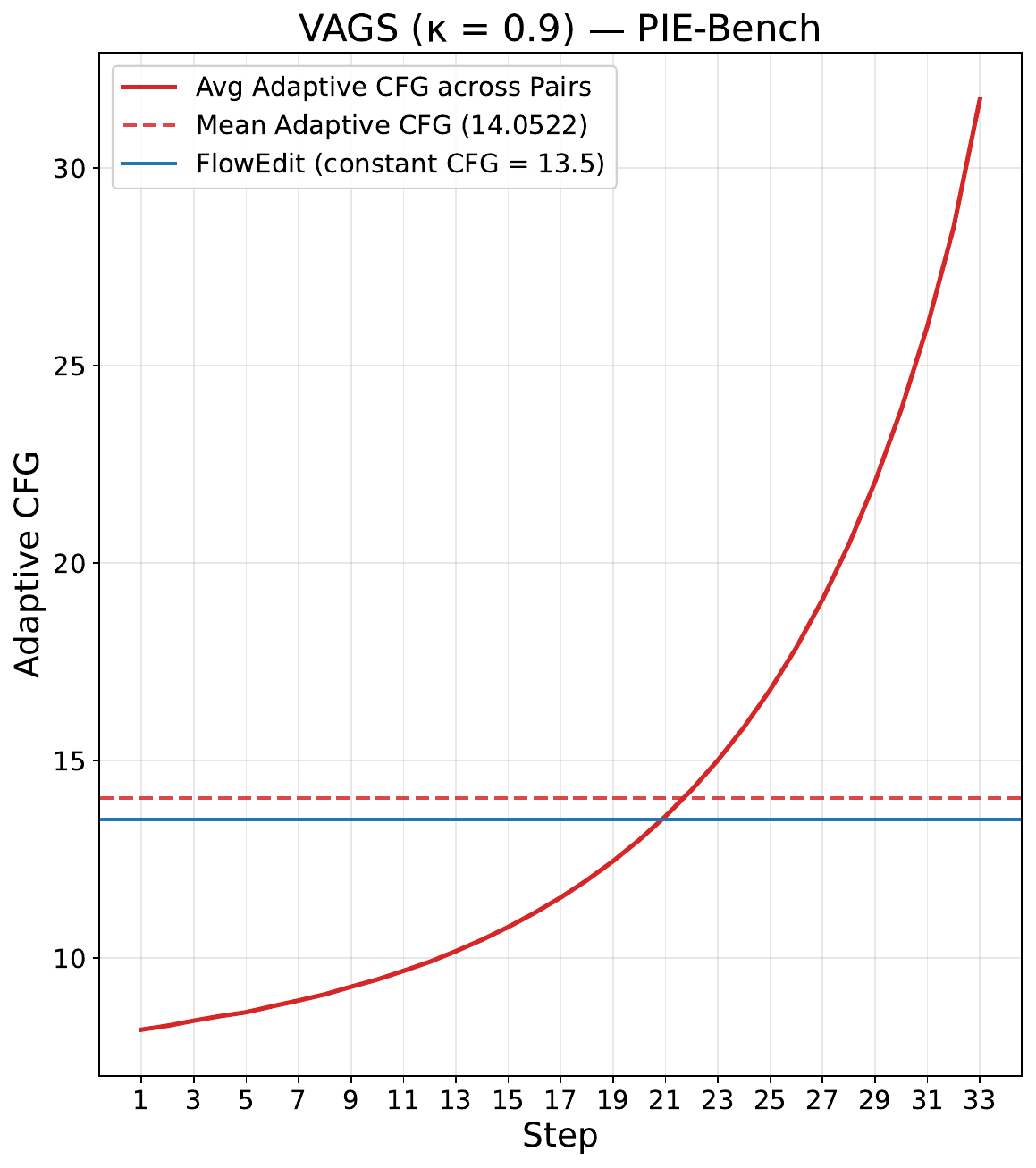}
        \caption{Step-wise target CFG}
        \label{fig:adaptive_cfg_step_edit}
    \end{subfigure}
    \hfill
    \begin{subfigure}[t]{0.32\textwidth}
        \centering
        \includegraphics[width=\linewidth]{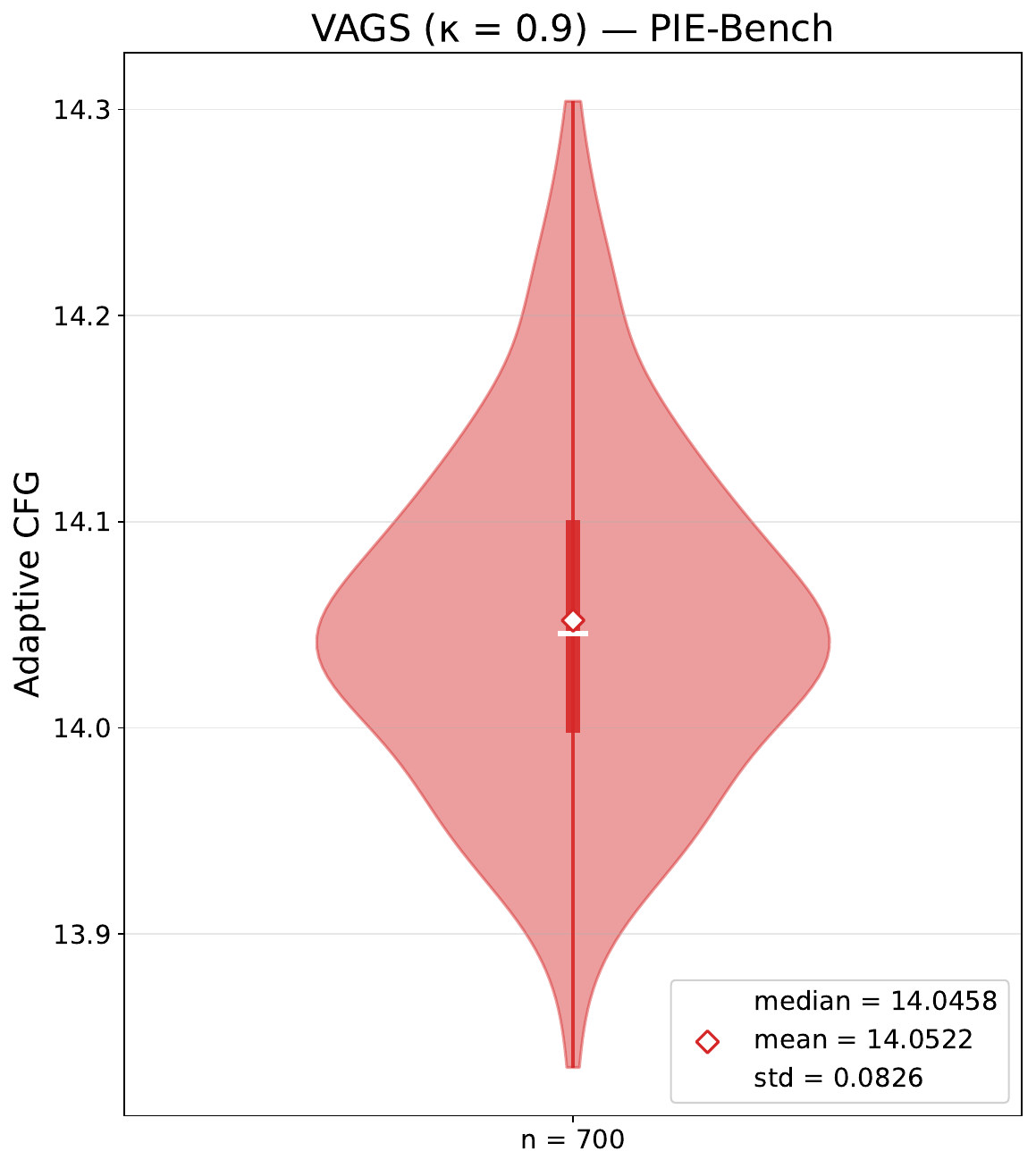}
        \caption{Pair-wise mean CFG}
        \label{fig:adaptive_cfg_prompt_edit}
    \end{subfigure}
    \caption{Editing diagnostics on PIE-Bench. The cosine signal tracks the
    source/target velocity relation, while the resulting VAGS target scale is
    stable across image-prompt pairs and rises toward clean-image steps.}
    \label{fig:editing_alignment_scale_diagnostics}
\end{figure*}

This section checks whether the editing gains are tied to the intended
source/target velocity geometry.
Fig.~\ref{fig:editing_alignment_scale_diagnostics} summarises the alignment
signal and the resulting adaptive target scale across PIE-Bench.
Within this grouped diagnostic, Fig.~\ref{fig:cos_sim_step_piebench} reports
the step-wise cosine signal, Fig.~\ref{fig:adaptive_cfg_step_edit} reports the
step-wise adaptive target scale, and
Fig.~\ref{fig:adaptive_cfg_prompt_edit} reports the pair-wise mean scale.

\FloatBarrier

\paragraph{Matched-mean editing control.}
Tab.~\ref{tab:average_cfg_fix_cfg_edit} compares VAGS with a constant target
scale set to the mean adaptive value \((\bar\lambda=14.05)\). The matched
constant scale stays close to the baseline, while VAGS substantially improves
preservation, confirming that the gain comes from step-wise modulation.

\FloatBarrier
% Required packages: booktabs, multirow, graphicx, xcolor
% Editing counterpart of generation_mean_cfg_ablation.tex.
\begin{table*}[t!]
\centering
\small
\setlength{\tabcolsep}{4.5pt}
\renewcommand{\arraystretch}{1.12}
\resizebox{\textwidth}{!}{%
\begin{tabular}{@{}l l c c c c c c c c@{}}
\toprule
\multirow[c]{2}{*}{\textbf{Dataset}} &
\multirow[c]{2}{*}{\textbf{Method}} &
\multirow[c]{2}{*}{\textbf{Dist.\,$\times10^{3}$} $\downarrow$} &
\multirow[c]{2}{*}{\textbf{PSNR} $\uparrow$} &
\multirow[c]{2}{*}{\textbf{LPIPS\,$\times10^{3}$} $\downarrow$} &
\multirow[c]{2}{*}{\textbf{MSE\,$\times10^{4}$} $\downarrow$} &
\multirow[c]{2}{*}{\textbf{SSIM\,$\times10^{2}$} $\uparrow$} &
\multirow[c]{2}{*}{\textbf{CLIP-I} $\uparrow$} &
\multicolumn{2}{c}{\textbf{CLIP Similarity}} \\
\cmidrule(lr){9-10}
& & & & & & & &
\textbf{Whole} $\uparrow$ &
\textbf{Edited} $\uparrow$ \\
\midrule

\multirow[c]{3}{*}{\textbf{PIE-Bench}}
& FlowEdit (CFG=13.5)
  & \underline{30.31} & \underline{21.87} & \underline{117.06} & \underline{92.70} & \underline{82.54} & \underline{83.69} & \textbf{27.62} & \underline{23.80} \\

& FlowEdit (CFG=14.05, $\kappa{=}0$)
  & 30.55 & 21.83 & 117.40 & 93.10 & 82.50 & 83.65 & \underline{27.60} & \textbf{23.82} \\

& \cellcolor{gray!15}FlowEdit + VAGS (CFG=13.5, $\kappa{=}0.9$)
  & \textbf{13.84} & \textbf{26.38} & \textbf{70.38} & \textbf{34.86} & \textbf{87.68} & \textbf{87.44} & 26.92 & 23.08 \\

\bottomrule
\end{tabular}%
}
\caption{Constant-CFG vs.\ VAGS editing on PIE-Bench. A fixed scale at the mean adaptive
value ($\bar{\lambda}=14.05$) performs on par with the baseline, confirming that the gains
stem from dynamic modulation rather than a shifted average guidance level.}
\label{tab:average_cfg_fix_cfg_edit}
\end{table*}

\clearpage
\subsection{Generation diagnostics}
\suppressfloats[t]
\label{sec:appendix_generation_diagnostics}

This section provides the generation-side analogue of the editing diagnostics.
The evidence is organised into qualitative comparisons, trajectory views, and
aggregate alignment statistics so that each figure group has nearby context.

\begin{figure}[t]
    \centering
    \includegraphics[width=\linewidth]{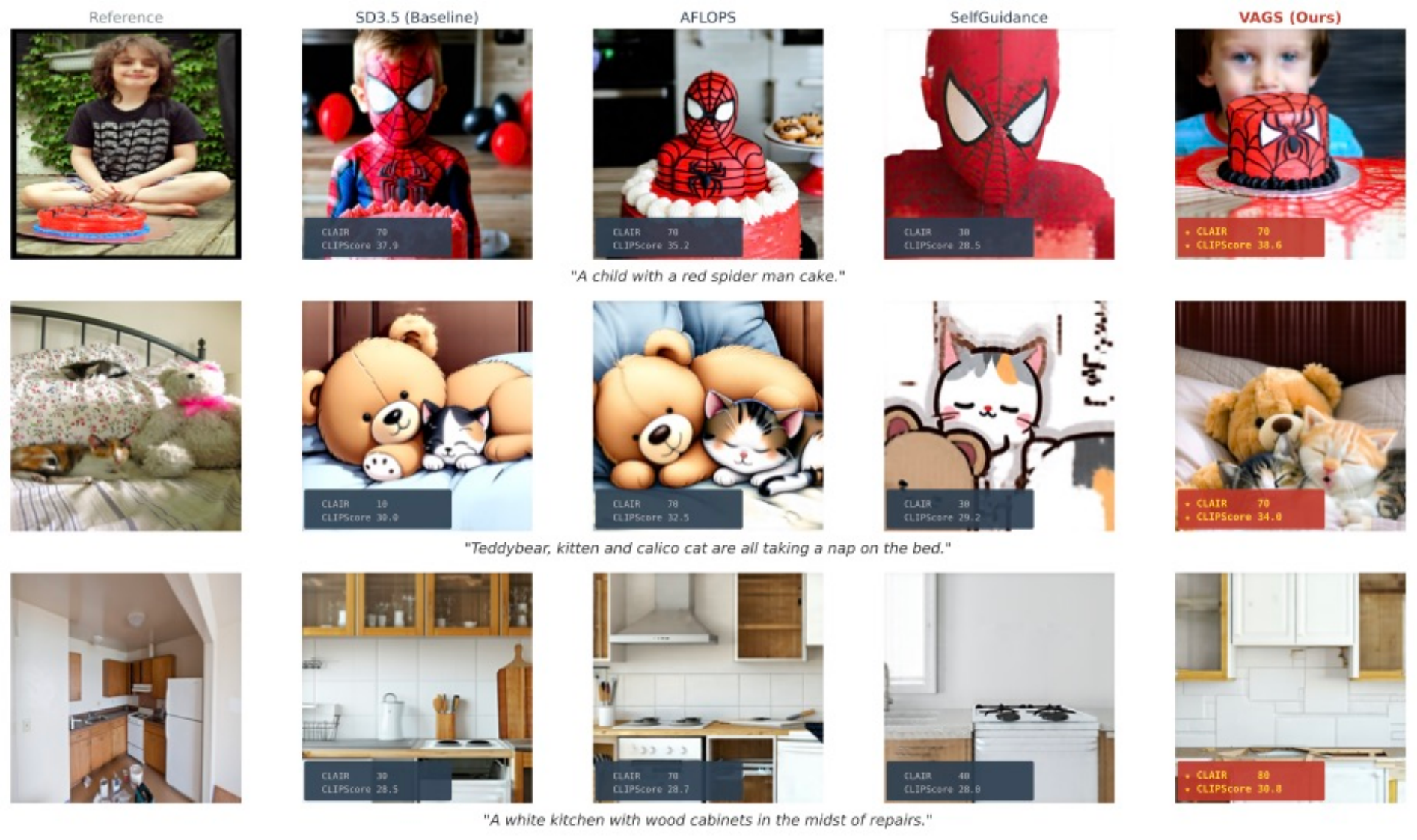}
    \caption{Qualitative generation comparison on COCO17 and CUB-200. VAGS-Gen
    improves prompt alignment and visual fidelity over fixed-CFG generation
    while preserving sample diversity.}
    \label{fig:qualitative_generation_main}
\end{figure}

\begin{figure}[t]
    \centering
    \includegraphics[width=\linewidth]{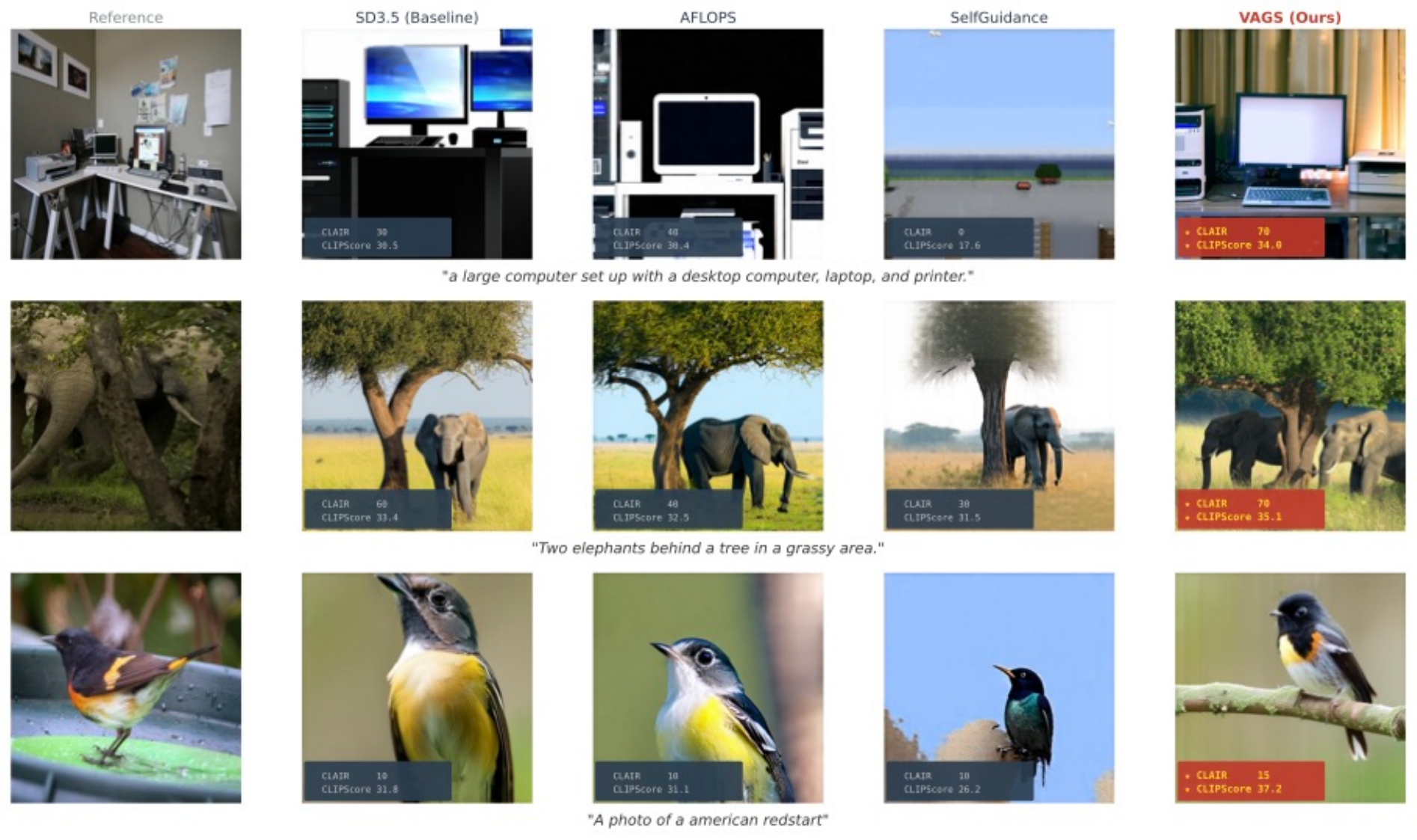}
    \caption{Generation qualitative comparison: additional examples on COCO17 and CUB-200.}
    \label{fig:qualitative_generation_supp}
\end{figure}

\paragraph{Qualitative comparisons.}
Figs.~\ref{fig:qualitative_generation_main} and
\ref{fig:qualitative_generation_supp} show where VAGS-Gen helps with counts,
fine-grained attributes, and composition while preserving sample diversity.

\FloatBarrier

\begin{figure}[t]
    \centering
    \includegraphics[width=\linewidth]{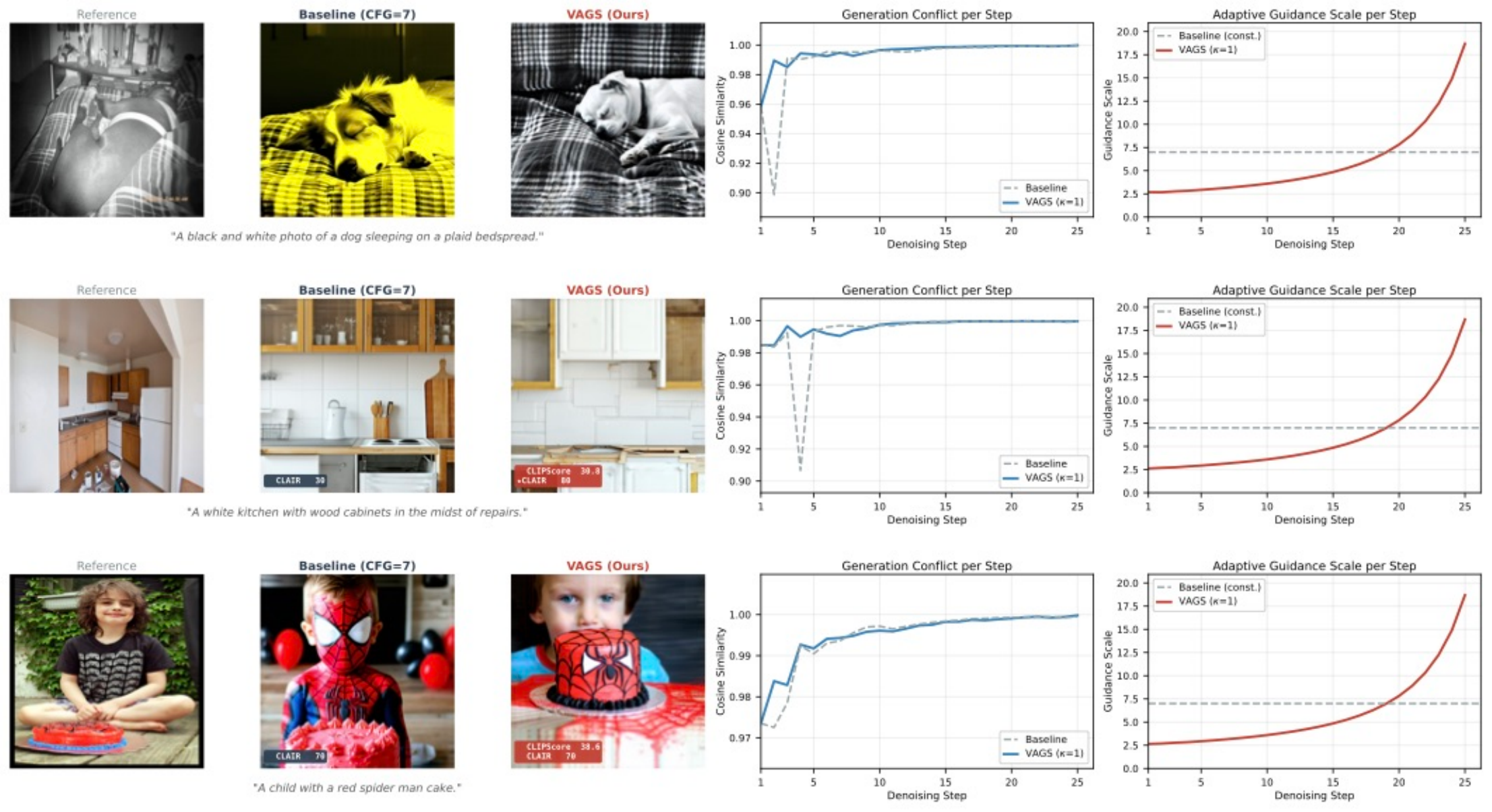}
    \caption{Generation trajectory analysis: per-step cosine similarity and adaptive guidance scale alongside reference, baseline, and VAGS samples.}
    \label{fig:trace_generation_main}
\end{figure}

\begin{figure}[t]
    \centering
    \includegraphics[width=\linewidth]{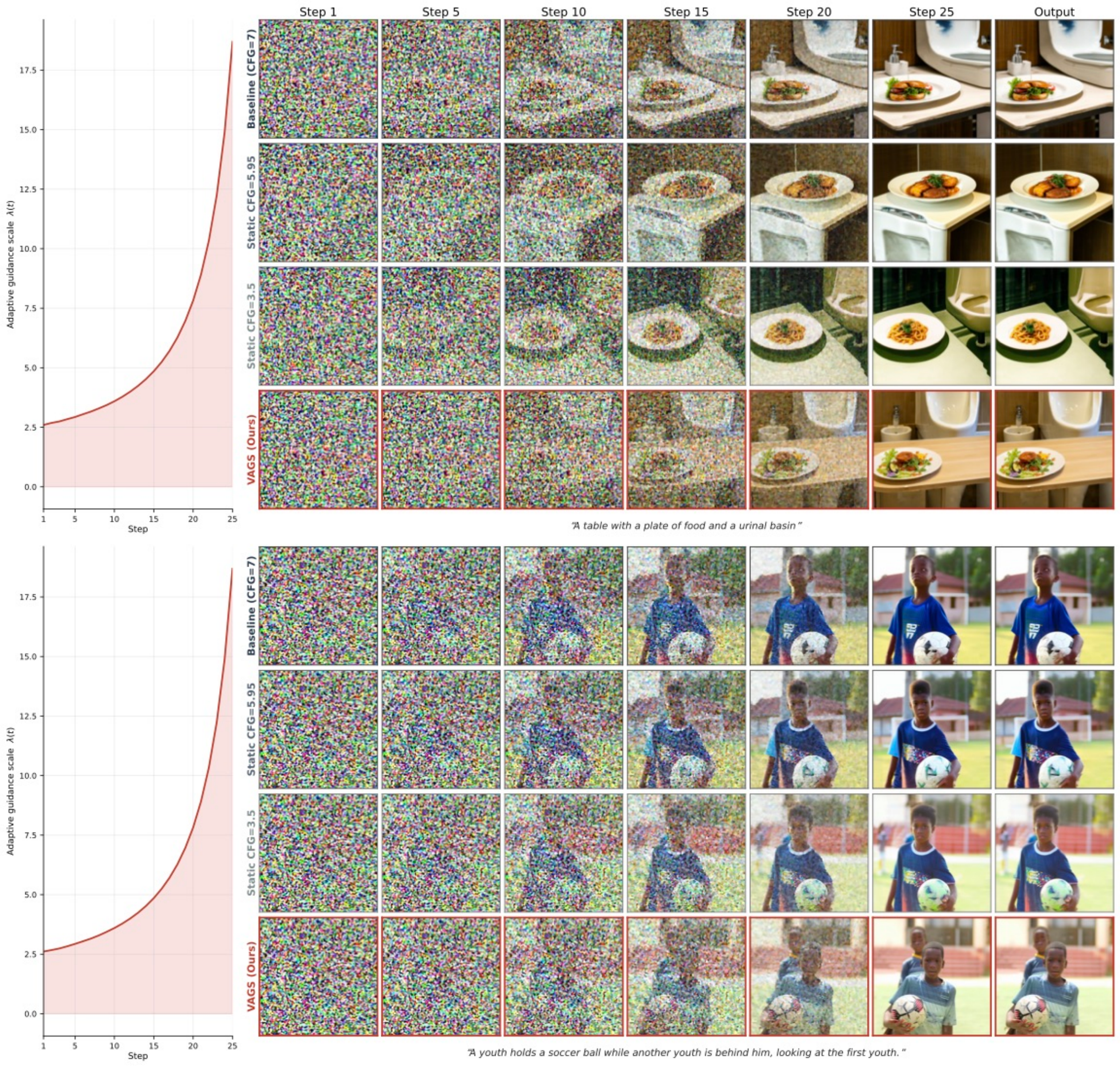}
    \caption{Progressive denoising on COCO17: per-step intermediate images for VAGS ($\kappa{=}1$) versus three static-CFG baselines (CFG=7 default, CFG=3.5 low, CFG=5.95 matched-mean of VAGS adaptive $\lambda$). Leftmost column shows the VAGS adaptive $\lambda(t)$ curve. Even when matched on mean guidance, static CFG denoises along a distinguishable trajectory from VAGS.}
    \label{fig:progressive_generation_main}
\end{figure}

\begin{figure}[t]
    \centering
    \includegraphics[width=\linewidth]{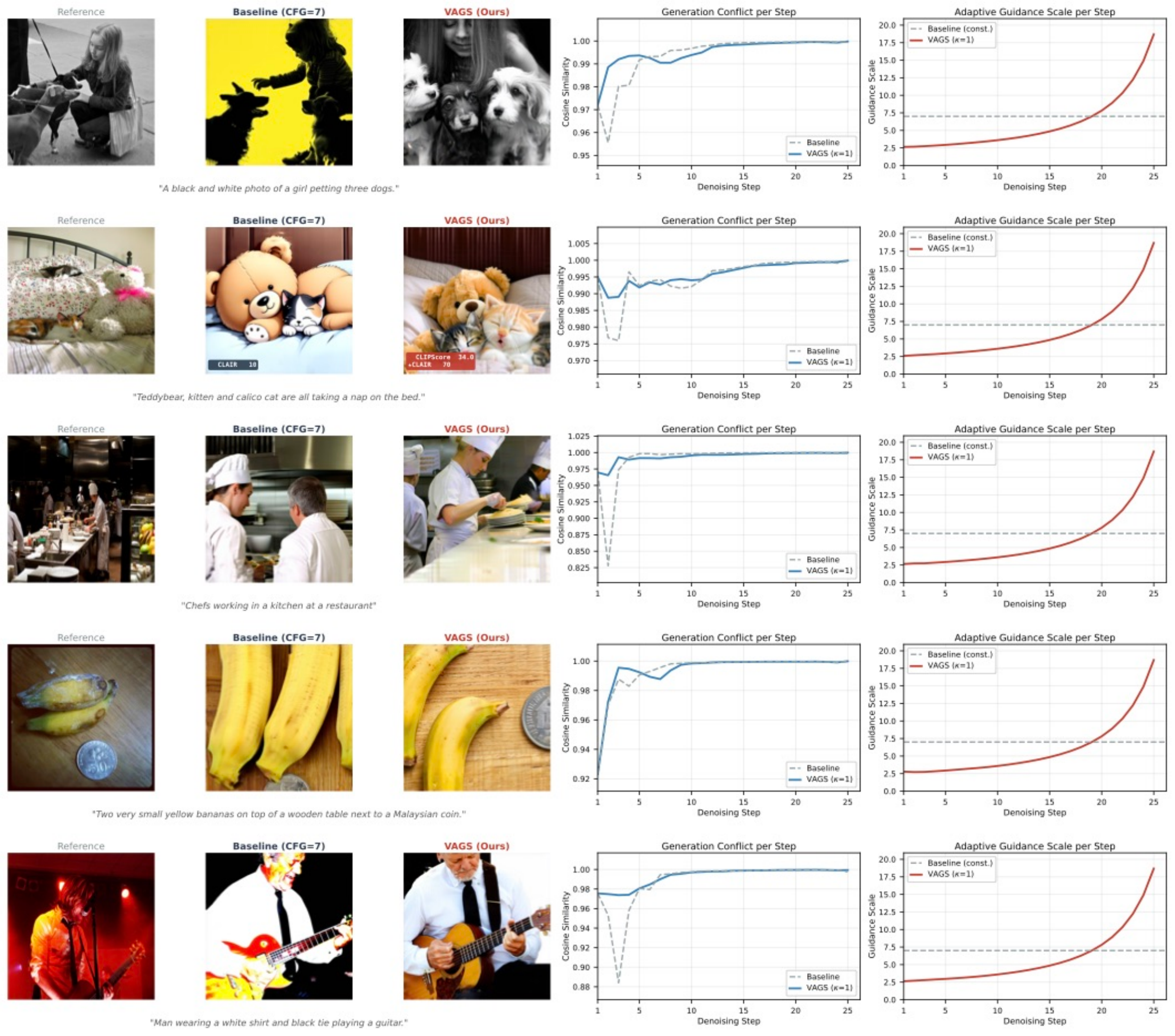}
    \caption{Generation trajectory analysis: additional per-prompt traces.}
    \label{fig:trace_generation_supp}
\end{figure}

\FloatBarrier
\paragraph{Trajectory diagnostics.}
Figs.~\ref{fig:trace_generation_main}, \ref{fig:progressive_generation_main},
and \ref{fig:trace_generation_supp} verify that the qualitative gains
correspond to a reshaped CFG schedule driven by unconditional/conditional
velocity alignment.

\FloatBarrier

\begin{figure*}[t]
    \centering
    \begin{subfigure}[t]{0.48\textwidth}
        \centering
        \includegraphics[width=\linewidth]{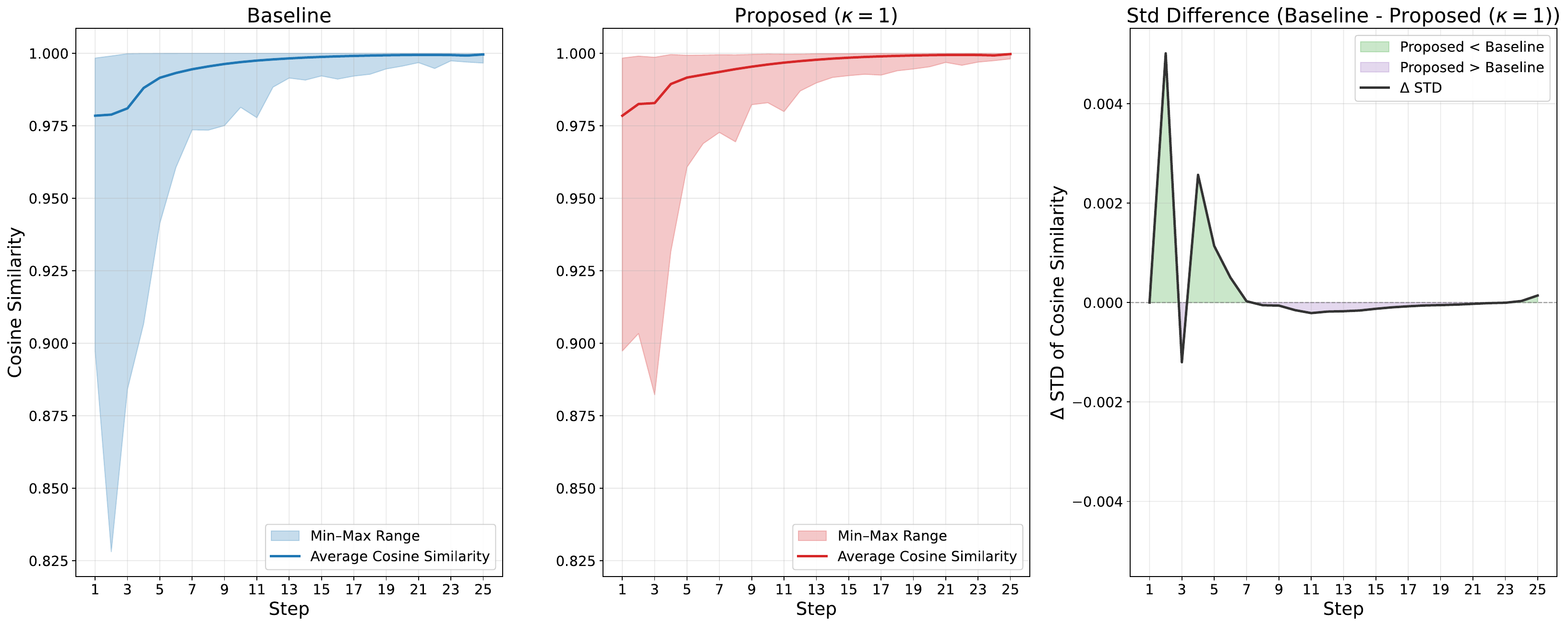}
        \caption{Step-wise velocity alignment}
        \label{fig:cos_sim_step_coco17}
    \end{subfigure}
    \hfill
    \begin{subfigure}[t]{0.48\textwidth}
        \centering
        \includegraphics[width=\linewidth]{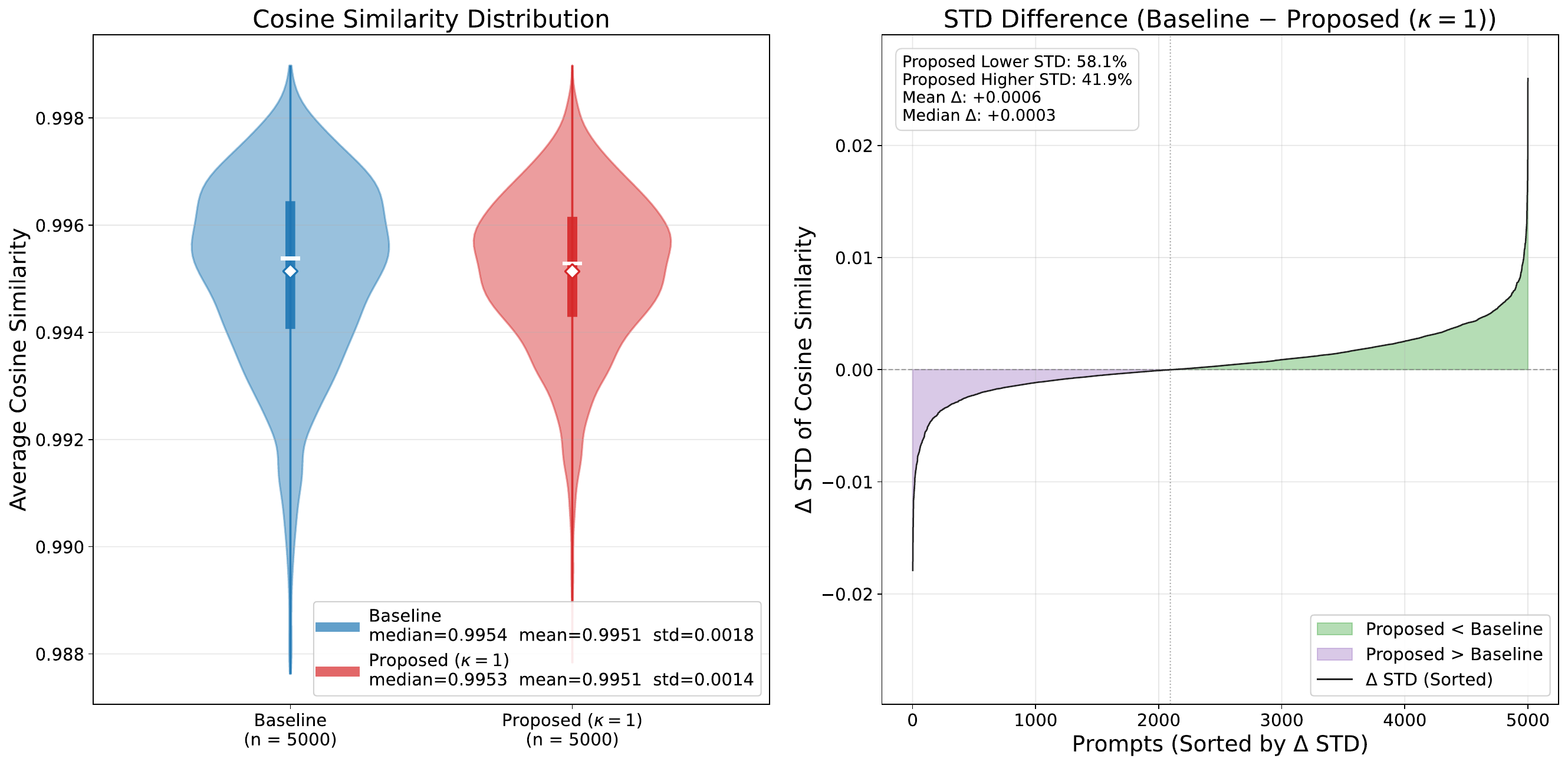}
        \caption{Prompt-wise alignment statistics}
        \label{fig:cos_sim_prompt_coco17}
    \end{subfigure}

    \vspace{0.5em}
    \begin{subfigure}[t]{0.48\textwidth}
        \centering
        \includegraphics[width=\linewidth]{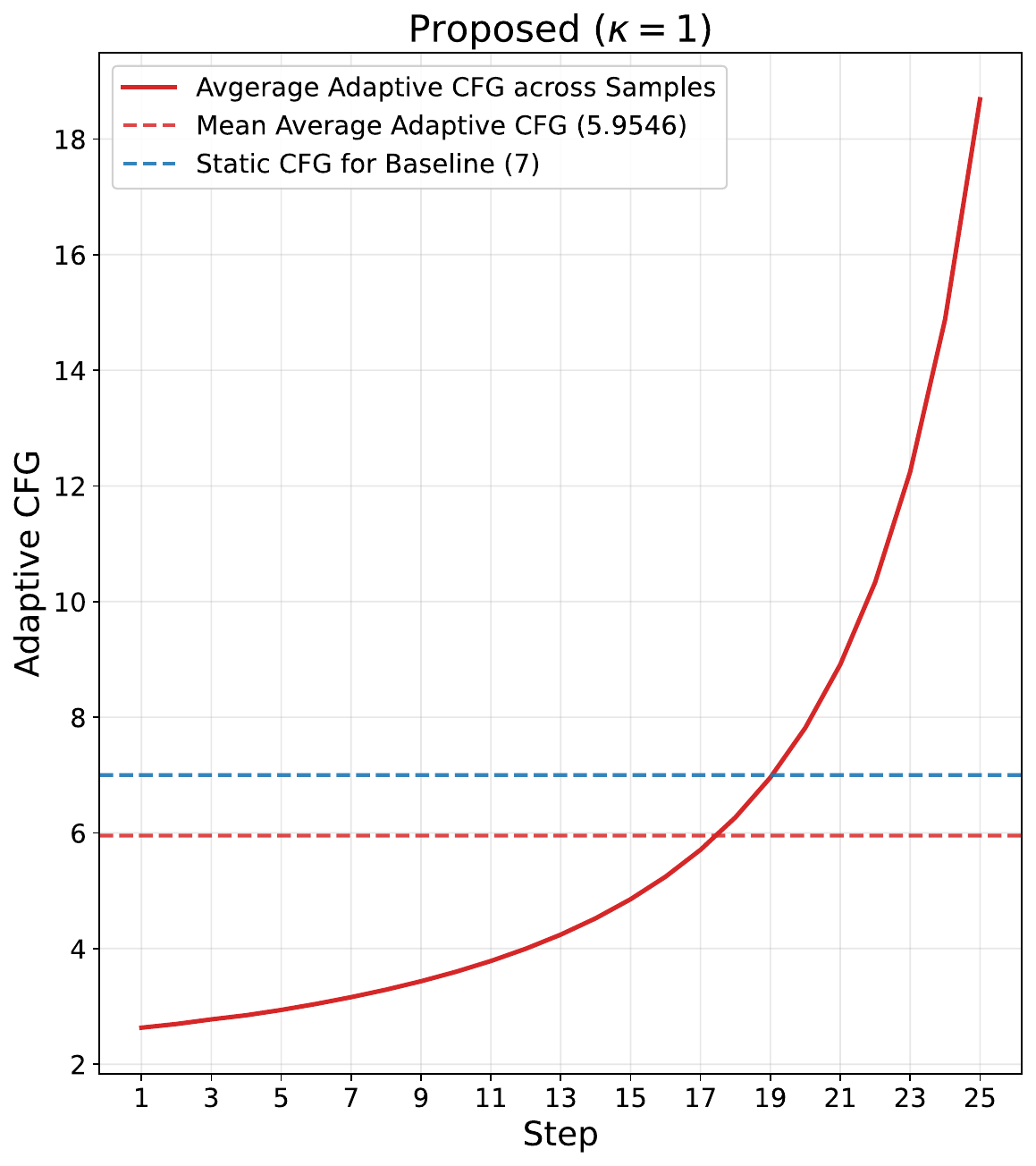}
        \caption{Step-wise adaptive CFG}
        \label{fig:adaptive_cfg_step}
    \end{subfigure}
    \hfill
    \begin{subfigure}[t]{0.48\textwidth}
        \centering
        \includegraphics[width=\linewidth]{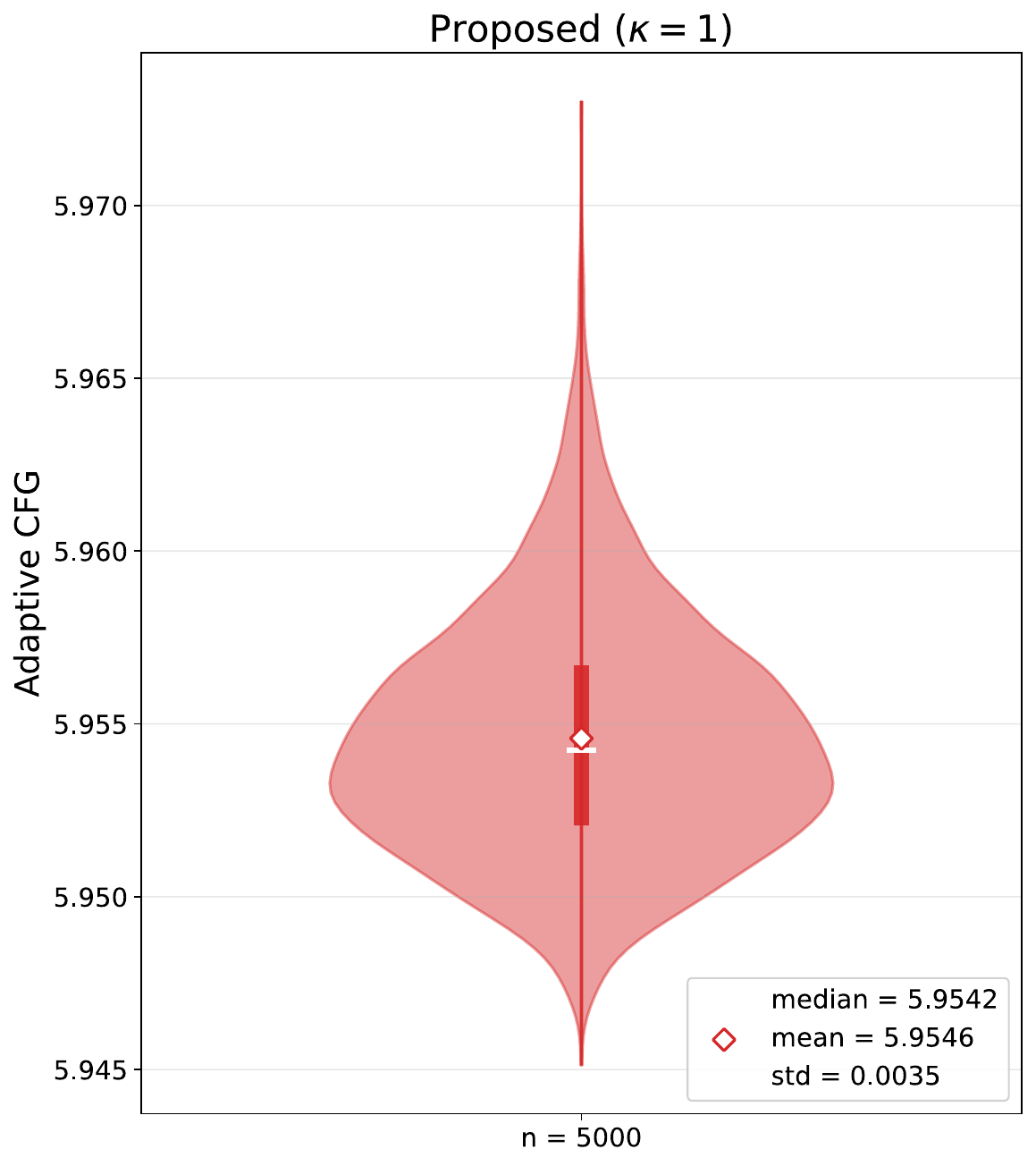}
        \caption{Prompt-wise mean CFG}
        \label{fig:adaptive_cfg_prompt}
    \end{subfigure}
    \caption{Generation diagnostics on COCO17. VAGS-Gen uses the
    unconditional/conditional velocity cosine to reshape the CFG trajectory
    while keeping the prompt-wise mean scale concentrated.}
    \label{fig:generation_alignment_scale_diagnostics}
\end{figure*}

\FloatBarrier
\paragraph{Aggregate alignment statistics.}
Fig.~\ref{fig:generation_alignment_scale_diagnostics} summarises the alignment
and adaptive-scale statistics across COCO17. Within this grouped diagnostic,
Figs.~\ref{fig:cos_sim_step_coco17} and \ref{fig:cos_sim_prompt_coco17}
report step-wise and prompt-wise alignment, while
Figs.~\ref{fig:adaptive_cfg_step} and \ref{fig:adaptive_cfg_prompt} report the
corresponding adaptive scales.

\FloatBarrier

\paragraph{Matched-mean generation control.}
Tab.~\ref{tab:average_cfg_fix_cfg} compares VAGS-Gen with a constant-CFG
baseline set to the mean adaptive value \((\bar{\lambda}=5.955)\). The
matched-mean constant scale yields only a small FID improvement over fixed
CFG, whereas VAGS-Gen obtains a much larger gain.

\FloatBarrier
% Required packages: booktabs, multirow, graphicx, xcolor
\begin{table*}[t!]
\centering
\small
\setlength{\tabcolsep}{5.0pt}
\renewcommand{\arraystretch}{1.12}
\resizebox{\textwidth}{!}{%
\begin{tabular}{@{}l l c c c c c c c@{}}
\toprule
\multirow[c]{2}{*}{\textbf{Dataset}} &
\multirow[c]{2}{*}{\textbf{Method}} &
\multirow[c]{2}{*}{\textbf{FID} $\downarrow$} &
\multirow[c]{2}{*}{\textbf{IS} $\uparrow$} &
\multirow[c]{2}{*}{\textbf{CLIPScore} $\uparrow$} &
\multicolumn{4}{c}{\textbf{Caption-based Metrics}} \\
\cmidrule(lr){6-9}
& & & & &
\textbf{BLEU-4} $\uparrow$ &
\textbf{METEOR} $\uparrow$ &
\textbf{ROUGE-L} $\uparrow$ &
\textbf{CLAIR} $\uparrow$ \\
\midrule

\multirow[c]{3}{*}{\textbf{COCO17}~\cite{lin2014microsoft}}
& SDv3.5 (CFG=7)
& 28.46 & 33.64 & \textbf{32.84}
& 7.99 & 29.17 & 35.11 & 71.45 \\

& SDv3.5 w/ VAGS (CFG=5.955, $\kappa{=}0$)
& \underline{28.28} & \underline{34.63} & \textbf{32.84}
& \underline{8.20} & \textbf{29.69} & \textbf{35.37} & \textbf{72.27} \\

& \cellcolor{gray!15}SDv3.5 w/ VAGS (CFG=7, $\kappa{=}1.0$)
& \cellcolor{gray!15}\textbf{26.07} & \cellcolor{gray!15}\textbf{35.15} & \cellcolor{gray!15}\textbf{32.84}
& \cellcolor{gray!15}\textbf{8.28} & \cellcolor{gray!15}\underline{29.53} & \cellcolor{gray!15}\underline{35.21} & \cellcolor{gray!15}\underline{71.62} \\

\bottomrule
\end{tabular}%
}
\caption{Constant-CFG vs.\ VAGS-Gen on COCO17. A fixed scale at the mean adaptive value
($\bar{\lambda}=5.955$) barely outperforms the baseline, confirming that the gains stem
from dynamic modulation rather than a shifted average guidance level.}
\label{tab:average_cfg_fix_cfg}
\end{table*}

\clearpage
\subsection{Kappa sensitivity}
\suppressfloats[t]
\label{sec:appendix_kappa}

\begin{figure*}[t]
    \centering
    \begin{subfigure}[t]{0.32\textwidth}
        \centering
        \includegraphics[width=\linewidth]{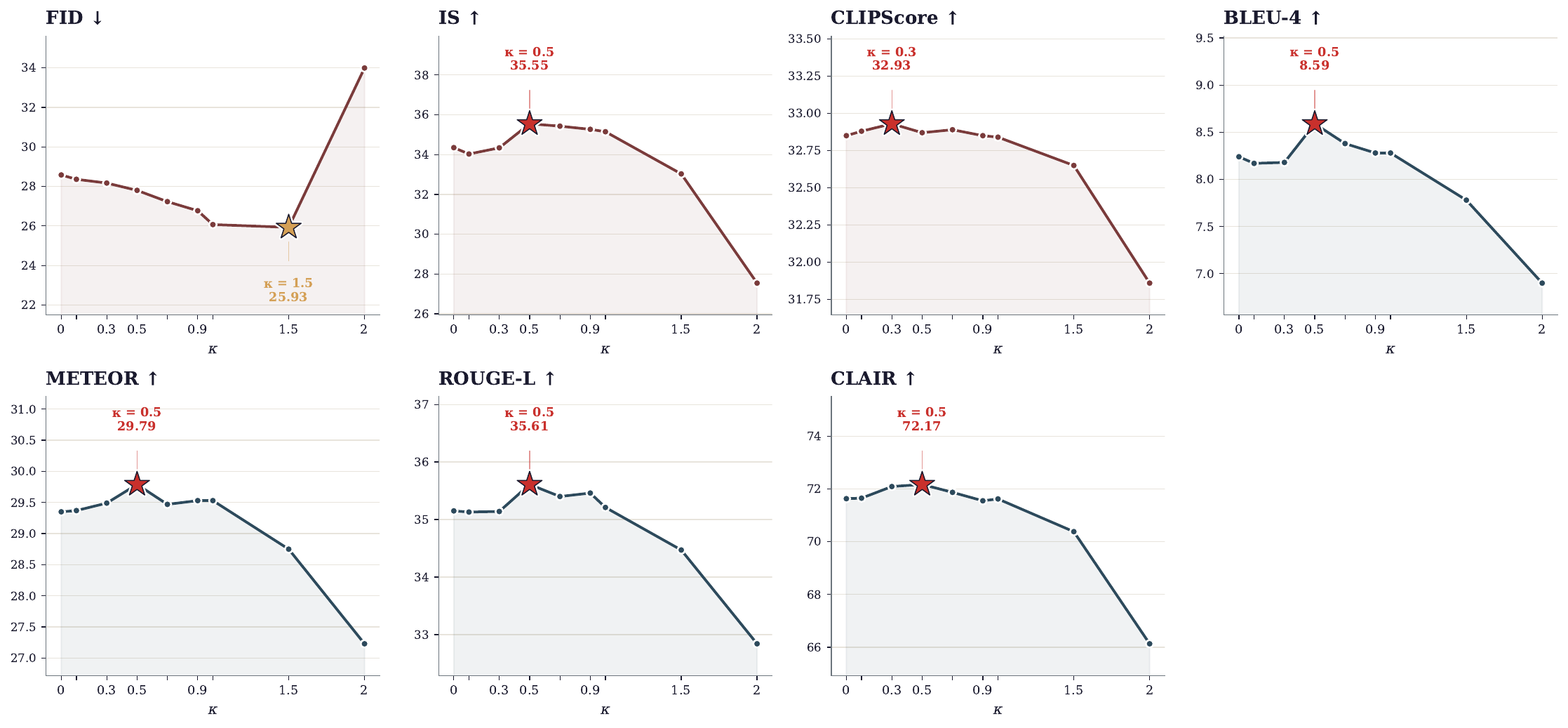}
        \caption{COCO17}
        \label{fig:ablation_kappa_impact_coco17}
    \end{subfigure}
    \hfill
    \begin{subfigure}[t]{0.32\textwidth}
        \centering
        \includegraphics[width=\linewidth]{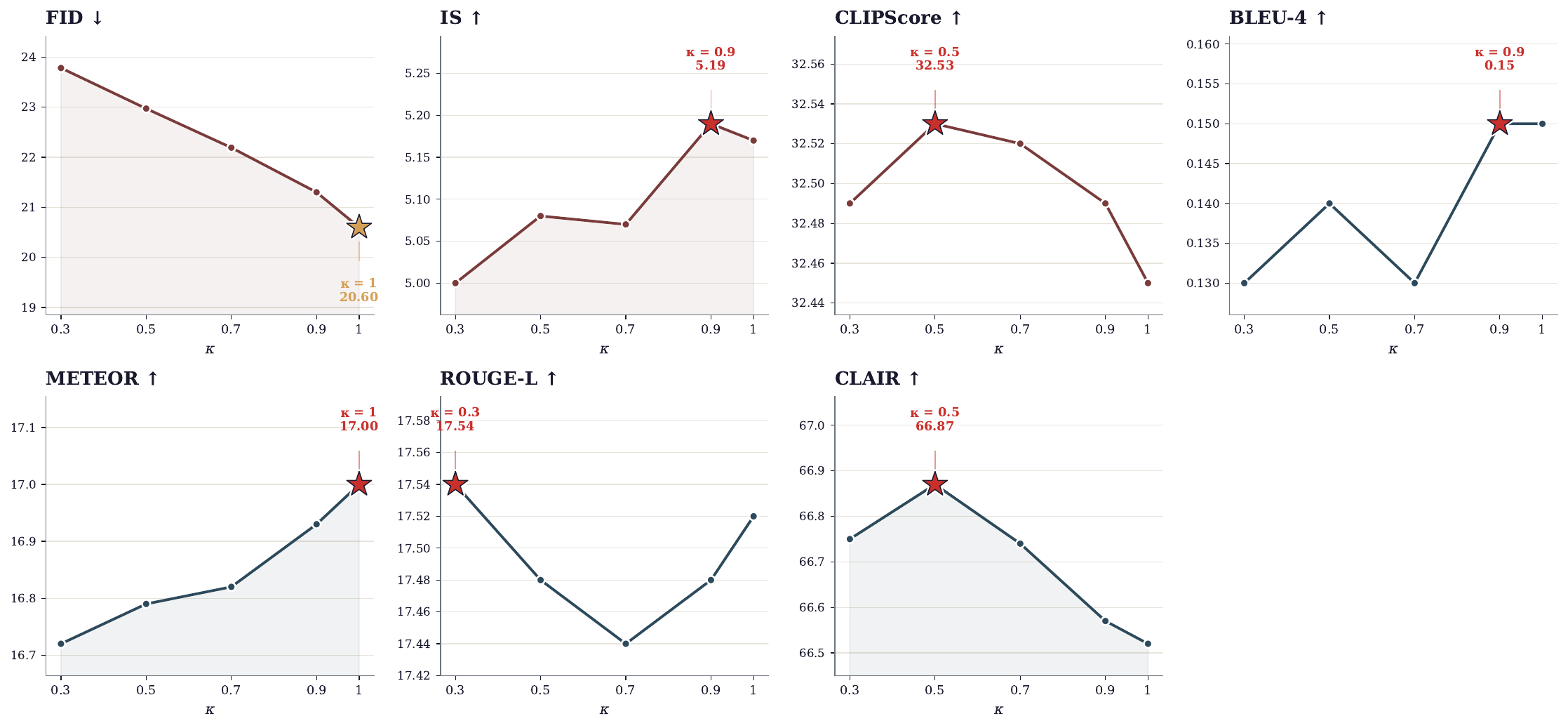}
        \caption{CUB-200}
        \label{fig:ablation_kappa_impact_cub200}
    \end{subfigure}
    \hfill
    \begin{subfigure}[t]{0.32\textwidth}
        \centering
        \includegraphics[width=\linewidth]{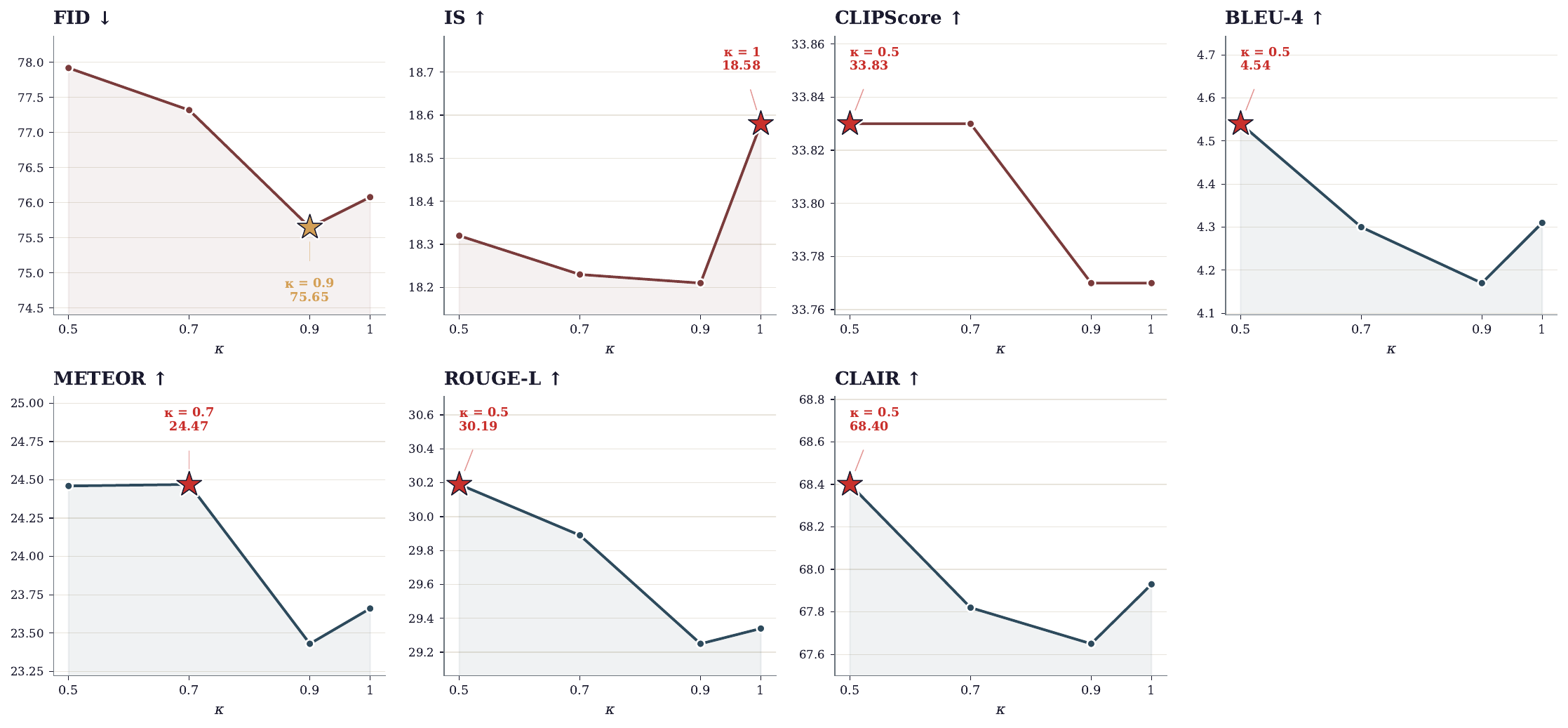}
        \caption{Flickr30K}
        \label{fig:ablation_kappa_impact_flickr30k}
    \end{subfigure}
    \caption{Effect of modulation strength \(\kappa\) on generation metrics.
    Moderate modulation improves fidelity, while overly large values begin to
    suppress or amplify guidance too aggressively.}
    \label{fig:generation_kappa_ablation_summary}
\end{figure*}

\begin{figure}[t]
    \centering
    \includegraphics[width=\linewidth]{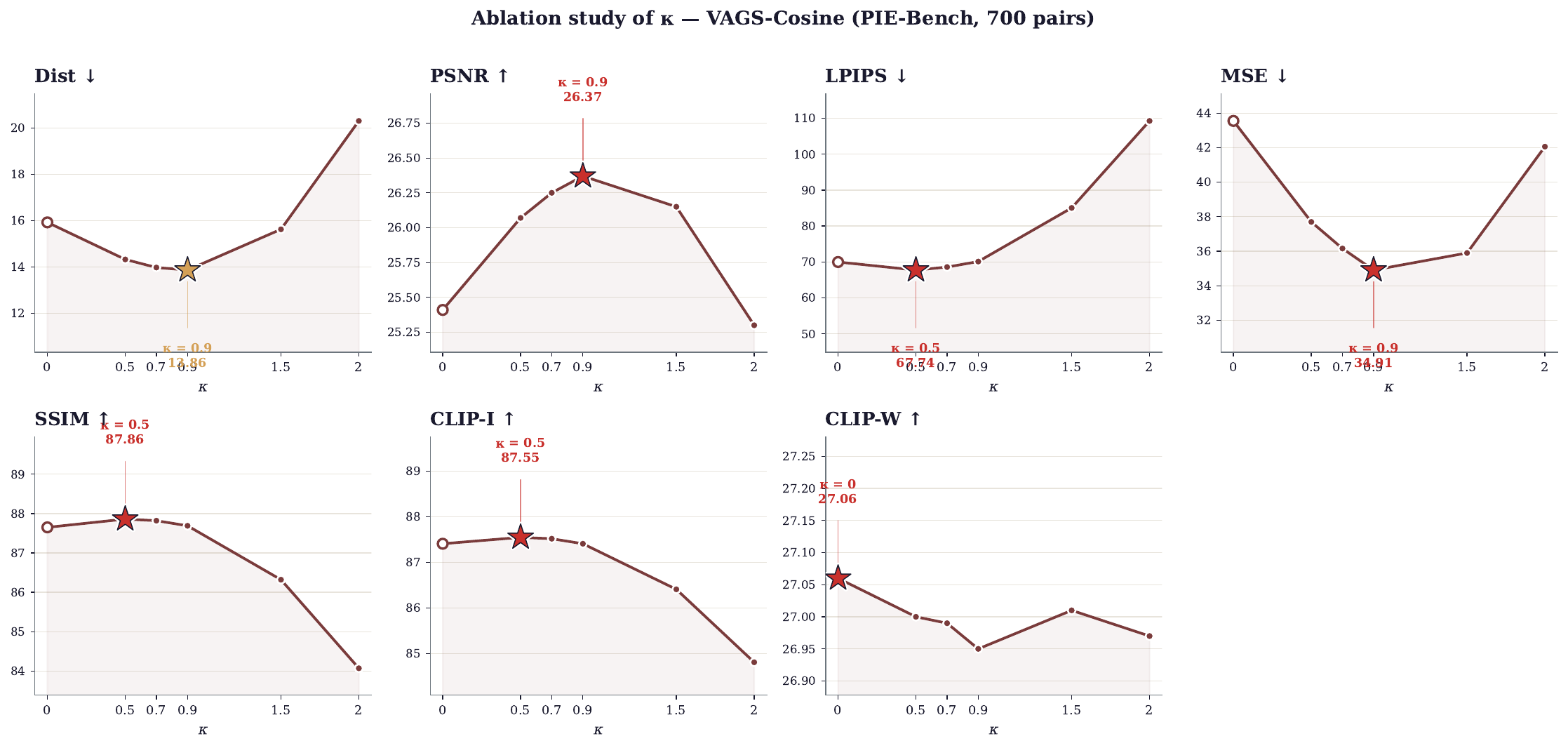}
    \caption{Effect of $\kappa$ on PIE-Bench editing metrics.}
    \label{fig:ablation_kappa_impact_piebench_cosine}
\end{figure}

The modulation strength \(\kappa\) controls the range of adaptive scales.
Tabs.~\ref{tab:ablation_kappa_generation} and
\ref{tab:ablation_kappa_editing} report the raw metrics, while
Figs.~\ref{fig:generation_kappa_ablation_summary},
\ref{fig:ablation_kappa_impact_coco17},
\ref{fig:ablation_kappa_impact_cub200},
\ref{fig:ablation_kappa_impact_flickr30k}, and
\ref{fig:ablation_kappa_impact_piebench_cosine} visualise the trends.
Moderate modulation improves preservation or FID; overly large values begin
to suppress or amplify guidance too aggressively.

\FloatBarrier
% Required packages: booktabs, multirow, adjustbox, xcolor
\begin{table*}[t]
\centering
\scriptsize
\setlength{\tabcolsep}{4.0pt}
\renewcommand{\arraystretch}{1.10}
\caption{Ablation of $\kappa$ on COCO17, CUB-200, and Flickr30K. Top results are in bold and second-best results are underlined. We adopt $\kappa{=}1.0$ as the default.}
\label{tab:ablation_kappa_generation}
\begin{adjustbox}{width=\textwidth}
\begin{tabular}{@{}l l c c c c c c c@{}}
\toprule
\multirow[c]{2}{*}{\textbf{Dataset}} &
\multirow[c]{2}{*}{\textbf{Method}} &
\multirow[c]{2}{*}{\textbf{FID} $\downarrow$} &
\multirow[c]{2}{*}{\textbf{IS} $\uparrow$} &
\multirow[c]{2}{*}{\textbf{CLIPScore} $\uparrow$} &
\multicolumn{4}{c}{\textbf{Caption-based Metrics}} \\
\cmidrule(lr){6-9}
& & & & &
\textbf{BLEU-4} $\uparrow$ &
\textbf{METEOR} $\uparrow$ &
\textbf{ROUGE-L} $\uparrow$ &
\textbf{CLAIR} $\uparrow$ \\
\midrule

\multirow[c]{10}{*}{\textbf{COCO17}~\cite{lin2014microsoft}}
& SDv3.5 & 28.46 & 33.64 & 32.84 & 7.99 & 29.17 & 35.11 & 71.45 \\
& VAGS, $\kappa{=}0$   & 28.58 & 34.35 & 32.85 & 8.24 & 29.35 & 35.15 & 71.63 \\
& VAGS, $\kappa{=}0.1$ & 28.36 & 34.03 & 32.88 & 8.17 & 29.37 & 35.13 & 71.65 \\
& VAGS, $\kappa{=}0.3$ & 28.17 & 34.34 & \textbf{32.93} & 8.18 & 29.49 & 35.14 & \underline{72.09} \\
& VAGS, $\kappa{=}0.5$ & 27.80 & \textbf{35.55} & 32.87 & \textbf{8.59} & \textbf{29.79} & \textbf{35.61} & \textbf{72.17} \\
& VAGS, $\kappa{=}0.7$ & 27.23 & \underline{35.43} & \underline{32.89} & \underline{8.38} & 29.47 & 35.40 & 71.87 \\
& VAGS, $\kappa{=}0.9$ & 26.77 & 35.27 & 32.85 & 8.28 & \underline{29.53} & \underline{35.46} & 71.55 \\
\rowcolor{gray!12}
& VAGS, $\kappa{=}1.0$ & \underline{26.07} & 35.15 & 32.84 & 8.28 & \underline{29.53} & 35.21 & 71.62 \\
& VAGS, $\kappa{=}1.5$ & \textbf{25.93} & 33.03 & 32.65 & 7.78 & 28.75 & 34.47 & 70.38 \\
& VAGS, $\kappa{=}2.0$ & 33.99 & 27.54 & 31.86 & 6.90 & 27.23 & 32.84 & 66.13 \\
\midrule

\multirow[c]{6}{*}{\textbf{CUB-200}~\cite{welinder2010caltech}}
& SDv3.5 & 24.92 & 4.81 & 32.40 & 0.13 & 16.57 & \textbf{17.64} & 66.69 \\
& VAGS, $\kappa{=}0.3$ & 23.78 & 5.00 & 32.49 & 0.13 & 16.72 & \underline{17.54} & \underline{66.75} \\
& VAGS, $\kappa{=}0.5$ & 22.97 & 5.08 & \textbf{32.53} & \underline{0.14} & 16.79 & 17.48 & \textbf{66.87} \\
& VAGS, $\kappa{=}0.7$ & 22.19 & 5.07 & \underline{32.52} & 0.13 & 16.82 & 17.44 & 66.74 \\
& VAGS, $\kappa{=}0.9$ & \underline{21.30} & \textbf{5.19} & 32.49 & \textbf{0.15} & \underline{16.93} & 17.48 & 66.57 \\
\rowcolor{gray!12}
& VAGS, $\kappa{=}1.0$ & \textbf{20.60} & \underline{5.17} & 32.45 & \textbf{0.15} & \textbf{17.00} & 17.52 & 66.52 \\
\midrule

\multirow[c]{5}{*}{\textbf{Flickr30K}~\cite{plummer2015flickr30k}}
& SDv3.5 & 79.58 & 17.95 & 33.79 & 4.12 & 23.88 & 29.30 & 68.03 \\
& VAGS, $\kappa{=}0.5$ & 77.92 & 18.32 & \textbf{33.83} & \textbf{4.54} & \underline{24.46} & \textbf{30.19} & \textbf{68.40} \\
& VAGS, $\kappa{=}0.7$ & 77.32 & 18.23 & \textbf{33.83} & 4.30 & \textbf{24.47} & \underline{29.89} & 67.82 \\
& VAGS, $\kappa{=}0.9$ & \textbf{75.65} & 18.21 & \underline{33.77} & 4.17 & 23.43 & 29.25 & 67.65 \\
\rowcolor{gray!12}
& VAGS, $\kappa{=}1.0$ & \underline{76.08} & \textbf{18.58} & \underline{33.77} & \underline{4.31} & 23.66 & 29.34 & \underline{67.93} \\
\bottomrule
\end{tabular}
\end{adjustbox}
\end{table*}

% Required packages: booktabs, multirow, adjustbox, xcolor
\begin{table*}[t]
\centering
\scriptsize
\setlength{\tabcolsep}{4.0pt}
\renewcommand{\arraystretch}{1.10}
\caption{Ablation of $\kappa$ for image editing on PIE-Bench (700 pairs). Top results are in bold and second-best results are underlined. We adopt $\kappa{=}0.9$ as the default.}
\label{tab:ablation_kappa_editing}
\begin{adjustbox}{width=\textwidth}
\begin{tabular}{@{}l l c c c c c c c c@{}}
\toprule
\multirow[c]{2}{*}{\textbf{Dataset}} &
\multirow[c]{2}{*}{\textbf{Method}} &
\multirow[c]{2}{*}{\textbf{Dist} $\downarrow$} &
\multirow[c]{2}{*}{\textbf{PSNR} $\uparrow$} &
\multirow[c]{2}{*}{\textbf{LPIPS} $\downarrow$} &
\multirow[c]{2}{*}{\textbf{MSE} $\downarrow$} &
\multirow[c]{2}{*}{\textbf{SSIM} $\uparrow$} &
\multicolumn{3}{c}{\textbf{CLIP Similarity}} \\
\cmidrule(lr){8-10}
& & & & & & &
\textbf{Whole} $\uparrow$ &
\textbf{Edited} $\uparrow$ &
\textbf{CLIP-I} $\uparrow$ \\
\midrule

\multirow[c]{7}{*}{\textbf{PIE-Bench}}
& FlowEdit
  & 30.31 & 21.87 & 117.06 & 92.70 & 82.54 & \textbf{27.62} & \textbf{23.80} & 83.69 \\
& FlowEdit + Monotone ($\kappa{=}0$)
  & 15.91 & 25.43 & 69.87 & 43.47 & 87.67 & \underline{27.05} & \underline{23.26} & 84.73 \\
& VAGS, $\kappa{=}0.5$
  & 14.32 & 26.07 & \textbf{67.74} & 37.70 & \textbf{87.86} & 27.00 & 23.25 & \textbf{87.55} \\
& VAGS, $\kappa{=}0.7$
  & \underline{13.97} & \underline{26.25} & \underline{68.58} & 36.16 & \underline{87.82} & 26.99 & 23.14 & \underline{87.52} \\
\rowcolor{gray!12}
& VAGS, $\kappa{=}0.9$
  & \textbf{13.84} & \textbf{26.38} & 70.38 & \textbf{34.86} & 87.68 & 26.92 & 23.08 & 87.44 \\
& VAGS, $\kappa{=}1.5$
  & 15.62 & 26.15 & 85.04 & \underline{35.90} & 86.32 & 27.01 & 23.00 & 86.41 \\
& VAGS, $\kappa{=}2.0$
  & 20.30 & 25.30 & 109.25 & 42.06 & 84.07 & 26.97 & 23.03 & 84.81 \\

\bottomrule
\end{tabular}
\end{adjustbox}
\end{table*}

\paragraph{Generation trends.}
On COCO17, FID decreases monotonically from \(28.58\) at \(\kappa{=}0\)
(constant CFG) to \(26.07\) at \(\kappa{=}1.0\), reaching its absolute
minimum of \(25.93\) at \(\kappa{=}1.5\). IS, BLEU-4, METEOR, ROUGE-L,
and CLAIR all peak earlier at \(\kappa{=}0.5\) (IS \(35.55\),
BLEU-4 \(8.59\), CLAIR \(72.17\)) and degrade past \(\kappa{=}1.0\);
\(\kappa{=}2.0\) is a clear collapse on every measure (FID \(33.99\),
IS \(27.54\), CLAIR \(66.13\)). We adopt \(\kappa{=}1.0\) rather than
the slightly lower \(\kappa{=}1.5\) (FID \(25.93\)) because moving from
\(\kappa{=}1.0\) to \(\kappa{=}1.5\) saves only \(0.14\) FID while
losing \(2.12\) on IS, \(0.50\) on BLEU-4, \(0.78\) on METEOR, \(0.74\)
on ROUGE-L, and \(1.24\) on CLAIR. On CUB-200, FID decreases
monotonically from \(23.78\) at \(\kappa{=}0.3\) to \(20.60\) at
\(\kappa{=}1.0\) over the tested range, with no observed collapse,
suggesting that fine-grained generation benefits from the full bound
allowed by \(\kappa{=}1.0\). On Flickr30K the optimum sits slightly
lower at \(\kappa{=}0.9\) (FID \(75.65\) vs.\ \(76.08\) at
\(\kappa{=}1.0\)); the more diverse Flickr30K scenes appear to favour a
marginally milder correction, hence the per-dataset choice.

\paragraph{Editing trends.}
On PIE-Bench, structure-oriented metrics reach their optimum at
\(\kappa_\mathrm{tar}{=}0.9\) (Dist \(13.84\), PSNR \(26.38\),
MSE \(34.86\)), whereas perceptual metrics peak earlier at
\(\kappa_\mathrm{tar}{=}0.5\) (LPIPS \(67.74\), SSIM \(87.86\)). We
adopt \(\kappa_\mathrm{tar}{=}0.9\) since the preservation/editability
trade-off central to inversion-free editing is governed primarily by
structural fidelity, and the perceptual loss between
\(\kappa_\mathrm{tar}{=}0.5\) and \(\kappa_\mathrm{tar}{=}0.9\) is
small: LPIPS rises by \(2.64\) and SSIM drops by \(0.18\) across that
range. Beyond \(\kappa_\mathrm{tar}{=}0.9\) every metric degrades: at
\(\kappa_\mathrm{tar}{=}2.0\), Dist rises from \(13.84\) to \(20.30\),
LPIPS rises to \(109.25\), and SSIM drops to \(84.07\), the same
clean-regime over-amplification observed on the generation side.

% \clearpage
% \subsection{Runtime and complexity}
% \suppressfloats[t]
% \label{sec:appendix_runtime}

% The dominant cost of every flow-based sampler considered here is the per-step
% forward pass through the velocity network \(v_\theta\), whose cost we denote
% \(T_{\mathrm{net}}\); all other operations are \(\mathcal{O}(d)\) in the
% latent dimension \(d\) and negligible by comparison. For editing, the
% inversion-free baseline performs a single batched forward pass of size four
% per step, covering unconditional and conditional predictions for both source
% and target prompts. VAGS reuses these raw predictions to assemble the pilot
% target velocity, compute one cosine similarity, and re-weight the target
% velocity in closed form. The total complexity therefore remains
% \(\mathcal{O}(N\,T_{\mathrm{net}})\) with no extra network evaluations.

% For generation, constant CFG performs one batched forward pass of size two per
% step. VAGS-Gen computes \(s^{\mathrm{gen}}_i\) directly from the same two raw
% predictions, again adding only one inner product per step. In measured runtime
% on a single NVIDIA H100, editing takes \(7.5\,\mathrm{s}\) per image for the
% baseline and \(7.6\,\mathrm{s}\) for VAGS, a \(1.3\%\) overhead. Generation
% takes \(11.3\,\mathrm{s}\) for the baseline and \(11.5\,\mathrm{s}\) for
% VAGS-Gen, a \(1.8\%\) overhead.

\clearpage
\subsection{CLIP-scaled metric tables}
\suppressfloats[t]
\label{sec:appendix_clipscaled}

Tables~\ref{tab:sota_comparison_clipscaled},
\ref{tab:guidance_ablation_clipscaled}, \ref{tab:perf_comparison_clipscaled},
\ref{tab:ablation_kappa_generation_clipscaled}, and
\ref{tab:average_cfg_fix_cfg_clipscaled} report the same comparisons as the
main paper and sensitivity studies after normalising non-CLIP metrics by
whole-image CLIP similarity.
The purpose is to penalise methods whose low distortion or high fidelity
scores are partly explained by weak text-image alignment.
For every cell except the CLIP scores themselves, the CLIP-scaled value in
parentheses is computed as
\(\mathrm{scaled} = \mathrm{raw}/(\mathrm{CLIP}_{\mathrm{W}}/100)^2\) for
lower-is-better metrics and
\(\mathrm{scaled} = \mathrm{raw}\cdot(\mathrm{CLIP}_{\mathrm{W}}/100)^2\) for
higher-is-better metrics.

\paragraph{Editing SOTA.}
On PIE-Bench, FTEdit's raw Structure Distance \(13.81\) sits within
\(0.03\) of FlowEdit+VAGS at \(13.84\), but its lower CLIP-W
(\(25.01\) vs.\ \(26.92\)) yields a substantially larger scaled value
(\(220.78\) vs.\ \(190.98\)); VAGS therefore leads by roughly \(13\%\)
on scaled Dist where the raw gap is essentially zero. FTEdit's bold raw
SSIM (\(92.62\)) and CLIP-I (\(92.40\)) likewise fall below the VAGS
row when scaled (\(5.79\) and \(5.78\) vs.\ \(6.35\) and \(6.34\)).
On DIV2K, VAGS is bold on every non-CLIP metric in raw form and
retains the lead under scaling.

\FloatBarrier
% Required packages: booktabs, multirow, graphicx (for \resizebox)
\begin{table*}[t!]
\centering
\small
\setlength{\tabcolsep}{4.5pt}
\renewcommand{\arraystretch}{1.08}
\resizebox{\textwidth}{!}{%
\begin{tabular}{l l c c c c c c c c c}
\toprule
\multirow{2}{*}{\textbf{Dataset}} &
\multirow{2}{*}{\textbf{Method}} &
\multirow{2}{*}{\textbf{Model}} &
\multicolumn{1}{c}{\textbf{Structure}} &
\multicolumn{5}{c}{\textbf{Background Preservation}} &
\multicolumn{2}{c}{\textbf{CLIP Similarity}} \\
\cmidrule(lr){4-4}\cmidrule(lr){5-9}\cmidrule(lr){10-11}
& & &
\textbf{Dist.\,$\times10^{3}\downarrow$} &
\textbf{PSNR $\uparrow$} &
\textbf{LPIPS $\times10^{3}\downarrow$} &
\textbf{MSE $\times10^{4}\downarrow$} &
\textbf{SSIM $\times10^{2}\uparrow$} &
\textbf{CLIP-I $\uparrow$} &
\textbf{Whole $\uparrow$} &
\textbf{Edited $\uparrow$}
\\
\midrule

%% ── PIE-Bench ──────────────────────────────────────────────────────────
\multirow{20}{*}{\textbf{PIE-Bench}}
& \multicolumn{9}{l}{\textit{Diffusion-based}} \\
& DDIM~\cite{song2021ddim} + P2P~\cite{hertz2022prompt2prompt}
  % & SD~1.4  & 69.40 & 17.87 & 208.80 & 219.88 & 22.44 & 71.14 & 25.01\\
  & SD~1.4 & 71.12\,(1059.39) & 17.84\,(1.20) & 207.62\,(3092.68) & 220.79\,(3288.85) & 71.62\,(4.81) & 82.44\,(5.53) & 25.91 & 22.75\\
& DDIM~\cite{song2021ddim} + PnP~\cite{tumanyan2023pnp}
  % & SD~1.4  & 28.22 & 22.28 & 113.46 & 83.64 & 22.55 & 79.05 & 25.41\\
  & SD~1.4 & 27.27\,(399.40) & 22.36\,(1.53) & 110.22\,(1614.29) & 81.94\,(1200.10) & 79.65\,(5.44) & 86.93\,(5.94) & 26.13 & 22.79\\

% & PnP-Inv~\cite{tumanyan2023pnp} + P2P~\cite{hertz2022prompt2prompt}
%   & SD~1.4 & \textbf{10.94\,(163.97)} & \textbf{27.29\,(1.82)} & \textbf{48.91\,(733.08)} & \textbf{32.57\,(488.17)} & 85.39\,(5.70) & \underline{91.20\,(6.08)} & 25.83 & 22.41\\
& Null-text~\cite{mokady2023nulltext} + P2P~\cite{hertz2022prompt2prompt}
  % & SD~2.1  & 13.44 & 27.03 & 60.67  & 35.86 & 21.86 & 84.11 & 24.75\\
  & SD~1.4 & 15.69\,(248.85) & \textbf{26.92\,(1.70)} & \textbf{66.42\,(1053.43)} & \underline{35.09\,(556.53)} & 82.67\,(5.21) & 85.42\,(5.39) & 25.11 & 22.01\\

% & PnP-Inv + PnP~\cite{tumanyan2023pnp}
%   & SD~1.4 & 23.19\,(337.06) & 22.57\,(1.55) & 103.10\,(1498.52) & 78.42\,(1139.80) & 80.42\,(5.53) & 87.78\,(6.04) & 26.23 & 22.87\\
\cmidrule(l){2-11}
& \multicolumn{9}{l}{\textit{Flow-based / Rectified Flow}} \\
& RF-Inversion~\cite{rout2024rfinversion}
  % & FLUX.1  & 40.60 & 20.82 & 184.80 & 129.10 & 22.11 & 71.92 & 25.20\\
  & FLUX.1 & 47.19\,(689.56) & 20.64\,(1.41) & 193.68\,(2830.15) & 123.33\,(1802.16) & 76.34\,(5.22) & 80.23\,(5.49) & 26.16 & 22.38\\
& RF-Solver~\cite{wang2024rfsolver}
  % & FLUX.1  & 31.10 & 22.90 & 135.81 & 80.11 & 22.88 & 81.90 & 26.00\\
  & FLUX.1 & 48.04\,(677.42) & 20.52\,(1.46) & 181.15\,(2554.44) & 128.54\,(1812.57) & 77.81\,(5.52) & 82.20\,(5.83) & 26.63 & 23.46\\
& FireFlow~\cite{deng2024fireflow} + RF-Solver~\cite{wang2024rfsolver}
  % & FLUX.1  & 28.30 & 23.28 & 120.82 & 70.39 & 22.94 & 82.82 & 25.98\\
  & FLUX.1 & 25.88\,(384.02) & 22.90\,(1.54) & 129.27\,(1918.18) & 74.73\,(1108.88) & 82.03\,(5.53) & 84.76\,(5.71) & 25.96 & 22.02\\
& FlowEdit~\cite{kulikov2024flowedit}
  % & FLUX.1  & 27.70 & 21.91 & 111.70 & 94.00 & 22.70 & 83.39 & 25.61\\
  & FLUX.1 & 27.93\,(407.19) & 21.74\,(1.49) & 113.83\,(1659.53) & 98.75\,(1439.68) & 83.46\,(5.72) & 86.54\,(5.94) & 26.19 & 22.19\\
& FTEdit~\cite{xu2025ftedit} + AdaLN
  % & SD~3.5  & 18.17 & 26.62 & 80.55  & 40.24 & 22.27 & 91.50 & 25.74\\
  & SD~3.5 & \textbf{13.81\,(220.78)} & 23.69\,(1.48) & 77.69\,(1242.05) & 36.10\,(577.14) & \textbf{92.62\,(5.79)} & \textbf{92.40\,(5.78)} & 25.01 & 21.81\\
& iRFDS~\cite{yang2025irfds}
  % & SD~3.5  & 45.04 & 16.25 & 312.24 & 305.83 & 25.38 & 57.03 & 25.46\\
  & SD~3.5 & 67.64\,(1044.31) & 19.42\,(1.26) & 150.68\,(2326.38) & 147.86\,(2282.84) & 79.47\,(5.15) & 77.69\,(5.03) & 25.45 & 21.73\\
& FlowEdit~\cite{kulikov2024flowedit}
  % & SD~3.5  & 30.32 & 18.31 & 207.09 & 169.11 & 27.56 & 68.96 & 27.63\\
  & SD~3.5 & 30.31\,(397.32) & 21.87\,(1.67) & 117.06\,(1534.48) & 92.70\,(1215.16) & 82.54\,(6.30) & 83.69\,(6.38) & \textbf{27.62} & \textbf{23.80}\\
& SplitFlow ~\cite{yoon2025splitflow}
  & SD~3.5  & 28.03\,(370.11) & 22.32\,(1.69) & 111.49\,(1472.11) & 85.31\,(1126.43) & 83.17\,(6.30) & 84.10\,(6.37) & \underline{27.52} & \underline{23.77}\\
%& SplitFlow (Official Repo)~\cite{yoon2025splitflow}
 % & SD~3  & 27.20 & 21.94 & 105.48 & 92.35 & 83.41 & - & 27.51 & 23.89\\
%& SplitFlow (SplitFlow Paper)~\cite{yoon2025splitflow}
 % & SD~3  & 25.96 & 22.45 & 102.14 & 81.99 & 83.91 & - & 26.96 & 23.83\\
\cmidrule(l){2-11}
& \multicolumn{9}{l}{\textit{Ours}} \\

  & \cellcolor{gray!15}FlowEdit + VAGS
  % & SD~3.5  & 23.52 & 19.74 & 186.19 & 125.96 & 27.44 & 71.43 & 27.53\\
  & \cellcolor{gray!15}SD~3.5 & \cellcolor{gray!15}\underline{13.84\,(190.98)} & \cellcolor{gray!15}\underline{26.38\,(1.91)} & \cellcolor{gray!15}\underline{70.38\,(971.18)} & \cellcolor{gray!15}\textbf{34.86\,(481.04)} & \cellcolor{gray!15}\underline{87.68\,(6.35)} & \cellcolor{gray!15}\underline{87.44\,(6.34)} & \cellcolor{gray!15}26.92 & \cellcolor{gray!15}23.08\\
  % & FlowEdit + VAGS-Cosine (Updated w/ $\kappa$=0.9) & SD~3.5 &  23.52 & 23.49 & 103.56 & 66.92 & 83.97 & 84.45 & 27.54 & 23.65\\
  % & FlowEdit + VAGS-Cosine (alt run,  $\kappa$=1.5) & SD~3.5 & 15.61 & 25.46 & 99.51 & 41.40 & 84.31 & 86.42 & 27.01 & 23.00\\
% & FlowEdit + VAGS-Ratio
  % & SD~3.5  & 24.21 & 19.44 & 183.17 & 131.64 & 27.40 & 71.86 & 27.50\\
  % & SD~3.5 & 13.54 & 26.29 & 64.35 & 35.50 & 88.24 & 87.85 & 26.93 & 23.04\\
  % & FlowEdit + VAGS-Cosine (Updated w/ $\kappa$=0.5) & SD~3.5 &  25.55 & 22.84 & 104.88 & 75.61 & 83.81 & 84.44 & 27.50 & 23.76\\
  % & FlowEdit + VAGS-Ratio (alt run,  $\kappa$=0.5) & SD~3.5 & 14.08 & 25.35 & 77.91 & 43.82 & 86.31 & 87.80 & 26.96 & 23.18\\

\midrule

%% ── DIV2K ───────────────────────────────────────────────────────────────
\multirow{20}{*}{\textbf{DIV2K}}
& \multicolumn{9}{l}{\textit{Diffusion-based}} \\
& DDIM~\cite{song2021ddim} + P2P~\cite{hertz2022prompt2prompt}
  % & SD~1.4  & 78.50 & 13.74 & 383.42 & 455.00 & 29.50 & 51.35 & 29.82\\
  % & SD~1.4 & 81.18 & 14.19 & 365.17 & 421.25 & 55.45 & 76.82 & 29.26 & -\\
  & SD~1.4 & 64.46\,(921.38) & 15.00\,(1.05) & 323.57\,(4625.05) & 359.10\,(5132.91) & 60.93\,(4.26) & 81.89\,(5.73) & 26.45 & -\\
& DDIM~\cite{song2021ddim} + PnP~\cite{tumanyan2023pnp}
  % & SD~1.4  & 31.60 & 18.02 & 228.35 & 179.00 & 28.30 & 65.31 & 28.56\\
  & SD~1.4 & 30.66\,(380.94) & 18.75\,(1.51) & 205.72\,(2555.98) & 154.73\,(1922.45) & 68.61\,(5.52) & 83.83\,(6.75) & 28.37 & -\\
& Null-text~\cite{mokady2023nulltext} + P2P~\cite{hertz2022prompt2prompt}
  % & SD~2.1  & 24.50 & 19.15 & 170.20 & 142.40 & 27.25 & 76.20 & 27.59\\
  & SD~1.4 & 15.82\,(207.53) & 21.35\,(1.63) & 137.28\,(1800.84) & 93.17\,(1222.20) & 74.37\,(5.67) & \underline{88.32\,(6.73)} & 27.61 & -\\
% & PnP-Inv~\cite{tumanyan2023pnp} + P2P~\cite{hertz2022prompt2prompt}
%   & SD~1.4 & \underline{12.15\,(157.10)} & 22.29\,(1.72) & \underline{110.24\,(1425.40)} & 79.47\,(1027.55) & 75.42\,(5.83) & 87.94\,(6.80) & 27.81 & -\\
% & PnP-Inv + PnP~\cite{tumanyan2023pnp}
%   & SD~1.4 & 26.09\,(325.30) & 19.08\,(1.53) & 190.37\,(2373.62) & 142.24\,(1773.52) & 69.57\,(5.58) & 84.11\,(6.75) & 28.32 & -\\
\cmidrule(l){2-11}
& \multicolumn{9}{l}{\textit{Flow-based / Rectified Flow}} \\
& RF-Inversion~\cite{rout2024rfinversion}
  % & FLUX.1  & 55.40 & 16.20 & 320.60 & 285.40 & 30.85 & 62.40 & 30.99\\
  & FLUX.1 & 31.87\,(363.75) & 20.05\,(1.76) & 252.75\,(2884.75) & 115.45\,(1317.68) & 73.37\,(6.43) & 79.57\,(6.97) & 29.60 & -\\
& RF-Solver~\cite{wang2024rfsolver}
  % & FLUX.1  & 48.20 & 16.55 & 245.20 & 272.10 & 30.80 & 64.15 & 30.96\\
  & FLUX.1 & 34.63\,(387.36) & 18.56\,(1.66) & 265.56\,(2970.44) & 161.60\,(1807.59) & 72.98\,(6.52) & 79.54\,(7.11) & 29.90 & -\\
& FireFlow~\cite{deng2024fireflow} + RF-Solver~\cite{wang2024rfsolver}
  % & FLUX.1  & 25.40 & 19.11 & 210.40 & 136.00 & 29.60 & 71.06 & 29.75\\
  & FLUX.1 & 20.52\,(247.91) & 21.02\,(1.74) & 197.26\,(2383.19) & 92.75\,(1120.56) & 77.88\,(6.45) & 82.86\,(6.86) & 28.77 & -\\
& FlowEdit~\cite{kulikov2024flowedit}
  % & FLUX.1  & 27.00 & 17.83 & 191.09 & 183.00 & 29.85 & 73.03 & 30.06\\
  & FLUX.1 & 24.84\,(293.94) & 18.73\,(1.58) & 173.21\,(2049.67) & 152.33\,(1802.58) & 78.81\,(6.66) & 83.93\,(7.09) & 29.07 & -\\
& FTEdit~\cite{xu2025ftedit} + AdaLN
  % & SD~3.5  & 26.00 & 19.80 & 184.76 & 118.00 & 29.05 & 79.50 & 29.13\\
  & SD~3.5 & \underline{14.43\,(197.50)} & \underline{23.02\,(1.68)} & \underline{125.97\,(1724.15)} & \underline{56.65\,(775.37)} & \underline{83.73\,(6.12)} & \textbf{88.97\,(6.50)} & 27.03 & -\\
& iRFDS~\cite{yang2025irfds}
  % & SD~3.5  & 65.40 & 18.50 & 208.30 & 195.50 & 28.35 & 71.20 & 28.55\\
  & SD~3.5 & 61.14\,(810.23) & 15.86\,(1.20) & 317.41\,(4206.33) & 320.07\,(4241.58) & 63.02\,(4.76) & 76.79\,(5.79) & 27.47 & -\\
& FlowEdit~\cite{kulikov2024flowedit}
  % & SD~3.5  & 25.15 & 20.27 & 145.63 & 103.77 & 30.25 & 82.15 & 30.33\\
  & SD~3.5 & 18.27\,(202.19) & 20.27\,(1.83) & 145.55\,(1610.77) & 103.37\,(1143.98) & 82.15\,(7.42) & 82.31\,(7.44) & \textbf{30.06} & -\\
& SplitFlow~\cite{yoon2025splitflow}
  % & SD~3.5  & 24.77 & 20.53 & 141.11 & 97.52 & 30.45 & 82.73 & 30.57\\
  & SD~3.5 & 17.61\,(196.32) & 20.53\,(1.84) & 141.99\,(1582.94) & 97.19\,(1083.50) & 82.63\,(7.41) & 82.44\,(7.39) & \underline{29.95} & -\\
\cmidrule(l){2-11}
& \multicolumn{9}{l}{\textit{Ours}} \\
& \cellcolor{gray!15}FlowEdit + VAGS
  % & SD~3.5  & NA & 21.70 & NA & 75.90 & 29.87 & 82.35 & 30.01\\
  & \cellcolor{gray!15}SD~3.5 & \cellcolor{gray!15}\textbf{8.62\,(98.32)} & \cellcolor{gray!15}\textbf{23.61\,(2.07)} & \cellcolor{gray!15}\textbf{98.43\,(1122.67)} & \cellcolor{gray!15}\textbf{50.22\,(572.80)} & \cellcolor{gray!15}\textbf{84.41\,(7.40)} & \cellcolor{gray!15}84.07\,(7.37) & \cellcolor{gray!15}29.61 & \cellcolor{gray!15}-\\
% & FlowEdit + VAGS-Ratio
  % & SD~3.5 & NA & 21.01 & NA & 86.87 & 29.81 & 81.56 & 29.91\\
  % & SD~3.5 & 9.38 & 23.14 & 95.70 & 54.67 & 84.39 & 84.39 & 29.62 & -\\

\bottomrule
\end{tabular}%
}
\caption{[CLIP-scaled] Quantitative comparison on PIE-Bench and DIV2K. FlowEdit + VAGS ranks first on all background preservation metrics while maintaining competitive edit strength.}
\label{tab:sota_comparison_clipscaled}
\end{table*}

\paragraph{Editing scheduler ablation.}
The scaled column reorders the failure modes more sharply than the raw
column. Zero-Init's PIE-Bench Structure Distance of \(194.67\) becomes
\(2406.80\) under scaling on FlowEdit, roughly a \(12\times\) penalty
over VAGS' \(190.98\). The penalty comes overwhelmingly from the
catastrophic raw Distance (\(194.67\) vs.\ \(13.84\)), since
Zero-Init's CLIP-W is comparable to the other rows. The bold ranking on Dist, PSNR,
MSE, SSIM, and CLIP-I is otherwise unchanged: VAGS leads on both
editors and on both datasets in raw and scaled form.

\FloatBarrier
% Required packages: booktabs, multirow, graphicx (for \resizebox)
\begin{table*}[t!]
\centering
\small
\setlength{\tabcolsep}{4.5pt}
\renewcommand{\arraystretch}{1.08}
\resizebox{\textwidth}{!}{%
\begin{tabular}{l l l c c c c c c c c}
\toprule
\multirow{2}{*}{\textbf{Dataset}} &
\multirow{2}{*}{\textbf{Editor}} &
\multirow{2}{*}{\textbf{Scheduler}} &
\multicolumn{1}{c}{\textbf{Structure}} &
\multicolumn{5}{c}{\textbf{Background Preservation}} &
\multicolumn{2}{c}{\textbf{CLIP Similarity}} \\
\cmidrule(lr){4-4}\cmidrule(lr){5-9}\cmidrule(lr){10-11}
& & &
\textbf{Dist.\,$\times10^{3}\downarrow$} &
\textbf{PSNR $\uparrow$} &
\textbf{LPIPS $\times10^{3}\downarrow$} &
\textbf{MSE $\times10^{4}\downarrow$} &
\textbf{SSIM $\times10^{2}\uparrow$} &
\textbf{CLIP-I $\uparrow$} &
\textbf{Whole $\uparrow$} &
\textbf{Edited $\uparrow$}
\\
\midrule

%% ── PIE-Bench ──────────────────────────────────────────────────────────
\multirow{10}{*}{\textbf{PIE-Bench}}
& \multirow{5}{*}{FlowEdit}
% & No Scheduler
%   & 11.65 & 27.22 & 54.55 & 32.86 & 22.10 & 84.76 & 25.02\\
 & No Scheduler
  % & 30.32 & 18.31 & 207.09 & 169.11 & 27.56 & 68.96 & 27.63\\
   & 30.31\,(397.32) & 21.87\,(1.67) & 117.06\,(1534.48) & 92.70\,(1215.16) & 82.54\,(6.30) & 83.69\,(6.38) & \underline{27.62} & \underline{23.80}\\
&&  Interval~\cite{kynkaanniemi2024applying} 
  % & 29.82 & 18.35 & 202.47 & 167.47 & 27.51 & 69.61 & 27.57\\
  & 29.82\,(392.03) & 21.93\,(1.67) & 114.14\,(1500.55) & 91.72\,(1205.80) & 82.88\,(6.30) & 84.03\,(6.39) & 27.58 & 23.75\\
&& Monotone~\cite{wang2024analysis}
  % & 15.91 & 21.49 & 132.86 & 85.86 & 26.96 & 78.48 & 27.05\\
  & \underline{15.91\,(217.44)} & \underline{25.43\,(1.86)} & \textbf{69.87\,(954.90)} & \underline{43.47\,(594.09)} & \underline{87.67\,(6.41)} & \underline{84.73\,(6.20)} & 27.05 & 23.26\\
&& Zero-Init~\cite{fan2025cfg} 
  % & 194.67 & 8.98 & 560.19 & 1385.85 & 28.46 & 32.46 & 28.44\\
  & 194.67\,(2406.80) & 11.69\,(0.946) & 354.57\,(4383.72) & 867.59\,(10726.43) & 57.64\,(4.66) & 75.28\,(6.09) & \textbf{28.44} & \textbf{24.24}\\
&& \cellcolor{gray!15}VAGS (Ours)
  % & 23.52 & 19.74 & 186.19 & 125.96 & 27.44 & 71.43 & 27.53\\
  & \cellcolor{gray!15}\textbf{13.84\,(190.98)} & \cellcolor{gray!15}\textbf{26.38\,(1.91)} & \cellcolor{gray!15}\underline{70.38\,(971.18)} & \cellcolor{gray!15}\textbf{34.86\,(481.04)} & \cellcolor{gray!15}\textbf{87.68\,(6.35)} & \cellcolor{gray!15}\textbf{87.44\,(6.34)} & \cellcolor{gray!15}26.92 & \cellcolor{gray!15}23.08\\
  % && VAGS-Ratio (Ours)
  %   & 24.21 & 19.44 & 183.17 & 131.64 & 27.40 & 71.86 & 27.50\\
  %   & 13.54 & 26.29 & 64.35 & 35.50 & 88.24 & 87.85 & 26.93 & 23.04\\
\cmidrule(l){2-11}

  & \multirow{5}{*}{ SplitFlow~\cite{yoon2025splitflow}}
& No Scheduler
  % & 28.04 & 18.62 & 199.44 & 158.13 & 27.45 & 69.99 & 27.52\\
  & 28.03\,(370.11) & 22.32\,(1.69) & 111.49\,(1472.11) & 85.31\,(1126.43) & 83.17\,(6.30) & 84.10\,(6.37) & \textbf{27.52} & \textbf{23.77}\\
&& Interval~\cite{kynkaanniemi2024applying}
  % & N/A & 18.96 & N/A & 147.42 & 27.39 & 71.49 & 27.43\\
  & 27.28\,(367.38) & 23.05\,(1.71) & 106.26\,(1430.99) & 77.64\,(1045.57) & 84.17\,(6.25) & 83.25\,(6.18) & 27.25 & 23.22\\
&& Monotone~\cite{wang2024analysis}
  % & N/A & 22.02 & N/A & 76.42 & 26.87 & 79.66 & 26.98\\
  & \underline{16.06\,(223.10)} & \textbf{26.25\,(1.89)} & \textbf{66.73\,(927.00)} & \underline{38.77\,(538.59)} & \textbf{88.12\,(6.34)} & \underline{85.94\,(6.19)} & 26.83 & 22.76\\
&& Zero-Init~\cite{fan2025cfg} 
  % & 21.66 & 19.85 & 170.21 & 121.57 & 27.23 & 73.37 & 27.30\\
  & 21.65\,(290.49) & 23.57\,(1.76) & 94.01\,(1261.39) & 65.07\,(873.08) & 84.90\,(6.33) & 85.36\,(6.36) & \underline{27.30} & \underline{23.57}\\
&& \cellcolor{gray!15}VAGS (Ours)
  & \cellcolor{gray!15}\textbf{14.67\,(200.34)} & \cellcolor{gray!15}\underline{25.99\,(1.90)} & \cellcolor{gray!15}\underline{73.43\,(1002.81)} & \cellcolor{gray!15}\textbf{38.41\,(524.55)} & \cellcolor{gray!15}\underline{87.31\,(6.39)} & \cellcolor{gray!15}\textbf{87.22\,(6.39)} & \cellcolor{gray!15}27.06 & \cellcolor{gray!15}23.28\\
  % && VAGS-Ratio (Ours)
  %   & 13.53 & 26.24 & 64.24 & 36.54 & 88.18 & 88.00 & 26.91 & 23.15\\
\midrule

%% ── DIV2K ──────────────────────────────────────────────────────────────
\multirow{10}{*}{\textbf{DIV2K}}
& \multirow{5}{*}{FlowEdit}
& No Scheduler
  % & 22.80 & 19.80 & 165.80 & 138.20 & 29.60 & 77.15 & 29.85\\
  & 18.27\,(202.19) & 20.27\,(1.83) & 145.55\,(1610.77) & 103.37\,(1143.98) & 82.15\,(7.42) & 82.31\,(7.44) & \textbf{30.06} & -\\
&& Interval~\cite{kynkaanniemi2024applying}
  % & N/A & 20.19 & N/A & 105.24 & 29.84 & 79.76 & 29.94\\
  & 18.11\,(200.82) & 19.94\,(1.80) & 144.88\,(1606.56) & 111.55\,(1236.97) & 79.51\,(7.17) & 82.32\,(7.42) & \underline{30.03} & -\\
&& Monotone~\cite{wang2024analysis} 
  % & N/A & 22.86 & N/A & 58.78 & 29.35 & 85.91 & 29.66\\
  & \underline{10.89\,(124.29)} & \underline{22.47\,(1.97)} & \underline{102.91\,(1174.56)} & \underline{63.91\,(729.43)} & \underline{83.64\,(7.33)} & \textbf{84.10\,(7.37)} & 29.60 & -\\
&& Zero-Init~\cite{fan2025cfg}
  % & N/A & 10.50 & N/A & 944.15 & 30.78 & 44.12 & 30.73\\
  & 194.67\,(2406.80) & 11.69\,(0.946) & 354.57\,(4383.72) & 867.59\,(10726.43) & 57.64\,(4.66) & 75.28\,(6.09) & 28.44 & -\\
&& \cellcolor{gray!15}VAGS (Ours)
  % & NA & 21.70 & NA & 75.90 & 29.87 & 82.35 & 30.01\\
  & \cellcolor{gray!15}\textbf{8.62\,(98.32)} & \cellcolor{gray!15}\textbf{23.61\,(2.07)} & \cellcolor{gray!15}\textbf{98.43\,(1122.67)} & \cellcolor{gray!15}\textbf{50.22\,(572.80)} & \cellcolor{gray!15}\textbf{84.41\,(7.40)} & \cellcolor{gray!15}\underline{84.07\,(7.37)} & \cellcolor{gray!15}29.61 & \cellcolor{gray!15}-\\
  % && VAGS-Ratio (Ours)
  %   & NA & 21.01 & NA & 86.87 & 29.81 & 81.56 & 29.91\\
  %   & 9.38 & 23.14 & 95.70 & 54.67 & 84.39 & 84.39 & 29.62 & -\\
\cmidrule(l){2-11}

  & \multirow{5}{*}{ SplitFlow~\cite{yoon2025splitflow}}
& No Scheduler & 17.61\,(196.32) & 20.53\,(1.84) & 141.99\,(1582.94) & 97.19\,(1083.50) & 82.63\,(7.41) & 82.44\,(7.39) & \underline{29.95} & -\\
% & 24.77 & 20.53 & 141.11 & 97.52 & 30.45 & 82.73 & 30.57\\
& & Interval~\cite{kynkaanniemi2024applying} & 16.01\,(178.96) & 20.53\,(1.84) & 136.04\,(1520.67) & 97.62\,(1091.20) & 80.55\,(7.21) & 82.65\,(7.39) & 29.91 & -\\
% & N/A & 20.82 & N/A & 91.16 & 29.62 & 81.19 & 29.87\\
& & Monotone~\cite{wang2024analysis} & 9.63\,(110.28) & 23.01\,(2.01) & \textbf{97.16\,(1112.69)} & \underline{56.81\,(650.59)} & \textbf{84.23\,(7.35)} & \textbf{84.43\,(7.37)} & 29.55 & -\\
% & N/A & 23.43 & N/A & 51.37 & 29.20 & 86.84 & 29.56\\
& & Zero-Init~\cite{fan2025cfg} & 46.72\,(468.47) & 14.83\,(1.48) & 378.94\,(3799.67) & 345.27\,(3462.06) & 48.95\,(4.88) & 75.93\,(7.57) & \textbf{31.58} & -\\
% & N/A & 21.86 & N/A & 73.26 & 29.30 & 82.98 & 29.39\\
& & \cellcolor{gray!15}VAGS (Ours) & \cellcolor{gray!15}\textbf{9.32\,(106.30)} & \cellcolor{gray!15}\textbf{23.28\,(2.04)} & \cellcolor{gray!15}\underline{102.58\,(1170.00)} & \cellcolor{gray!15}\textbf{53.57\,(611.01)} & \cellcolor{gray!15}\underline{84.01\,(7.37)} & \cellcolor{gray!15}\underline{84.08\,(7.37)} & \cellcolor{gray!15}29.61 & \cellcolor{gray!15}-\\
  % & 22.19 & 22.23 & 117.55 & 69.94 & 26.64 & 83.20 & 26.70\\
  % & & VAGS-Ratio (Ours) & 9.43 & 23.17 & 96.62 & 54.62 & 84.34 & 84.37 & 29.46 & -\\
  % & 23.63 & 21.01 & 131.95 & 90.77 & 26.62 & 81.12 & 26.74\\
\bottomrule
\end{tabular}%
}
\caption{[CLIP-scaled] Ablation of guidance scheduling strategies on FlowEdit and SplitFlow. VAGS gives the strongest overall preservation and edit-quality trade-off under the CLIP-scaled normalization.}
\label{tab:guidance_ablation_clipscaled}
\end{table*}

\paragraph{Generation SOTA.}
Scaling preserves the raw conclusion that VAGS-Gen leads on COCO17,
CUB-200, and Flickr30K. CLIPScore is the divisor and is therefore
reported unchanged; the scaled FID, IS, BLEU-4, METEOR, ROUGE-L, and
CLAIR rows keep VAGS-Gen at the top or tied position on all three
datasets. Self-Guidance's degradation, already visible in the raw
rows (FID \(40.98\) on COCO17), is amplified under scaling
(\(420.71\)) since its CLIPScore is the lowest in its block
(\(31.21\) vs.\ \(32.84\) for the other COCO17 rows).

\FloatBarrier
% Required packages: booktabs, multirow, graphicx, xcolor
\begin{table*}[t!]
\centering
\small
\setlength{\tabcolsep}{4.5pt}
\renewcommand{\arraystretch}{1.08}
\resizebox{\textwidth}{!}{%
\begin{tabular}{l l c c c c c c c}
\toprule
\multirow{2}{*}{\textbf{Dataset}} &
\multirow{2}{*}{\textbf{Method}} &
\multirow{2}{*}{\textbf{FID} {$\downarrow$}} &
\multirow{2}{*}{\textbf{IS} {$\uparrow$}} &
\multirow{2}{*}{\textbf{CLIPScore} {$\uparrow$}} &
\multicolumn{4}{c}{\textbf{Caption-based Metrics}} \\
\cmidrule(lr){6-9}
& & & & &
\textbf{BLEU-4} {$\uparrow$} &
\textbf{METEOR} {$\uparrow$} &
\textbf{ROUGE-L} {$\uparrow$} &
\textbf{CLAIR} {$\uparrow$} \\
\midrule

\multirow{4}{*}{\textbf{COCO17} \cite{lin2014microsoft}}
& SDv3.5 (CFG=7) & 28.46\,(263.89) & 33.64\,(3.63) & \textbf{32.84} & \underline{7.99\,(0.862)} & \underline{29.17\,(3.15)} & 35.11\,(3.79) & \underline{71.45\,(7.71)} \\
& SDv3.5 w/ A-Euler \cite{jin2025flops} & \underline{27.64\,(256.29)} & \underline{34.63\,(3.73)} & \textbf{32.84} & 7.96\,(0.858) & 29.09\,(3.14) & \underline{35.17\,(3.79)} & 71.23\,(7.68) \\
& SDv3.5 w/ Self-Guidance \cite{li2025self} & 40.98\,(420.71) & 28.54\,(2.78) & 31.21 & 6.52\,(0.635) & 26.18\,(2.55) & 31.59\,(3.08) & 62.26\,(6.06) \\
& \cellcolor{gray!15}SDv3.5 w/ VAGS (CFG=7, $\kappa$=1.0)
  & \cellcolor{gray!15}\textbf{26.07\,(241.73)} & \cellcolor{gray!15}\textbf{35.15\,(3.79)} & \cellcolor{gray!15}\textbf{32.84}
  & \cellcolor{gray!15}\textbf{8.28\,(0.893)} & \cellcolor{gray!15}\textbf{29.53\,(3.18)} & \cellcolor{gray!15}\textbf{35.21\,(3.80)} & \cellcolor{gray!15}\textbf{71.62\,(7.72)} \\
\midrule

\multirow{4}{*}{\textbf{CUB-200} \cite{welinder2010caltech}}
& SDv3.5 (CFG=7) & 24.92\,(237.39) & 4.81\,(0.505) & 32.40 & 0.13\,(0.014) & 16.57\,(1.74) & \underline{17.64\,(1.85)} & \textbf{66.69\,(7.00)} \\
& SDv3.5 w/ A-Euler \cite{jin2025flops} & \underline{22.98\,(217.56)} & \underline{5.05\,(0.533)} & \textbf{32.50} & 0.12\,(0.013) & 16.39\,(1.73) & 17.51\,(1.85) & 66.22\,(6.99) \\
& SDv3.5 w/ Self-Guidance \cite{li2025self} & 61.16\,(626.28) & 4.98\,(0.486) & 31.25 & \textbf{0.18\,(0.018)} & \textbf{17.85\,(1.74)} & \textbf{17.78\,(1.74)} & 62.28\,(6.08) \\
& \cellcolor{gray!15}SDv3.5 w/ VAGS (CFG=7, $\kappa$=1.0)
  & \cellcolor{gray!15}\textbf{20.60\,(195.63)} & \cellcolor{gray!15}\textbf{5.17\,(0.544)} & \cellcolor{gray!15}\underline{32.45}
  & \cellcolor{gray!15}\underline{0.15\,(0.016)} & \cellcolor{gray!15}\underline{17.00\,(1.79)} & \cellcolor{gray!15}17.52\,(1.84) & \cellcolor{gray!15}\underline{66.52\,(7.00)} \\
\midrule

\multirow{4}{*}{\textbf{Flickr30K} \cite{plummer2015flickr30k}}
& SDv3.5 (CFG=7) & 79.58\,(696.99) & \underline{17.95\,(2.05)} & \textbf{33.79} & \underline{4.12\,(0.470)} & \textbf{23.88\,(2.73)} & \textbf{29.30\,(3.35)} & \textbf{68.03\,(7.77)} \\
& SDv3.5 w/ A-Euler \cite{jin2025flops} & \underline{78.57\,(691.42)} & 17.83\,(2.03) & 33.71 & 3.91\,(0.444) & 22.99\,(2.61) & 29.00\,(3.30) & \underline{67.56\,(7.68)} \\
& SDv3.5 w/ Self-Guidance \cite{li2025self} & 93.07\,(932.04) & 14.44\,(1.44) & 31.60 & 3.35\,(0.335) & 20.80\,(2.08) & 26.27\,(2.62) & 57.35\,(5.73) \\
& \cellcolor{gray!15}SDv3.5 w/ VAGS (CFG=7, $\kappa$=0.9)
  & \cellcolor{gray!15}\textbf{75.65\,(663.36)} & \cellcolor{gray!15}\textbf{18.21\,(2.08)} & \cellcolor{gray!15}\underline{33.77}
  & \cellcolor{gray!15}\textbf{4.17\,(0.476)} & \cellcolor{gray!15}\underline{23.43\,(2.67)} & \cellcolor{gray!15}\underline{29.25\,(3.34)} & \cellcolor{gray!15}\underline{67.65\,(7.71)} \\
\bottomrule
\end{tabular}%
}
\caption{[CLIP-scaled] Text-to-image generation on COCO17, CUB-200, and Flickr30K. BLIP-2 \cite{li2023blip} is used to generate captions from the generated images, enabling evaluation with the caption-based metrics. VAGS-Gen yields consistent FID and IS gains, with the largest improvement on fine-grained generation.}
\label{tab:perf_comparison_clipscaled}
\end{table*}

\paragraph{Generation \(\kappa\) ablation.}
The optimal \(\kappa\) on each dataset is unchanged under scaling
(\(\kappa{=}1.0\) on COCO17 and CUB-200, \(\kappa{=}0.9\) on
Flickr30K), and the same collapse at \(\kappa{=}2.0\) on COCO17 is
visible in the scaled column (FID \(334.86\) vs.\ VAGS at
\(\kappa{=}1.0\) at \(241.73\)). The chosen modulation strength
therefore does not depend on whether non-CLIP metrics are reported
raw or scaled by alignment.

\FloatBarrier
% Required packages: booktabs, multirow, adjustbox, xcolor
\begin{table*}[t]
\centering
\scriptsize
\setlength{\tabcolsep}{4.0pt}
\renewcommand{\arraystretch}{1.10}
\caption{[CLIP-scaled] Ablation of $\kappa$ on COCO17, CUB-200, and Flickr30K. Top results are in bold and second-best results are underlined. We adopt $\kappa{=}1.0$ as the default.}
\label{tab:ablation_kappa_generation_clipscaled}
\begin{adjustbox}{width=\textwidth}
\begin{tabular}{@{}l l c c c c c c c@{}}
\toprule
\multirow[c]{2}{*}{\textbf{Dataset}} &
\multirow[c]{2}{*}{\textbf{Method}} &
\multirow[c]{2}{*}{\textbf{FID} $\downarrow$} &
\multirow[c]{2}{*}{\textbf{IS} $\uparrow$} &
\multirow[c]{2}{*}{\textbf{CLIPScore} $\uparrow$} &
\multicolumn{4}{c}{\textbf{Caption-based Metrics}} \\
\cmidrule(lr){6-9}
& & & & &
\textbf{BLEU-4} $\uparrow$ &
\textbf{METEOR} $\uparrow$ &
\textbf{ROUGE-L} $\uparrow$ &
\textbf{CLAIR} $\uparrow$ \\
\midrule

\multirow[c]{10}{*}{\textbf{COCO17}~\cite{lin2014microsoft}}
& SDv3.5 & 28.46\,(263.89) & 33.64\,(3.63) & 32.84 & 7.99\,(0.862) & 29.17\,(3.15) & 35.11\,(3.79) & 71.45\,(7.71) \\
& VAGS, $\kappa{=}0$   & 28.58\,(264.84) & 34.35\,(3.71) & 32.85 & 8.24\,(0.889) & 29.35\,(3.17) & 35.15\,(3.79) & 71.63\,(7.73) \\
& VAGS, $\kappa{=}0.1$ & 28.36\,(262.33) & 34.03\,(3.68) & 32.88 & 8.17\,(0.883) & 29.37\,(3.18) & 35.13\,(3.80) & 71.65\,(7.75) \\
& VAGS, $\kappa{=}0.3$ & 28.17\,(259.78) & 34.34\,(3.72) & \textbf{32.93} & 8.18\,(0.887) & 29.49\,(3.20) & 35.14\,(3.81) & \underline{72.09\,(7.82)} \\
& VAGS, $\kappa{=}0.5$ & 27.80\,(257.30) & \textbf{35.55\,(3.84)} & 32.87 & \textbf{8.59\,(0.928)} & \textbf{29.79\,(3.22)} & \textbf{35.61\,(3.85)} & \textbf{72.17\,(7.80)} \\
& VAGS, $\kappa{=}0.7$ & 27.23\,(251.72) & \underline{35.43\,(3.83)} & \underline{32.89} & \underline{8.38\,(0.907)} & 29.47\,(3.19) & 35.40\,(3.83) & 71.87\,(7.77) \\
& VAGS, $\kappa{=}0.9$ & 26.77\,(248.07) & 35.27\,(3.81) & 32.85 & 8.28\,(0.894) & \underline{29.53\,(3.19)} & \underline{35.46\,(3.83)} & 71.55\,(7.72) \\
\rowcolor{gray!12}
& VAGS, $\kappa{=}1.0$ & \underline{26.07\,(241.73)} & 35.15\,(3.79) & 32.84 & 8.28\,(0.893) & \underline{29.53\,(3.18)} & 35.21\,(3.80) & 71.62\,(7.72) \\
& VAGS, $\kappa{=}1.5$ & \textbf{25.93\,(243.24)} & 33.03\,(3.52) & 32.65 & 7.78\,(0.829) & 28.75\,(3.06) & 34.47\,(3.67) & 70.38\,(7.50) \\
& VAGS, $\kappa{=}2.0$ & 33.99\,(334.86) & 27.54\,(2.80) & 31.86 & 6.90\,(0.700) & 27.23\,(2.76) & 32.84\,(3.33) & 66.13\,(6.71) \\
\midrule

\multirow[c]{6}{*}{\textbf{CUB-200}~\cite{welinder2010caltech}}
& SDv3.5 & 24.92\,(237.39) & 4.81\,(0.505) & 32.40 & 0.13\,(0.014) & 16.57\,(1.74) & \textbf{17.64\,(1.85)} & 66.69\,(7.00) \\
& VAGS, $\kappa{=}0.3$ & 23.78\,(225.27) & 5.00\,(0.528) & 32.49 & 0.13\,(0.014) & 16.72\,(1.76) & \underline{17.54\,(1.85)} & \underline{66.75\,(7.05)} \\
& VAGS, $\kappa{=}0.5$ & 22.97\,(217.07) & 5.08\,(0.538) & \textbf{32.53} & \underline{0.14\,(0.015)} & 16.79\,(1.78) & 17.48\,(1.85) & \textbf{66.87\,(7.08)} \\
& VAGS, $\kappa{=}0.7$ & 22.19\,(209.82) & 5.07\,(0.536) & \underline{32.52} & 0.13\,(0.014) & 16.82\,(1.78) & 17.44\,(1.84) & 66.74\,(7.06) \\
& VAGS, $\kappa{=}0.9$ & \underline{21.30\,(201.78)} & \textbf{5.19\,(0.548)} & 32.49 & \textbf{0.15\,(0.016)} & \underline{16.93\,(1.79)} & 17.48\,(1.85) & 66.57\,(7.03) \\
\rowcolor{gray!12}
& VAGS, $\kappa{=}1.0$ & \textbf{20.60\,(195.63)} & \underline{5.17\,(0.544)} & 32.45 & \textbf{0.15\,(0.016)} & \textbf{17.00\,(1.79)} & 17.52\,(1.84) & 66.52\,(7.00) \\
\midrule

\multirow[c]{5}{*}{\textbf{Flickr30K}~\cite{plummer2015flickr30k}}
& SDv3.5 & 79.58\,(696.99) & 17.95\,(2.05) & 33.79 & 4.12\,(0.470) & 23.88\,(2.73) & 29.30\,(3.35) & 68.03\,(7.77) \\
& VAGS, $\kappa{=}0.5$ & 77.92\,(680.84) & 18.32\,(2.10) & \textbf{33.83} & \textbf{4.54\,(0.520)} & \underline{24.46\,(2.80)} & \textbf{30.19\,(3.46)} & \textbf{68.40\,(7.83)} \\
& VAGS, $\kappa{=}0.7$ & 77.32\,(675.60) & 18.23\,(2.09) & \textbf{33.83} & 4.30\,(0.492) & \textbf{24.47\,(2.80)} & \underline{29.89\,(3.42)} & 67.82\,(7.76) \\
& VAGS, $\kappa{=}0.9$ & \textbf{75.65\,(663.36)} & 18.21\,(2.08) & \underline{33.77} & 4.17\,(0.476) & 23.43\,(2.67) & 29.25\,(3.34) & 67.65\,(7.71) \\
\rowcolor{gray!12}
& VAGS, $\kappa{=}1.0$ & \underline{76.08\,(667.13)} & \textbf{18.58\,(2.12)} & \underline{33.77} & \underline{4.31\,(0.492)} & 23.66\,(2.70) & 29.34\,(3.35) & \underline{67.93\,(7.75)} \\
\bottomrule
\end{tabular}
\end{adjustbox}
\end{table*}

\paragraph{Matched-mean control.}
Setting CFG to the per-trajectory mean adaptive value
(\(\bar\lambda{=}5.955\), \(\kappa{=}0\)) improves COCO17 scaled FID
marginally over fixed CFG (\(262.22\) vs.\ \(263.89\)), while
VAGS-Gen at \(\kappa{=}1.0\) reaches \(241.73\), a further \(7.8\%\)
reduction. Scaling agrees with the raw conclusion: the gain comes
from step-level modulation, not from a shifted average scale.

\FloatBarrier
% Required packages: booktabs, multirow, graphicx, xcolor
\begin{table*}[t!]
\centering
\small
\setlength{\tabcolsep}{5.0pt}
\renewcommand{\arraystretch}{1.12}
\resizebox{\textwidth}{!}{%
\begin{tabular}{@{}l l c c c c c c c@{}}
\toprule
\multirow[c]{2}{*}{\textbf{Dataset}} &
\multirow[c]{2}{*}{\textbf{Method}} &
\multirow[c]{2}{*}{\textbf{FID} $\downarrow$} &
\multirow[c]{2}{*}{\textbf{IS} $\uparrow$} &
\multirow[c]{2}{*}{\textbf{CLIPScore} $\uparrow$} &
\multicolumn{4}{c}{\textbf{Caption-based Metrics}} \\
\cmidrule(lr){6-9}
& & & & &
\textbf{BLEU-4} $\uparrow$ &
\textbf{METEOR} $\uparrow$ &
\textbf{ROUGE-L} $\uparrow$ &
\textbf{CLAIR} $\uparrow$ \\
\midrule

\multirow[c]{3}{*}{\textbf{COCO17}~\cite{lin2014microsoft}}
& SDv3.5 (CFG=7)
& 28.46\,(263.89) & 33.64\,(3.63) & \textbf{32.84}
& 7.99\,(0.862) & 29.17\,(3.15) & 35.11\,(3.79) & 71.45\,(7.71) \\

& SDv3.5 w/ VAGS (CFG=5.955, $\kappa{=}0$)
& \underline{28.28\,(262.22)} & \underline{34.63\,(3.73)} & \textbf{32.84}
& \underline{8.20\,(0.884)} & \textbf{29.69\,(3.20)} & \textbf{35.37\,(3.81)} & \textbf{72.27\,(7.79)} \\

& \cellcolor{gray!15}SDv3.5 w/ VAGS (CFG=7, $\kappa{=}1.0$)
& \cellcolor{gray!15}\textbf{26.07\,(241.73)} & \cellcolor{gray!15}\textbf{35.15\,(3.79)} & \cellcolor{gray!15}\textbf{32.84}
& \cellcolor{gray!15}\textbf{8.28\,(0.893)} & \cellcolor{gray!15}\underline{29.53\,(3.18)} & \cellcolor{gray!15}\underline{35.21\,(3.80)} & \cellcolor{gray!15}\underline{71.62\,(7.72)} \\

\bottomrule
\end{tabular}%
}
\caption{[CLIP-scaled] Constant-CFG vs.\ VAGS-Gen on COCO17. A fixed scale at the mean adaptive value
($\bar{\lambda}=5.955$) barely outperforms the baseline, confirming that the gains stem
from dynamic modulation rather than a shifted average guidance level.}
\label{tab:average_cfg_fix_cfg_clipscaled}
\end{table*}

%%%%%%%%%%%%%%%%%%%%%%%%%%%%%%%%%%%%%%%%%%%%%%%%%%%%%%%%%%%%

% \clearpage
% \input{checklist.tex}

\end{document}